\documentclass[11pt]{article}
\usepackage[a4paper,margin=2.4cm]{geometry}
\usepackage[round,authoryear]{natbib}
\usepackage{setspace}
\usepackage[section]{placeins}
\usepackage{amsmath,amssymb,amsthm,mathtools}
\usepackage{graphicx}
\usepackage{booktabs}
\usepackage{array}
\usepackage{xcolor}
\usepackage{enumitem}
\usepackage{xurl}
\usepackage[hidelinks]{hyperref}
\usepackage[capitalise,noabbrev]{cleveref}

\newcommand{\R}{\mathbb{R}}
\newcommand{\norm}[1]{\lVert #1\rVert}
\newcommand{\ip}[2]{\langle #1,#2\rangle}
\newcommand{\tr}{\operatorname{tr}}
\newcommand{\range}{\operatorname{range}}
\newcommand{\spn}{\operatorname{span}}
\newcommand{\supp}{\operatorname{supp}}
\newcommand{\din}{d_{\mathrm{in}}}
\newcommand{\dout}{d_{\mathrm{out}}}
\newcommand{\T}{^{\top}}
\newcommand{\M}{\mathcal{M}}
\newcommand{\F}{\mathcal{F}}

\newtheorem{theorem}{Theorem}
\newtheorem{lemma}{Lemma}
\newtheorem{proposition}{Proposition}
\newtheorem{corollary}{Corollary}
\theoremstyle{definition}
\newtheorem{definition}{Definition}
\theoremstyle{remark}

\crefname{theorem}{Theorem}{Theorems}
\crefname{proposition}{Proposition}{Propositions}
\crefname{lemma}{Lemma}{Lemmas}
\crefname{corollary}{Corollary}{Corollaries}
\crefname{definition}{Definition}{Definitions}
\crefname{remark}{Remark}{Remarks}

\newcommand{\mref}[1]{\cref{#1}}
\newcommand{\Mref}[1]{\Cref{#1}}

\title{\textbf{Novelty geometry, plasticity locus, and interference in hierarchical neural networks}}
\author{
  Jasmeet Singh Bindra\thanks{Corresponding author: \texttt{d24059@students.iitmandi.ac.in}. Indian Knowledge System and Mental Health Applications Centre, Indian Institute of Technology Mandi, Kamand, Mandi 175075, Himachal Pradesh, India.}
  \and
  Siddharth Panwar\thanks{School of Computing and Electrical Engineering, Indian Institute of Technology Mandi, Kamand, Mandi 175075, Himachal Pradesh, India.}
}
\date{}

\begin{document}
\maketitle

\begin{abstract}
When a hierarchical neural network adapts to a new item, which levels change and what interference cost follows? We analyse this in deep linear networks and test selected predictions in nonlinear predictive-coding, convolutional, and residual networks. Onset gradients lie on the contiguous plasticity window reached by both bottom-up activity and top-down error. In balanced networks, pure novelty can be classified into four types relative to learned input and output subspaces. A novel input with a familiar outcome activates only the bottom level, where one-shot or bottom-only updates spare old inputs in the learned subspace; joint all-level gradient flow obeys a conservation law yielding exact single-mode interference, while mode-locality confines output interference to touched modes under arbitrary step sizes. A familiar input with a novel outcome activates only the readout. For doubly novel items in balanced networks with spare hidden capacity and old inputs spanning the learned input subspace, depth $L\ge3$ yields a saddle where every negative-curvature direction perturbs at least one old output at first order while aligned interference-free flow escapes in time $\Theta(\varepsilon^{-(L-2)})$. We bound upper-level interference, derive a cross-moment condition for safe partial replay, and show that raw-error precision contrasts localise predictive-coding plasticity. Nonlinear experiments support this localisation, cutting median old-digit accuracy loss from 23.8 to 0.45 percentage points on unseen MNIST digits, while pre-learning layer tangent kernels forecast stage-wise interference in CIFAR-10 CNNs ($r=0.79$, $\rho=0.83$) and old-class Kullback--Leibler (KL) divergence across nonlinear ResNet-18 blocks under cross-entropy ($r=0.942$, $\rho=0.90$; optimal stage in 52/60 familiar-output conditions).
\end{abstract}


\bigskip

\section{Introduction}
\label{sec:intro}

Gradient-based learning in neural networks suffers from catastrophic interference: acquiring new associations can overwrite previously stored ones \citep{mccloskey1989,french1999}. Standard remedies penalise changes to important parameters \citep{kirkpatrick2017,zenke2017}, interleave old experience through replay \citep{mcclelland1995,vandeven2020}, or separate fast episodic storage from slow semantic integration \citep{mcclelland1995,kumaran2016}. Although these approaches address retention at the whole-network level, biological cortex and deep artificial architectures are layered hierarchies. A foundational question therefore remains open: \emph{when a trained hierarchy acquires a new item, which of its levels change, and how does that locus govern both acquisition speed and interference with prior knowledge?}

The locus of plasticity carries direct functional consequences: modifying early levels alters representations shared across downstream computations, whereas modifying upper levels alters how existing representations are read out. Five lines of work bear on this problem from complementary perspectives. In perceptual learning, reverse hierarchy theory proposes that adaptation begins at high levels and descends to sensory stages only when fine discriminations require it \citep{ahissar1997,ahissar2004}, a pattern observed empirically in deep visual networks \citep{wenliang2018}. In memory consolidation, schema theory documents rapid neocortical integration of schema-consistent associations alongside slow, interfering learning of inconsistent ones \citep{tse2007,gilboa2017}, modelled in connectionist accounts that treat neocortex as an undivided block \citep{mcclelland2013}. In theoretical deep learning, exact dynamical solutions for deep linear networks characterise learning trajectories from prior knowledge and task-similarity effects on forgetting \citep{saxe2014,saxe2019,braun2022,lee2021,domine2025}, while tangent-kernel analyses show that forgetting concentrates in low-rank output eigenmodes and motivates parameter-efficient continual fine-tuning \citep{hidekel2026,liu2026}. In continual learning algorithms, orthogonal gradient and subspace-projection methods actively project parameter updates onto subspaces orthogonal to prior gradients or layerwise feature spaces \citep{farajtabar2020,saha2021,tang2021,saha2023}, whereas empirical fine-tuning studies show that early blocks adapt best to input shifts, upper layers suit label shifts \citep{surgical2023}, and forgetting concentrates in higher layers \citep{ramasesh2021}. Unlike subspace-projection algorithms that impose orthogonality explicitly, what remains missing is a unified, layer-resolved theory of where \emph{unconstrained} gradient-based plasticity naturally lands for a given item and trained network, how rapidly it progresses, and what interference cost each locus entails.

Here we develop that theory analytically in deep linear networks and evaluate its predictions in nonlinear predictive-coding, convolutional, and residual networks. The mathematical starting point is elementary: the weight gradient at level $l$ is the outer product of a top-down error signal and a bottom-up activity vector, so a level changes only where both signals are nonzero. The substance lies in the structural and dynamical consequences of this factorisation in a trained hierarchy, where learned subspaces constrain signal propagation. Our contributions centre on four results.
\begin{enumerate}[leftmargin=*,itemsep=1pt]
\item \textbf{Plasticity window, novelty geometry, and layer-resolved kernel} (\cref{thm:window,cor:types,cor:blocks,prop:relu,thm:kernel,thm:cost,cor:aniso}). Onset gradients in deep linear networks (and single-item fixed patterns of positively homogeneous piecewise-linear units) lie on the contiguous band $[\dout,\din]$ reached by both forward activity and backward error, yielding four pure novelty types in balanced networks. A layer-resolved tangent kernel governs onset speed, transfer, and first-order interference, separating the minimum cost of upper-level updates from the Kantorovich excess of the local target delta rule.
\item \textbf{Input-novelty dynamics, mode-locality, and depth-dependent saddles} (\cref{thm:typeB,thm:modes,thm:saddle}). For novel inputs with familiar outcomes (type~B), bottom-only and one-shot updates leave old inputs in $\spn V$ unchanged; joint all-level updates obey an exact single-mode interference law under gradient flow and mode-locality under arbitrary gradient-descent step-size schedules. For doubly novel items (type~D) at depth $L\ge3$, every negative-curvature direction perturbs at least one old output at first order whereas aligned interference-free flow escapes in time $\Theta(\varepsilon^{-(L-2)})$.
\item \textbf{Precision-selected localisation and safe partial replay} (\cref{thm:pc,thm:learnedprec,thm:replay}). In equilibrium predictive coding, raw-error updates under a precision contrast localise plasticity whereas precision-weighted updates lack inverse-precision attenuation; normalised error-energy shares under free-energy precision updates converge doubly exponentially in the common-eigenvector error case. For partial replay, vanishing of an uncentred mode cross-moment protects unrehearsed memories across all time.
\item \textbf{Empirical evaluation in nonlinear hierarchies} (\cref{sec:mechanism,sec:cnn}). Nonlinear predictive-coding benchmarks (four-layer $\tanh$/ReLU and modified MNIST) support raw-error precision localisation alongside explicit layer-1 and learning-rate-scaled controls. In CIFAR-10 six-layer convolutional networks (squared loss) and ResNet-18 (cross-entropy), output-side novelty concentrates onset gradients in the readout, intermediate stages remain optimal under pixel-space shifts consistent with local patch overlap, and pre-learning layer tangent kernels forecast stage-wise interference, old-class Kullback--Leibler (KL) divergence, and optimal stage selection.
\end{enumerate}

\Cref{tab:summary} summarises the scope of each analytical result; complete proofs, robustness bounds, auxiliary schema and perceptual-learning illustrations, and computational details appear in the Supplementary Information (SI).

\begin{table}[!htbp]
\centering\footnotesize
\caption{Summary of analytical results and their scope. A0: linear and bias-free; A1: balanced; A2: learned-map spectral structure (compact SVD $USV\T$, distinct singular values); A3: spare hidden capacity (\cref{sec:setting}).}
\label{tab:summary}
\begin{tabular}{@{}p{0.30\textwidth}p{0.28\textwidth}p{0.34\textwidth}@{}}
\toprule
Result & Scope & Kind of statement \\
\midrule
\Cref{thm:window}, plasticity window & A0; gradient signals & exact \\
\Cref{cor:types,cor:blocks}, novelty types & A0--A2 & exact (onset window; first-order blocks) \\
\Cref{prop:relu}, piecewise linear & positively homogeneous activations & exact for single item within one pattern \\
\Cref{thm:kernel}, layer kernel & A0 & first order in $\eta$; exact for one plastic level \\
\Cref{thm:cost}, cost by locus & A0; reachability conditions stated & lower bound; exact local-rule cost \\
\Cref{thm:typeB}, type B & A0--A2; single mode for (c)--(d) & exact; small-$g$ expansion labelled \\
\Cref{thm:modes}, mode-locality & A0--A2, $e_\perp=0$ & exact for any step sizes \\
\Cref{thm:saddle}, type-D saddle & A0--A3 & local curvature; aligned gradient flow \\
\Cref{thm:pc}, precision & A0 & asymptotic in $\Pi$; first order in $\eta$ for (d) \\
\Cref{thm:learnedprec}, learned precision & A0; common-eigenvector $e$ for (a) & exact recursion; unique maximiser \\
\Cref{thm:replay}, partial replay & A0--A2, $e_\perp=0$, $\lambda>0$ for (b)--(c) & exact invariance; exact leakage rate \\
\bottomrule
\end{tabular}
\end{table}

\section{Setting and novelty geometry}
\label{sec:setting}

\paragraph{Network, loss and interference}
We study deep linear networks $f(x)=W_LW_{L-1}\cdots W_1x$ with $W_l\in\R^{n_l\times n_{l-1}}$, $L\ge2$ and no biases, the standard analytically tractable model of hierarchical learning \citep{saxe2014,saxe2019}. We write $W_{b:a}:=W_b\cdots W_a$ for $b\ge a$ and $W_{a-1:a}:=I$. For an input $x$, the \emph{bottom-up activity} at level $j$ is $h_j(x):=W_{j:1}x$ ($h_0=x$). A new item $(x,y)$ has error $e:=y-f(x)$ and loss $\ell=\tfrac12\norm{e}^2$; the \emph{top-down error signal} at level $l$ is $\beta_l(e):=W_{L:l+1}\T e$ ($\beta_L=e$). The old items have inputs $x_1,\dots,x_P$, collected in $X_o$. \emph{Interference} is the summed squared change of the old outputs, $I:=\sum_o\norm{f'(x_o)-f(x_o)}^2$, where $f'$ is the network after learning. In the convolutional-network experiments we report the mean over a retention set instead, together with the change in old-task error (\cref{sec:cnn}).

\paragraph{Assumptions}
Each result states which of the following it uses.
\begin{itemize}[leftmargin=2.2em,itemsep=0pt]
\item[A0] Linear and bias-free. (\Cref{prop:relu} and SI \cref{sec:relu} treat piecewise-linear networks.)
\item[A1] Balanced: $W_{l+1}\T W_{l+1}=W_lW_l\T$ for $l=1,\dots,L-1$.
\item[A2] Learned-map spectral structure: $W_{L:1}=USV\T$ is a compact singular value decomposition of rank $k$ with distinct singular values $s_1>\dots>s_k>0$.
\item[A3] Spare capacity: $n_l>k$ for every hidden level.
\item[A4] Isotropic old inputs: $X_oX_o\T=cVV\T$ for some $c>0$ (used only in \cref{cor:aniso}).
\end{itemize}
A1 and A2 describe networks trained from small initial weights: under gradient flow the differences $W_{l+1}\T W_{l+1}-W_lW_l\T$ are conserved \citep{arora2018,du2018}, so small initialisation gives approximate balance, and for whitened inputs training converges to the input--output correlation \citep{saxe2014}. Under A1--A2, $W_l=R_{l+1}S^{1/L}R_l\T$ with orthonormal $R_l$, $R_1=V$ and $R_{L+1}=U$ (SI \cref{lem:balanced}). The columns of $V$ and $U$ span the \emph{learned input and output subspaces}. SI \cref{sec:robust} quantifies what happens when A1--A2 hold only approximately.

\paragraph{Novelty types}
Novelty is defined relative to the learner's current input and output subspaces.

\begin{definition}[Pure novelty types]\label{def:types}
Assume A0--A2. A new item $(x,y)$ with nonzero input $x\neq0$ and nonzero error $e=y-f(x)\neq0$ has \emph{familiar input} if $x\in\spn V$ and \emph{novel input} if $x\perp\spn V$; it has \emph{familiar output} if $e\in\spn U$ and \emph{novel output} if $e\perp\spn U$. The four pure types are A (familiar input, familiar output), B (novel input, familiar output), C (familiar input, novel output) and D (novel input, novel output). Any item with $x\neq0$ and $e\neq0$ that is none of these is \emph{mixed}.
\end{definition}

Every item decomposes into the four blocks $(x_\parallel,e_\parallel)$, $(x_\perp,e_\parallel)$, $(x_\parallel,e_\perp)$ and $(x_\perp,e_\perp)$, where $\parallel$ and $\perp$ denote the components in and orthogonal to $\spn V$ (for $x$) and $\spn U$ (for $e$). Note that the learned input subspace $\spn V$ need not coincide with the old-data input span $\spn X_o$ in general (for instance, if a pre-trained network discarded low-variance old input directions); exact protection statements below therefore specify when old inputs lie in $\spn V$ or satisfy $x_o\perp x$. The exact results below are stated for pure types or for these blocks. Two coarser quantities remain meaningful for mixed items and for general networks: the \emph{reach} of the input and of the error (\cref{def:reach}), which fixes the window at the onset of learning, and the layer-resolved tangent kernel (\cref{thm:kernel}). A type-B item is a new cue mapped onto a known outcome, for example a new flavour paired with a location in a known arena \citep{tse2007}; type C is a new response to known stimuli; type D is a new cue mapped onto a new outcome.

\paragraph{Learning rules}
Gradient descent (backpropagation) is the reference learner for \cref{sec:window,sec:dynamics}; its update at level $l$ is the outer product of $\beta_l$ and $h_{l-1}$. Other rules that update weights by outer products, such as predictive coding \citep{rao1999,whittington2017,millidge2022} and target propagation \citep{lee2015,meulemans2020}, compute their pre- and postsynaptic factors differently, so the window does not transfer to them automatically. \Cref{sec:pc} analyses equilibrium predictive coding separately.

\section{The plasticity window and the layer-resolved kernel}
\label{sec:window}

\subsection{The window}

Writing $f=W_{L:l+1}W_lh_{l-1}$ gives
\begin{equation}\label{eq:grad}
\nabla_{W_l}\ell=-\beta_l(e)\,h_{l-1}(x)\T,\qquad \norm{\nabla_{W_l}\ell}_F=\norm{\beta_l}\,\norm{h_{l-1}}.
\end{equation}
A level therefore has a nonzero gradient if and only if both signals are nonzero there. In a linear network their reach is nested: once the input is annihilated at some level it remains zero above, and once the error is annihilated on its way down it remains zero below, because $h_{j+1}=W_{j+1}h_j$ and $\beta_{l-1}=W_l\T\beta_l$.

\begin{definition}[Reach]\label{def:reach}
For $x\neq0$ and $e\neq0$, the \emph{input reach} is $\din(x):=1+\max\{j\in\{0,\dots,L-1\}:h_j(x)\neq0\}$ and the \emph{error reach} is $\dout(e):=\min\{l:\beta_l(e)\neq0\}$.
\end{definition}

\begin{theorem}[Plasticity window]\label{thm:window}
Assume A0. For every $x\neq0$ and $e\neq0$,
\[
\nabla_{W_l}\ell\neq0\iff\dout(e)\le l\le\din(x).
\]
If $\dout(e)>\din(x)$, all gradients vanish although $\ell>0$: the network is at a critical point of the item's loss that is not a global minimum. Under A1--A3 and for a pure type-D item it is a saddle (\cref{thm:saddle}).
\end{theorem}

\begin{proof}
By \cref{eq:grad}, $\nabla_{W_l}\ell\neq0$ if and only if $h_{l-1}(x)\neq0$ and $\beta_l(e)\neq0$. By nesting, $\{l:h_{l-1}\neq0\}=\{1,\dots,\din\}$ and $\{l:\beta_l\neq0\}=\{\dout,\dots,L\}$; both sets are nonempty because $h_0=x$ and $\beta_L=e$. If the window is empty, $\ell>0$ while $\ell=0$ is attainable by some choice of weights, so the critical point is not a global minimum.
\end{proof}

The theorem concerns the gradient, whose factors are the feedforward activity $h_{l-1}$ and the transpose-propagated error $\beta_l$. Any rule whose update at level $l$ is an outer product vanishes where either of its factors vanishes, but whether its factors have the same reach depends on how they are computed; for equilibrium predictive coding they do not (\cref{thm:pc}d).

\begin{figure}[!htbp]
\centering
\includegraphics[width=\textwidth]{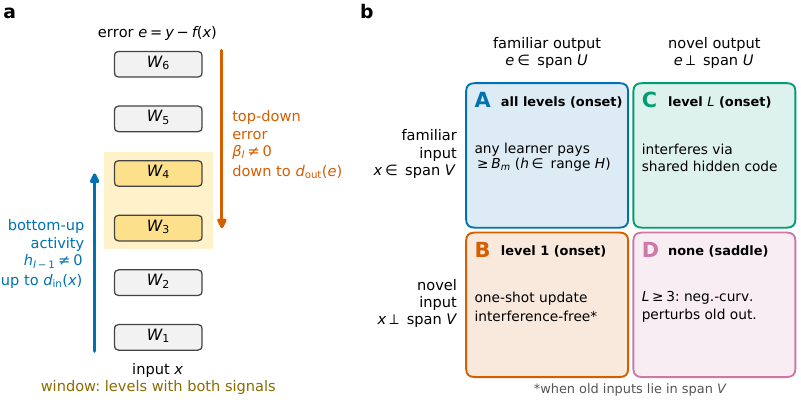}
\caption{\textbf{Plasticity window and pure novelty types.} (a) In deep linear networks, the gradient at level $l$ is the outer product of the top-down error signal $\beta_l$ and the bottom-up activity $h_{l-1}$ (\cref{eq:grad}). Activity reaches up to $\din(x)$ and error down to $\dout(e)$, confining onset gradient-based plasticity to $[\dout,\din]$ (\cref{thm:window}). (b) Measured against the learned input and output subspaces $\spn V$ and $\spn U$, nonzero items partition into four pure novelty types (\cref{def:types}), each with a distinct onset locus and dynamical consequence (\cref{cor:types,thm:cost,thm:typeB,thm:saddle}).}
\label{fig:concept}
\end{figure}

In a balanced network with learned spectral structure (A1--A2), the reach is binary and the onset window is set by the novelty type (\cref{fig:concept}).

\begin{corollary}[Novelty types and onset windows]\label{cor:types}
Assume A0--A2. Then $\din(x)=L$ if $V\T x\neq0$ and $\din(x)=1$ otherwise, and $\dout(e)=1$ if $U\T e\neq0$ and $\dout(e)=L$ otherwise. The window is $\{1,\dots,L\}$ for type A, $\{1\}$ for type B, $\{L\}$ for type C and empty for type D; a mixed item has the window of the pure type with the same pattern of vanishing and nonvanishing projections $V\T x$, $U\T e$. The gradient norms are $\norm{x}\norm{S^{\frac{L-1}{L}}U\T e}$ at $l=1$, $\norm{S^{\frac{l-1}{L}}V\T x}\norm{S^{\frac{L-l}{L}}U\T e}$ for $1<l<L$, and $\norm{S^{\frac{L-1}{L}}V\T x}\norm{e}$ at $l=L$.
\end{corollary}

\begin{corollary}[Block decomposition]\label{cor:blocks}
Assume A0--A2. For any update $\{\Delta W_l\}$, the first-order change $\Delta F=\sum_lW_{L:l+1}\Delta W_lW_{l-1:1}$ of the input--output map satisfies $P_{U^\perp}\Delta F\,P_{V^\perp}=0$. For a gradient step, the $(x_\perp,e_\parallel)$ block is written only at level 1, the $(x_\parallel,e_\perp)$ block only at level $L$, and the $(x_\parallel,e_\parallel)$ block at every level.
\end{corollary}

Beyond first order the window can widen during learning, and the blocks of a mixed item then interact. For example, let $U=V=\spn(e_1)$, $x=e_2$ and $e=e_1+e_2$ ($L=3$, $W_l=\mathrm{diag}(1,0)$). At onset the window is $\{1\}$, and no update of level 1 alone can produce the $e_2$ output component, which is a type-D block. Under gradient descent on all levels, however, the level-1 update opens a channel for $x$ through the old mode, the $e_2$ component is then written at level $L$ through that channel, and the old input $e_1$ acquires an $e_2$ response (interference 1.46; SI \cref{sec:verification}, Example~1). Exact fitting and interference-free statements below therefore specify the blocks they apply to.

\paragraph{Piecewise-linear networks}
The window carries over to a single item in networks with positively homogeneous piecewise-linear activations within its fixed activation pattern.

\begin{proposition}[Piecewise-linear networks]\label{prop:relu}
Let \[f(x)=W_L\phi(W_{L-1}\cdots\phi(W_1x))\] with no biases and $\phi$ applied elementwise, where $\phi(z)=z$ for $z>0$ and $\phi(z)=az$ for $z<0$ with $a\ge0$ (ReLU for $a=0$, leaky ReLU for $a>0$). At a single input $x\neq0$ whose pre-activations $z_l(x)$ are all nonzero and with $e\neq0$, let $D_l=\mathrm{diag}(\phi'(z_l(x)))$ for $l<L$ and $D_L=I$, and set $\tilde W_l=D_lW_l$. Define the activation-gated forward and backward signals $\tilde h_{l-1}:=\tilde W_{l-1:1}x$ ($\tilde h_0=x$) and $\tilde\beta_l:=D_l\tilde W_{L:l+1}\T e$ ($\tilde\beta_L=e$). Then $\phi(z_l)=D_lz_l$,
\[\nabla_{W_l}\ell=-\tilde\beta_l\,\tilde h_{l-1}\T,\]
and the plasticity-window characterisation of \cref{thm:window} holds within this fixed activation pattern when input reach and error reach are defined from $\tilde h_{l-1}$ and $\tilde\beta_l$.
\end{proposition}

The effective linear representation is local to the single-item activation pattern at $x$: it describes the onset gradient at $x$, not a finite update that switches unit states. Positive homogeneity is required: for a clipped activation $\phi(z)=\min(\max(z,0),1)$, the identity $\phi(z)=D(z)z$ fails and the effective linearisation misestimates the gradient (SI \cref{sec:verification}, Example~2). In ReLU networks an input exactly orthogonal to all first-layer weight rows lies on the kink, so the ReLU analogue of type B is an input that drives only units silent for all old items (SI \cref{thm:unit}). Trainable biases introduce a shared constant input coordinate, so orthogonality in the original input space does not generally imply orthogonality in the augmented representation ($\tilde x\T\tilde x_o = x\T x_o + 1$). Consequently, exact orthogonality-based protection need not persist when biases are also updated; we state the theory for bias-free networks and evaluate biased networks numerically (\cref{sec:cnn}). When A1--A2 hold only approximately, out-of-window gradients are bounded by the residual weight on directions the training data never use, which gradient descent leaves unchanged without weight decay and shrinks by $(1-\eta\lambda)$ per step with decoupled weight decay $\lambda$ (SI \cref{lem:R1,prop:soft}).

\Cref{fig:dln}a shows the window in networks trained from small random weights, with novelty types defined from each network's own singular value decomposition: for type-B items 99.9\% of the squared gradient norm falls on level 1, for type C 99.8\% falls on level $L$, and type-D gradients are more than $10^4$ times smaller than type-A gradients.

\subsection{Speed and first-order interference: a layer-resolved kernel}

\begin{theorem}[Layer-resolved kernel]\label{thm:kernel}
Assume A0 and apply the step $\Delta W_l=-\eta\nabla_{W_l}\ell$ to a set of levels $\mathcal L$. For every probe input $x'$,
\[
f'(x')-f(x')=\eta\sum_{l\in\mathcal L}\ip{h_{l-1}(x)}{h_{l-1}(x')}\,K_l\,e+O(\eta^2),\qquad K_l:=W_{L:l+1}W_{L:l+1}\T.
\]
If $\mathcal L$ contains a single level, the expansion is exact.
\end{theorem}

$\Theta(x',x)=\sum_l\ip{h_{l-1}(x)}{h_{l-1}(x')}K_l$ is the neural tangent kernel \citep{jacot2018} of the network, resolved by level: each level contributes the product of a bottom-up overlap and a top-down kernel. The same kernel governs the speed of learning ($x'=x$), interference ($x'=x_o$) and transfer ($x'$ a related item).

\begin{corollary}[Speed and one-step interference]\label{cor:speed}
Assume A0--A2, unit-norm $x$ and $e$, and that every familiar input or error component is aligned with the corresponding learned singular mode $i$ (that is, $V\T x=e_i$ for types A and C, and $U\T e=e_i$ for types A and B). The initial relative rate of loss decrease under gradient flow, $-\dot\ell/\ell$, is $2L\,s_i^{2(L-1)/L}$ for type A, $2s_i^{2(L-1)/L}$ for types B and C, and $0$ for type D. One gradient step changes old outputs by exactly $\eta\ip{x}{x_o}K_1e$ for type B, which vanishes for every $x_o\perp x$, and by exactly $\eta\ip{h_{L-1}(x)}{h_{L-1}(x_o)}e$ for type C, which gives old items outputs along the novel direction $e$ in proportion to their hidden-code overlap with $x$.
\end{corollary}

Under these matched conditions, familiar items are learned $L$ times faster at onset than one-sided novelty because all $L$ levels contribute, and doubly novel items make no progress (\cref{fig:dln}b).

\section{Interference and learning dynamics}
\label{sec:dynamics}

\subsection{The cost of learning at a given level}

We bound the interference of any update confined to levels $m,\dots,L$ and compute the interference of the local \emph{target delta rule}, which changes one level so that the upper levels map its new output onto the target.

\begin{theorem}[Cost of learning at locus $m$]\label{thm:cost}
Assume A0 and fix $m$. Let the levels below $m$ be frozen, and write $H:=h_{m-1}(X_o)$, $h:=h_{m-1}(x)$, $G:=HH\T$ and $\Delta:=f'(x)-f(x)$.
\begin{enumerate}[leftmargin=2em,itemsep=1pt]
\item[(a)] (Any learner.) If $h=0$, no update of levels $m,\dots,L$ changes $f(x)$. If $h\neq0$ and $h\in\range H$, every update of levels $m,\dots,L$ satisfies $I\ge B_m:=\norm{\Delta}^2/\big(h\T G^+h\big)$.
\item[(b)] (Attainment.) Let $e\in\range W_{L:m+1}$ and $h\neq0$. If $h\in\range H$, some update of $W_m$ alone achieves $\Delta=e$ with $I=B_m$. If $h\notin\range H$, some update of $W_m$ alone achieves $\Delta=e$ with $I=0$.
\item[(c)] (Target delta rule.) Let $e\in\range W_{L:m+1}$ and $h\neq0$. The update $\Delta W_m=W_{L:m+1}^+e\,h\T/\norm{h}^2$ achieves $\Delta=e$ and causes $I^\delta_m=\norm{e}^2\,h\T Gh/\norm{h}^4$. If moreover $h\in\range H$, then
\[
I^\delta_m=B_m\,\rho_m,\qquad 1\le\rho_m\le\frac{(\kappa_m+1)^2}{4\kappa_m},
\]
where $\kappa_m$ is the ratio of the largest to the smallest nonzero eigenvalue of $G$, and $\rho_m=1$ if and only if $h$ is an eigenvector of $G$.
\end{enumerate}
\end{theorem}

The three parts separate what any learner must pay (a), whether that price can be met (b), and what a particular local rule pays (c). If $e\notin\range W_{L:m+1}$, no update of $W_m$ alone fits the item, and the target delta rule leaves the residual $(I-W_{L:m+1}W_{L:m+1}^+)e$. If $h\notin\range H$, an interference-free update exists, but the target delta rule still interferes whenever $h$ has a component in $\range H$ (SI \cref{sec:verification}, Example~3). The bound $B_m$ is non-decreasing in $m$ in general (SI \cref{cor:monotone}). In balanced networks whose old inputs lie in the learned input subspace $\spn V$, however, each frozen lower layer acts as a full-rank linear map on $\spn V$, so $B_m$ is independent of $m$ for familiar inputs $x\in\spn X_o$ and the bound alone does not favour low loci (whereas when old inputs also carry components outside $\spn V$, $B_m$ can increase strictly with $m$). In that balanced case, locus dependence enters through the local rule's excess $\rho_m$, which is controlled by the anisotropy of old-item activity at the input to level $m$.

\begin{corollary}[Worst-case excess of the local rule]\label{cor:aniso}
Assume A0--A2 and A4. The old items' second-moment matrix at the input to level $m$ is $G_{m-1}=cR_mS^{2(m-1)/L}R_m\T$, with $\kappa_m=(s_1/s_k)^{2(m-1)/L}$. For familiar inputs ($x\in\spn V$) and reachable errors, the target delta rule attains the bound at level 1 ($\rho_1=1$), and the worst-case excess $(\kappa_m+1)^2/4\kappa_m$ over such items increases strictly with $m$ whenever $s_1>s_k$.
\end{corollary}

The corollary concerns the worst case. The excess of an individual item need not increase with the locus. For 200 familiar items the median excess rose from 1 at level 1 to 1.08, 1.30 and 1.60 at levels 2--4, below the worst case (\cref{fig:dln}c). Under these assumptions, a learner restricted to the local rule pays least by learning at the lowest level at which the error is reachable.

\subsection{Input-side novelty}

\begin{theorem}[Type B]\label{thm:typeB}
Assume A0--A2 and a type-B item ($x\perp\spn V$, $e\in\spn U$), with old inputs in $\spn V$.
\begin{enumerate}[leftmargin=2em,itemsep=1pt]
\item[(a)] (One shot.) The target delta rule at level 1, $\Delta W_1=W_{L:2}^+e\,x\T/\norm{x}^2$, gives $f'(x)=y$ and $f'(x_o)=f(x_o)$ for every old item. If $e$ lies on one mode $i$, the single gradient step with $\eta=1/(\norm{x}^2s_i^{2(L-1)/L})$ does the same.
\item[(b)] (First level only.) Gradient descent on $W_1$ alone, with the other levels frozen and any sequence of step sizes, leaves every old output unchanged; for any fixed step size $0<\eta<2/(\norm{x}^2s_1^{2(L-1)/L})$ (or any schedule $\{\eta_t\}$ satisfying $\lim_{T\to\infty}\prod_{t=0}^{T-1}(1-\eta_t\norm{x}^2s_i^{2(L-1)/L})=0$ for all $i=1,\dots,k$), it converges to an exact fit.
\item[(c)] (Joint gradient flow: reduction and conservation law.) Let $\norm{x}=1$ and $e=\gamma u_i$ with $\gamma>0$, and write $s=s_i$. Under gradient flow on all levels the weights remain in the family $W_1=R_2[S^{1/L}V\T+p\,e_ix\T]$, $W_l=R_{l+1}(S^{1/L}+\delta_lE_{ii})R_l\T$ ($l\ge2$). With $a_l:=s^{1/L}+\delta_l$, the new synapse $p$ and the old mode-$i$ synapses above it satisfy
\[
a_l(t)^2-p(t)^2=s^{2/L}\qquad\text{for all }l\ge2\text{ and }t\ge0.
\]
\item[(d)] (Exact cost of joint gradient flow.) The flow in (c) converges to $p^*=s^{1/L}q$, where $q>0$ is the unique root of $q(1+q^2)^{(L-1)/2}=g:=\gamma/s$. Every old item $x_o\in\spn V$ with mode-$i$ coefficient $c=e_i\T V\T x_o$ changes by exactly
\[
\norm{f_\infty(x_o)-f_0(x_o)}^2=c^2s^2\Big(\frac gq-1\Big)^2,
\]
and old items with $c=0$ are unchanged. For small $g$ this is $c^2(L-1)^2\gamma^4/(4s^2)+O(\gamma^6/s^4)$.
\end{enumerate}
\end{theorem}

\begin{figure}[!htbp]
\centering
\includegraphics[width=\textwidth]{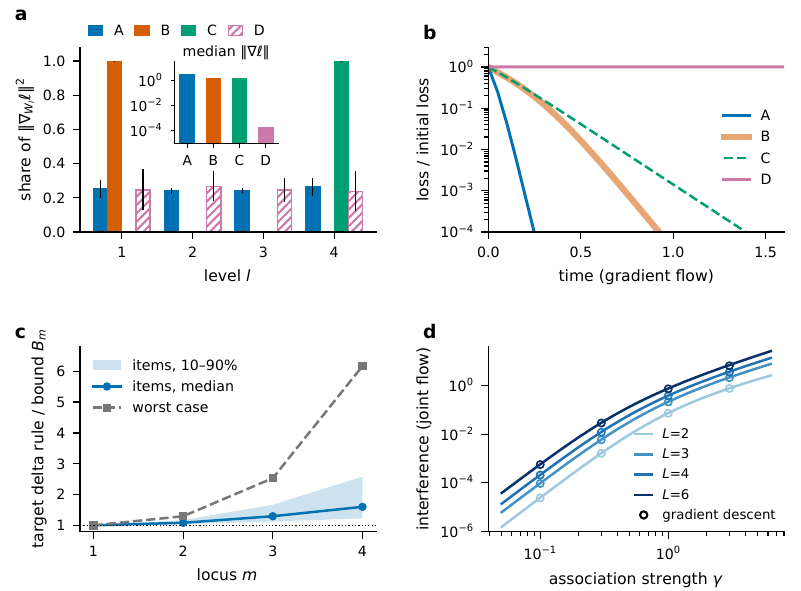}
\caption{\textbf{Window, speed, and cost in deep linear networks.} (a) Squared gradient-norm share per level across novelty types in trained networks ($L=4$; 5 seeds, 20 items/type; mean $\pm$ s.d.; inset: median $\norm{\nabla\ell}$). (b) Gradient-flow loss for unit single-mode items ($s_i=2$, aligned residual $\varepsilon=0.1$): types B and C share onset rate ($L$-fold slower than A), while D makes negligible progress over the plotted interval. (c) Target delta rule interference relative to $B_m$ over 200 familiar items ($L=4$, isotropic old inputs): median, 10--90\% range, and worst-case bound (\cref{cor:aniso}). (d) Type-B joint-flow interference versus strength $\gamma$ ($s_i=1$, $c=1$): \cref{thm:typeB}d (lines) and finite-step GD (circles) for $L\in\{2,3,4,6\}$; one-shot and level-1-only updates stayed below $10^{-30}$.}
\label{fig:dln}
\end{figure}

The conservation law has a direct reading: joint gradient flow cannot grow the new synapse at level 1 without increasing the squared strengths of the corresponding shared mode-$i$ synapses above it by the same amount, and those synapses are shared with old memories of outcome $i$. Interference is thus a property of joint all-level dynamics rather than of gradual learning as such: gradual learning confined to level 1 (b) and the one-shot update (a) are both interference-free, because the upper weights remain frozen and every rank-one level-1 update $\Delta W_1\propto\beta_1x\T$ annihilates every old input $x_o\perp x$. \Cref{fig:dln}d compares the exact endpoint of (d) with finite-step gradient descent on all levels; the relative discrepancy shrinks linearly with the step size (SI \cref{sec:verification}, Example~4). For errors spread over several modes the joint flow reduces exactly to an ordinary differential equation whose dimension does not depend on the width, and the modes couple (SI \cref{sec:multimode}).

\subsection{Interference is mode-local}

\begin{theorem}[Mode-locality]\label{thm:modes}
Assume A0--A2 and an item with $e\in\spn U$ (types A and B and their mixtures). Let $\M:=\supp(V\T x)\cup\supp(U\T e)$ be the set of touched modes. Under gradient descent on the item's loss with any sequence of step sizes, and under gradient flow, every old input $x_o\in\spn V$ whose mode support is disjoint from $\M$ satisfies $f_t(x_o)=f_0(x_o)$ for all $t$.
\end{theorem}

Onset gradients are thus confined to a band of levels, while subsequent learning remains confined to a block of modes. A new cue for outcome $i$ can disturb only memories that involve the touched modes: in simulation, other old items stayed below $4\times10^{-29}$ while mode-sharing items changed by 0.29 (\cref{fig:replay}a). This is the single-item, continual-learning counterpart of the mode decoupling of batch learning \citep{saxe2014}. The exact statement requires the balanced structure. Separately, if the balanced weights are perturbed by $\delta$, protected outputs change by $O(\delta)$ over any finite horizon, and under gradient flow that converges to an endpoint with $h_{L-1}(x)\neq0$ the bound holds uniformly over infinite time (SI \cref{prop:robust}).

\subsection{Doubly novel items}

A type-D item has an empty window: the network is at a critical point of the item's loss that is not a minimum. What happens next depends on depth.

\begin{theorem}[Type-D saddle]\label{thm:saddle}
Assume A0--A3, old inputs spanning $\spn V$, and a type-D item ($x\perp\spn V$, $y\perp\spn U$), so that $e=y$.
\begin{enumerate}[leftmargin=2em,itemsep=1pt]
\item[(a)] $\nabla\ell=0$ and $\ell=\tfrac12\norm{y}^2>0$.
\item[(b)] (Fresh path.) Choose unit vectors $q_0=\hat x$, $q_L=\hat y$ and $q_l\perp\spn R_{l+1}$ for $0<l<L$. Along $W_l\mapsto W_l+\varepsilon q_lq_{l-1}\T$ the item's output is $\varepsilon^L\norm{x}\hat y$, and every old output is unchanged for every $\varepsilon$.
\item[(c)] (Negative-curvature directions interfere if $L\ge3$.) The second-order term in the Taylor expansion of $\ell(W+\delta W)-\ell(W)$ along $\{\delta W_l\}$ is $\tfrac12\delta^2\ell=\tfrac12\norm{W_{L:2}\delta W_1x}^2-e\T\delta W_LW_{L-1:2}\delta W_1x$. For $L\ge3$, every direction of negative curvature changes at least one old output along $\spn U^\perp$ at first order, and every direction that leaves old outputs unchanged to first order has nonnegative curvature. For $L=2$ there are directions of negative curvature that leave old outputs unchanged.
\item[(d)] (Aligned gradient flow is slow.) If the residual weights are aligned with the fresh path, $W_l=W_l^{\mathrm{old}}+a_lq_lq_{l-1}\T$ with $a_l(0)=\varepsilon$ and $0<\varepsilon<(\sigma/(2c))^{1/L}$, gradient flow stays on the path and follows $\dot a=c\,a^{L-1}(\sigma-c\,a^L)$ with $c=\norm{x}$ and $\sigma=\norm{y}$. The time until the output along $\hat y$ reaches $\sigma/2$ is $\tfrac{1}{2c\sigma}\ln\tfrac{\sigma/c-\varepsilon^2}{\varepsilon^2}$ for $L=2$, and $\tfrac{\varepsilon^{-(L-2)}}{(L-2)c\sigma}+O(1)$ for $L\ge3$.
\end{enumerate}
\end{theorem}

\begin{figure}[!htbp]
\centering
\includegraphics[width=0.9\textwidth]{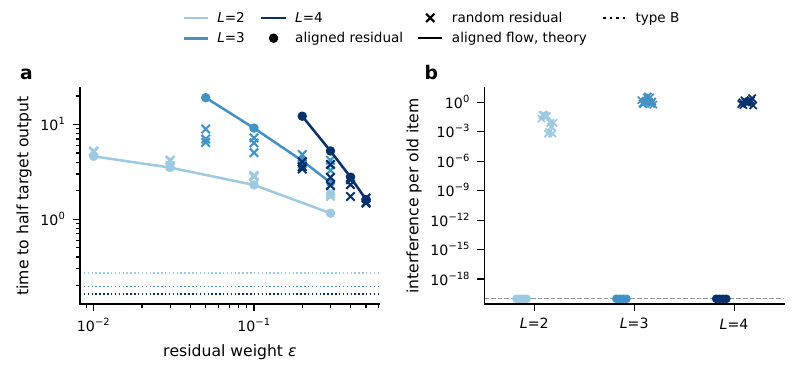}
\caption{\textbf{Gradient dynamics at the type-D saddle (\cref{thm:saddle}).} (a) Time until a type-D item reaches half target output ($\sigma/2$, corresponding along the aligned path to one-quarter of the unperturbed saddle loss) against residual weight $\varepsilon$ ($L=2,3,4$). Circles: aligned fresh-path residual; lines: \cref{thm:saddle}d; crosses: random residuals (3 draws); dotted lines: type-B half-output time. (b) Mean old-item interference at half target output (rounding-level values below $10^{-20}$ drawn at the floor).}
\label{fig:saddle}
\end{figure}

For $L\ge3$, gradient-based escape from the saddle along directions of negative curvature necessarily perturbs at least one old output at first order by reusing channels that old items depend on, whereas gradient flow along an aligned interference-free path started from a small residual weight is slow; for type B, the learning time does not depend on $\varepsilon$. Depth two is exempt. The theorem does not say that no fast interference-free update exists: the coordinated finite update $W_l+a\,q_lq_{l-1}\T$ with $a=(\norm{y}/\norm{x})^{1/L}$ fits the item exactly without interference (SI \cref{sec:verification}, Example~5). The obstacle is that local gradient dynamics near the saddle do not find it quickly. \Cref{fig:saddle} shows both behaviours: aligned residual weights matched (d) to three significant digits with rounding-level interference, whereas random residual weights of the same scale at $L\ge3$ perturbed old outputs by 0.5--4 per item while escaping earlier at small $\varepsilon$ (at $L=3$, $\varepsilon=0.05$: 6.5--8.9 versus 19.2 time units; at $L=2$, old-output change was one to three orders of magnitude smaller).

These dynamics give a mechanistic rationale, consistent with complementary learning systems theory \citep{mcclelland1995,kumaran2016}, for storing doubly novel items in a fast auxiliary episodic store that teaches the hierarchy gradually with interleaved replay: under the gradient dynamics analysed here, items with a familiar outcome (types A and B) can be integrated directly without that detour, whereas doubly novel items exhibit a speed--interference trade-off at depth three or more. If the hierarchy has a saturated bottleneck ($r=\operatorname{rank}F_o$) and the new output is orthogonal to old outputs, interference is bounded below by the energy of the weakest old mode (SI \cref{thm:capacity}).

\section{Precision selects the locus; which partial replay is safe}
\label{sec:mechanism}

\Cref{sec:dynamics} characterises what learning at each level costs. A hierarchy with local plasticity would need a way to place learning at a chosen level. In predictive coding, precision provides one. We also derive when partial replay protects untouched memories.

\subsection{Predictive coding: a precision contrast selects the plastic level}
\label{sec:pc}

Consider linear supervised predictive coding with level precisions $\pi_l>0$ \citep{rao1999,friston2005,whittington2017,bogacz2017}:
\begin{equation}\label{eq:pc}
E(u)=\tfrac12\sum_{l=1}^L\pi_l\norm{u_l-W_lu_{l-1}}^2,\qquad u_0=x,\ u_L=y,\qquad \varepsilon_l:=u_l-W_lu_{l-1}.
\end{equation}
Inference minimises $E$ over the hidden activities $u_1,\dots,u_{L-1}$, a strictly convex quadratic problem. Plasticity then acts either on the \emph{raw} equilibrium errors, $\Delta W_l=\eta\,\varepsilon_lu_{l-1}\T$, or on the \emph{precision-weighted} errors, $\Delta W_l=\eta\,\pi_l\varepsilon_lu_{l-1}\T$.

\begin{theorem}[Precision selects the locus]\label{thm:pc}
Assume A0.
\begin{enumerate}[leftmargin=2em,itemsep=1pt]
\item[(a)] (Equilibrium.) With $\xi_l:=\pi_l\varepsilon_l$, the equilibrium satisfies $\xi_l=W_{L:l+1}\T\xi_L$ and $\xi_L=K_\pi^{-1}e$, where $K_\pi:=\sum_l\pi_l^{-1}K_l\succ0$.
\item[(b)] (Raw-error plasticity localises.) Let $\pi_m=1$ and $\pi_l=\Pi$ for $l\neq m$, let $e\in\range W_{L:m+1}$, and let $h:=h_{m-1}(x)\neq0$. As $\Pi\to\infty$,
\[
\varepsilon_m=W_{L:m+1}^+e+O(\Pi^{-1}),\quad \varepsilon_{l}=O(\Pi^{-1})\ (l\neq m),\quad u_{m-1}=h+O(\Pi^{-1}).
\]
Hence the raw-error update at level $m$ is, up to $O(\Pi^{-1})$, proportional to the target delta rule of \cref{thm:cost}c, and equal to it for $\eta=1/\norm{h}^2$ (or $\eta=1/\norm{u_{m-1}}^2$, which has the same limit). The updates of the other levels are $O(\Pi^{-1})$. For a type-B item with old inputs $x_o\in\spn V$ (or $x_o\perp x$), $m=1$, and $\eta=1/\norm{x}^2$, the interference of one update is $O(\Pi^{-2})$.
\item[(c)] (Precision-weighted plasticity does not guarantee localisation.) Precision-weighted error signals $\xi_l=W_{L:l+1}\T\xi_L$ do not contain the explicit inverse-precision ($\pi_l^{-1}$) attenuation of raw-error signals. Consequently, precision contrast alone does not guarantee localisation of precision-weighted weight updates $\Delta W_l=\eta\xi_lu_{l-1}\T$ to the low-precision level; whenever both limiting factors $\xi_l$ and $u_{l-1}$ remain nonzero as $\Pi\to\infty$, the update remains nonvanishing at level $l$.
\item[(d)] (Standard predictive coding is not interference-free for type B.) For a type-B item with old inputs $x_o\in\spn V$, all $\pi_l=1$, and simultaneous updates of all levels, old outputs change by
\[
\eta\sum_{l\geq2}K_l\xi_L\,\ip{u_{l-1}}{h_{l-1}(x_o)}+O(\eta^2),
\]
which is generally nonzero because the equilibrium activity $u_{l-1}=\sum_{j<l}W_{l-1:j+1}\varepsilon_j$ carries the error upward, whereas one gradient step causes no interference for type B (\cref{cor:speed}).
\end{enumerate}
\end{theorem}

The level with the lowest precision absorbs the equilibrium error and becomes the plastic level, provided reachability ($e\in\range W_{L:m+1}$) and nonzero presynaptic support ($h_{m-1}(x)\neq0$) hold and plasticity reads raw rather than precision-weighted errors. Because $\Delta W_l^{\mathrm{raw}}=\pi_l^{-1}\Delta W_l^{\mathrm{weighted}}$, raw-error plasticity under a precision contrast acts like layerwise learning-rate preconditioning ($\eta_l=\eta/\pi_l$) coupled with equilibrium activity inference: the precision contrast suppresses off-locus updates as $\Pi^{-1}$, while activity inference transforms the error at level $m$ into the pseudoinverse target $W_{L:m+1}^+e$ rather than the transpose gradient $W_{L:m+1}\T e$. The limiting update is therefore the target delta rule, attaining the lower bound of \cref{thm:cost}a only when $\rho_m=1$. The $O(\eta^2)$ term in (d) is nonzero in finite steps: in a three-level network with equal precisions, the old output changes by $\eta/3+2\eta^2/81$, matching the first-order term $\eta/3$ as $\eta\to0$ (SI \cref{sec:verification}, Example~6). With a precision contrast, global minimisers of the nonlinear predictive-coding energy likewise approach a nonlinear target-propagation step at level $m$ under reachability and boundedness conditions (SI \cref{prop:nonlinearpc}).

\Cref{fig:precision} compares learners at a matched fit (relative misfit $10^{-3}$) in linear networks. For a type-B item with locus 1, interference under raw-error predictive coding falls as $\Pi^{-2}$, from 5.3 with equal precisions to $1.1\times10^{-6}$ at $\Pi=10^4$, where a single update meets the criterion. All-layer backpropagation caused 2.45 at the same fit, whereas level-1-only gradient descent and the target delta rule caused rounding-level interference. For a familiar (type-A) item, every learner paid a similar price at matched fit (2.2--2.5), consistent with \cref{thm:cost}a.

\paragraph{Nonlinear predictive coding}
To test whether precision-selected localisation survives nonlinearity and iterative activity inference, we evaluated four-layer bias-free $\tanh$ and ReLU predictive-coding networks on a structured nonlinear task ($n=10$ seeds, relative misfit $\le10^{-2}$) and three-layer $\tanh$ networks on an MNIST familiar-output adaptation benchmark ($n=10$ seeds, pre-trained on digits 0--4 to 99.2\% test accuracy, adapting to digits 5--9 mapped onto old labels 0--4 to relative misfit $\le0.05$; SI \cref{sec:pcnonlinmethods}). Because not every baseline reached the stopping criterion on every seed in the ReLU and raw-digit MNIST conditions, we distinguish \emph{all-seed endpoint comparisons} ($n=10$ seeds under the prescribed stopping budget) from \emph{strictly matched-fit paired comparisons} restricted to seeds where both compared learners met the criterion, reporting exact two-sided paired Wilcoxon signed-rank $p$-values (and Holm--Bonferroni adjustment for $m=3$ planned baseline comparisons where all 10 seeds reached criterion).

In four-layer $\tanh$ networks on type-B items (\cref{fig:generalization}c, \cref{tab:quant}), every learner reached the $10^{-2}$ criterion in 10 of 10 seeds. Equal-precision predictive coding ($\Pi=1$) again interfered more than all-layer backpropagation (median $5.88\times10^{-2}$ versus $3.08\times10^{-2}$), whereas layer-1-only backpropagation achieved $5.55\times10^{-14}$. Increasing precision contrast with raw-error updates drove interference down as $\Pi^{-1.87}$ to $5.00\times10^{-7}$ at $\Pi=10^3$ (median layer-1 weight-change share $0.999998$; strictly matched-fit $p=0.0020$, Holm-adjusted $p=0.0060$, against all-layer backpropagation, equal-precision predictive coding, and precision-weighted predictive coding), whereas precision-weighted updates at $\Pi=10^3$ failed to localise (layer-1 share 0.38, median interference $3.48\times10^{-1}$), consistent with \cref{thm:pc}c. On familiar (type-A) items all learners incurred comparable interference (0.54--0.68). Four-layer ReLU networks serve as a supplementary sensitivity test under an approximately projected, numerically perturbed configuration: exact subspace projection $W_1\leftarrow W_1P_V$ in float32 leaves rounding residuals ($\sim10^{-7}$) along $V^\perp$ that yield small nonzero initial pre-activations for roughly half the hidden units (whereas exact zero pre-activation can stall feedforward-initialised ReLU inference as well as backpropagation; SI \cref{prop:nonlinearpc}). Under this perturbed configuration, layer-1-only backpropagation reached criterion in 0 of 10 seeds within 12{,}000 steps (median relative misfit $0.108$, interference $2.71\times10^{-12}$), all-layer backpropagation reached criterion in 9 of 10 seeds (median interference 1.13), raw-error predictive coding ($\Pi=10^3$) reached criterion in 10 of 10 seeds in a median of 5 updates (layer-1 share 0.96, median interference 0.25 versus 2.01 for equal-precision predictive coding, 10/10 reached), and precision-weighted predictive coding ($\Pi=10^3$) reached criterion in 4 of 10 seeds (median interference 9.08). All-seed endpoint comparisons gave $p=0.0020$ ($n=10$) against each baseline, while strictly matched-fit paired subsets yielded $p=0.0020$ versus equal-precision predictive coding ($n=10$), $p=0.0039$ versus all-layer backpropagation ($n=9$, medians 0.235 versus 1.003), and $p=0.125$ versus precision-weighted predictive coding ($n=4$, medians 0.115 versus 8.02, the minimum attainable two-sided $p$ for $n=4$).

On MNIST familiar-output adaptation (\cref{fig:generalization}d), raw-error contrast at $\Pi=500$ concentrated $>99.99\%$ of squared weight change in layer 1 (median share $0.99995$--$0.99997$; 10/10 seeds reached the 5\% criterion, achieved median relative misfit $0.049$). On raw unseen digits 5--9 mapped to old labels, raw-error contrast cut median interference 24-fold (0.023 versus 0.57 for all-layer backpropagation and 0.54 for equal-precision predictive coding; matched-fit $p=0.0020$, $n=10$) and reduced old-digit accuracy drop from 23.8 pp (max 61.7 pp) under backpropagation and 22.3 pp (max 55.3 pp) under equal-precision predictive coding to 0.45 pp (IQR 0.13--0.70, max 0.80 pp; $p=0.0020$, $n=10$). Against precision-weighted predictive coding ($\Pi=500$, 7/10 reached, median interference 0.273, accuracy drop 1.9 pp), all-seed endpoint comparisons gave $p=0.0039$ (interference) and $p=0.0195$ (accuracy drop; $n=10$), and strictly matched-fit paired runs gave $p=0.0156$ for both metrics ($n=7$, median interference 0.023 versus 0.315). On pixel-orthogonalised shifts, every learner reached criterion in 10 of 10 seeds: raw-error contrast reduced median interference roughly $10^3$-fold ($6.50\times10^{-4}$ versus 0.69 for backpropagation and 0.64 for equal-precision and precision-weighted predictive coding; matched-fit $p=0.0020$, $n=10$) and lowered old-digit accuracy drop from 31.4 pp (IQR 22.3--35.3, max 37.0 pp) under backpropagation and 24.9 pp (max 36.1 pp) under equal-precision predictive coding to 0.0 pp (IQR 0.0--0.0, max 0.0 pp; precision-weighted $\Pi=500$: median 3.6 pp, max 30.1 pp; $p=0.0020$, $n=10$). Across 200 class-balanced held-out orthogonalised items (5-class chance 20\%, pre-adaptation baseline 9.3\%), however, held-out accuracy remained below chance for all localised learners (12.0\% for raw-error PC and layer-1 BP, 12.5\% for LR-scaled BP, versus 20.3\% for all-layer BP); thus, zero old-task loss in the orthogonalised condition demonstrates protected acquisition of specific new associations rather than generalisable classification. Controls with layer-1-only and learning-rate-scaled ($\eta_l=\eta/500$) backpropagation matched old-task retention (0.00 pp orthogonal; 0.70 and 0.65 pp raw unseen), confirming that localisation itself drives retention. On raw unseen digits, activity inference improved \emph{weight-update efficiency} 5.3--5.8-fold (median 50.5 updates versus 291.5 for layer-1 and 265.5 for LR-scaled BP), but incurred an 84--92-fold CPU wall-clock overhead (12.4 s versus 0.15 s and 0.13 s, including all L-BFGS inference iterations, line searches, and stopping checks); localising adaptation to layer 1 also traded held-out new-digit generalisation (43.0\% for raw-error PC and 48.0\% for layer-1/LR-scaled BP versus 59.0\% for all-layer BP, above a 38.3\% baseline) for substantial retention gains (SI \cref{tab:mnistcontrols}).

\subsection{Learned precision}

In predictive coding, precisions are estimated by minimising variational free energy, $F=\tfrac12\sum_l\pi_l\norm{\varepsilon_l}^2-\tfrac12\sum_ln_l\log\pi_l$, whose stationarity condition is $\pi_l=n_l/\norm{\varepsilon_l}^2$ \citep{friston2005,feldman2010,bogacz2017}. While equilibrium activity inference depends only on precision ratios and is invariant under common rescaling of $\{\pi_l\}$, the unregularised free-energy objective $F$ is not scale-invariant, and vanishing prediction errors ($\varepsilon_l=0$) make unregularised precision estimates unbounded. We therefore track the normalised energy shares $\sigma_l:=\pi_l\norm{\varepsilon_l}^2/\sum_j\pi_j\norm{\varepsilon_j}^2$.

\begin{theorem}[Learned precision]\label{thm:learnedprec}
Alternate the equilibrium of \cref{thm:pc}a with the update $\pi_l\leftarrow n_l/\norm{\varepsilon_l}^2$ on active levels ($\varepsilon_l\neq0$; levels with $\varepsilon_l=0$ have $\sigma_l=0$ and are excluded from the ratio update).
\begin{enumerate}[leftmargin=2em,itemsep=1pt]
\item[(a)] (Share recursion, common-eigenvector error case.) If $K_le=k_le$ for every $l$ (for example, a single-mode error in a balanced network, $k_l=s^{2(L-l)/L}$), then for all active levels ($k_l>0$),
\[
\sigma_l'=\frac{\sigma_l^2/n_l}{\sum_j\sigma_j^2/n_j},\qquad\text{so}\qquad \sigma^{(t)}_l\propto n_l\big(\sigma_l^{(0)}/n_l\big)^{2^t}.
\]
The normalised shares $\sigma_l^{(t)}$ converge doubly exponentially to the set of active levels maximising $\sigma^{(0)}_l/n_l$, divided among them in proportion to $n_l$. If the maximiser $j^*$ is unique, all precision ratios $\pi_l/\pi_{j^*}$ for non-maximising levels with $k_l>0$ diverge and $j^*$ becomes the dominant error-energy level. From uniform initial precisions with equal widths, the bottom level is the unique maximiser if $s>1$ and the top level if $s<1$; for $s=1$ the shares remain equal.
\item[(b)] (Reachability.) In general $\varepsilon_l=\pi_l^{-1}W_{L:l+1}\T K_\pi^{-1}e$. Under A0--A2, a pure type-C or type-D item has $\varepsilon_l=0$ for $l<L$, so only the top level can be selected. Without the balanced structure, $K_le=0$ does not imply $\varepsilon_l=0$.
\item[(c)] (Presynaptic-support limitation.) Under the uniform-initial-precision, equal-width, single-mode conditions of (a), a type-B item with $s<1$ selects the top level as the dominant error-energy level, but its presynaptic activity $u_{L-1}$ vanishes as the lower-level precision ratios diverge, so the limiting raw-error update vanishes, preventing learning in this limit.
\end{enumerate}
\end{theorem}

\begin{figure}[!htbp]
\centering
\includegraphics[width=\textwidth]{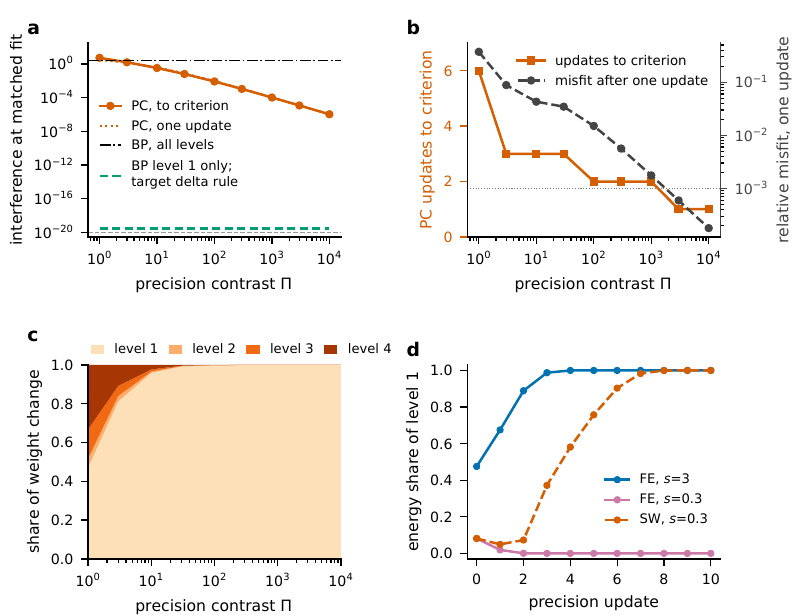}
\caption{\textbf{Precision selects the locus (\cref{thm:pc,thm:learnedprec}).} Balanced network ($L=4$, $s=(3,2,1,0.6,0.3)$, 20 old items), raw-error updates with $\eta=1/\norm{u_{m-1}}^2$, locus 1 precision 1 and other levels $\Pi$. (a) Matched-fit interference (relative misfit $10^{-3}$) versus $\Pi$ for a type-B item: repeated predictive-coding updates (solid), single update (dotted), all-layer backpropagation (BP), and level-1-only GD / target delta rule (below $10^{-28}$, shown above the $10^{-20}$ floor). (b) Updates to criterion (left axis) and single-update relative misfit (right axis) for type B. (c) Weight-change share per level after one update. (d) Level-1 energy share over precision updates for single-mode type B: free-energy (FE) precision selects level 1 for $s=3$ but stalls at level $L$ for $s=0.3$, whereas support-weighted (SW) precision selects level 1.}
\label{fig:precision}
\end{figure}

In (b), a non-balanced three-level network with $K_1e=0$ has $\varepsilon_1=(-1/7,0)\T$ at equilibrium (SI \cref{sec:verification}, Example~8). The limitation in (c) arises because error energy does not register whether a level has the presynaptic activity needed to change its weights. We therefore propose a \emph{support-weighted} heuristic, $\pi_l\leftarrow n_l/(\norm{\varepsilon_l}^2\norm{u_{l-1}}^2)$, which weights each level by the energy of its own raw-error update rather than deriving from a generative model. In the example of (c) it selected level 1, after which one update fitted the item to misfit $6\times10^{-11}$ (\cref{fig:precision}d); for a familiar item, however, it selected the top level whereas \cref{cor:aniso} favours the bottom under A4. Beyond the common-eigenvector case of (a), with errors spread over several modes, the free-energy update selected the level with the largest initial energy share in 60 of 60 random balanced networks and items (an empirical observation). A Gamma hyperprior bounds absolute precisions and makes locus selection graded rather than winner-take-all (SI \cref{sec:precision}).

\subsection{Which partial replay is safe}

Replaying old items while learning (rehearsal) trades interference against misfit. With full rehearsal of all old items and penalty $\lambda>0$, achievable interference and misfit lie on an exact frontier that no linear-network update can beat (SI \cref{prop:frontier}); if a familiar input's new target conflicts with its old output, rehearsal cannot eliminate interference (SI \cref{sec:verification}, Example~9). For a type-B item, full rehearsal removes interference at every global minimiser. Rehearsal in biological and artificial systems is typically partial, however, and partial replay can damage memories that \cref{thm:modes} would otherwise protect.

\begin{theorem}[Partial replay]\label{thm:replay}
Assume A0--A2 and an item with $e\in\spn U$ and touched modes $\M$. Rehearse a set $R$ of old items $x_r\in\spn V$ with mode vectors $c_r=V\T x_r$ by gradient descent (any step sizes) or gradient flow on $\ell+\tfrac\lambda2\sum_{r\in R}\norm{f(x_r)-f_0(x_r)}^2$ with $\lambda\ge0$, and define the uncentred cross-moment $C_R:=\sum_{r\in R}c_{r,\M}\,c_{r,\M^c}\T$.
\begin{enumerate}[leftmargin=2em,itemsep=1pt]
\item[(a)] (Sufficiency.) If $C_R=0$, every old item supported on $\M^c$, rehearsed or not, is unchanged for all time.
\item[(b)] (Leakage rate.) Assume $\lambda>0$ and let the weights lie in the invariant family of \cref{thm:modes}, with touched-block factors $A_l$ and touched-block change $\Delta:=A_{L:1}-S_\M$. The output of a probe $x_o=Vc$ with $c$ supported on $\M^c$ then changes at the rate
\[
\dot f(x_o)=-\lambda\,U_\M\sum_{l=1}^LA_{L:l+1}A_{L:l+1}\T\,\Delta\,C_R\,S_{\M^c}^{2(l-1)/L}\,c .
\]
For a single touched mode the rate is $-\lambda\,U_\M\,\Delta\,C_R\Lambda c$ with the positive diagonal matrix $\Lambda:=\sum_lA_{L:l+1}^2S_{\M^c}^{2(l-1)/L}$, which is nonzero if and only if $\Delta\neq0$ and $C_R\Lambda c\neq0$: once the touched block has changed, every unrehearsed old item with $C_R\Lambda c\neq0$ starts to change.
\item[(c)] (Endpoint, type B.) Assume $\lambda>0$ and let the item be of type B ($x\perp\spn V$). Every global minimiser of the objective within the invariant family has zero loss, and its touched-block change annihilates $\spn\{c_{r,\M}\}$. If these vectors span $\R^{|\M|}$, every old item is unchanged at such a minimiser.
\end{enumerate}
\end{theorem}

\begin{figure}[!htbp]
\centering
\includegraphics[width=0.9\textwidth]{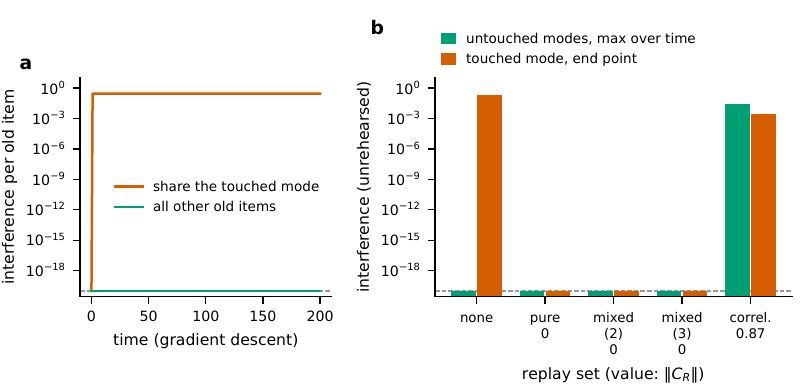}
\caption{\textbf{Mode-locality and partial replay (\cref{thm:modes,thm:replay}).} (a) Gradient descent on a type-B item touching mode 1 ($L=4$): interference on mode-sharing versus untouched old items (rounding-level floor $10^{-20}$). (b) Joint learning with partial replay ($L=3$, $\lambda=0.5$): peak trajectory interference on unrehearsed untouched-mode items (green) and final interference on unrehearsed touched-mode items (orange), with $\norm{C_R}$ below each replay set. Decorrelated sets ($C_R=0$) spanning the touched mode protect both groups, whereas a correlated set damages previously spared memories.}
\label{fig:replay}
\end{figure}

Preservation of untouched modes thus depends on the cross-moment structure of the replay set: replayed items may individually mix touched and untouched dimensions provided the mixing cancels in $C_R$ (\cref{fig:replay}b). A correlated replay set ($\norm{C_R}=0.87$) damaged unrehearsed memories on untouched modes that were spared without replay ($3\times10^{-2}$ versus $5\times10^{-29}$), whereas sets with $C_R=0$ kept them at rounding level throughout learning (SI \cref{sec:verification}, Example~13).

\section{Convolutional and residual networks trained on natural images}
\label{sec:cnn}

We tested selected predictions of the theory beyond the linear setting in two deep visual architectures trained on CIFAR-10 \citep{krizhevsky2009}: a six-layer convolutional network trained with squared loss, and a standard ResNet-18 \citep{he2016} with skip connections, batch normalisation, biases, and cross-entropy loss. Throughout this section, the labels A--D denote \emph{operational adaptation conditions} designed to probe input-side and output-side novelty in natural-image networks rather than exact realisations of the pure linear subspaces in \cref{def:types}. Across both architectures we asked whether output-side novelty confines onset plasticity to the readout, how input-side novelty behaves under natural and controlled pixel-level shifts, and whether the pre-learning layer-resolved tangent kernel predicts stage-wise interference and optimal locus selection.

\subsection{Six-layer convolutional networks with squared loss}

\paragraph{Experimental protocol}
A bias-free six-layer ReLU convolutional network (four convolutional layers, global average pooling, one hidden fully connected layer, and a linear readout) was pre-trained with squared loss on five CIFAR-10 classes \citep{hui2021} to 81.4\% test accuracy (mean over 5 independent seeds). As a controlled test of pure output-side novelty, five additional output units were reserved with incoming readout weights initialised and maintained at zero during pre-training; robustness to nonzero readout initialisation is evaluated separately via a residual-weight sweep below. Sets of 20 new items were adapted to 5\% of initial loss across six conditions (3 draws per seed, 90 conditions total): \textbf{A} (held-out old-class images remapped to another old unit), \textbf{B} (images of unseen classes mapped to old units), \textbf{B$'$} (colour-negative old-class images with original labels), \textbf{C} (old images acquiring a response on a reserved unit while retaining old outputs), \textbf{D} (unseen-class images acquiring a response on a reserved unit while retaining old outputs), and \textbf{D$^*$} (standard one-hot class-incremental targets on a reserved unit, zero on old units).

Each condition was adapted both with all layers plastic (at a seed-common step size) and with one layer plastic at a time. Interference is the mean squared output change over 1{,}000 held-out old-class images, paired with old-class accuracy drop (percentage points, pp) and old-task squared error. Before any parameter update, we computed a \emph{prospective prediction} for each single-layer run by integrating linearised gradient descent under that layer's tangent kernel (\cref{thm:kernel}) to the 5\% criterion and propagating the resulting parameter update to the retention images via Jacobian--vector products (SI \cref{sec:cnnmethods}). Out of 540 candidate single-layer configurations (90 conditions $\times$ 6 layers), 150 lay below the readout for C and D where zero readout weights yield identically zero lower-layer gradients, and 75 lay below the readout for D$^*$ where 49--68\% of initial error rests on reserved outputs (reported separately as partial fits). Of the 315 reachable single-layer runs, 305 reached the 5\% adaptation criterion within 4{,}000 steps and 10 (at conv1--conv2) did not; all 90 all-layer runs reached criterion. Prospective kernel evaluations use the 305 reached runs; note that cluster bootstrap confidence intervals reflect variation across the 5 independent pre-trained seeds.

\begin{figure}[!htbp]
\centering
\includegraphics[width=\textwidth]{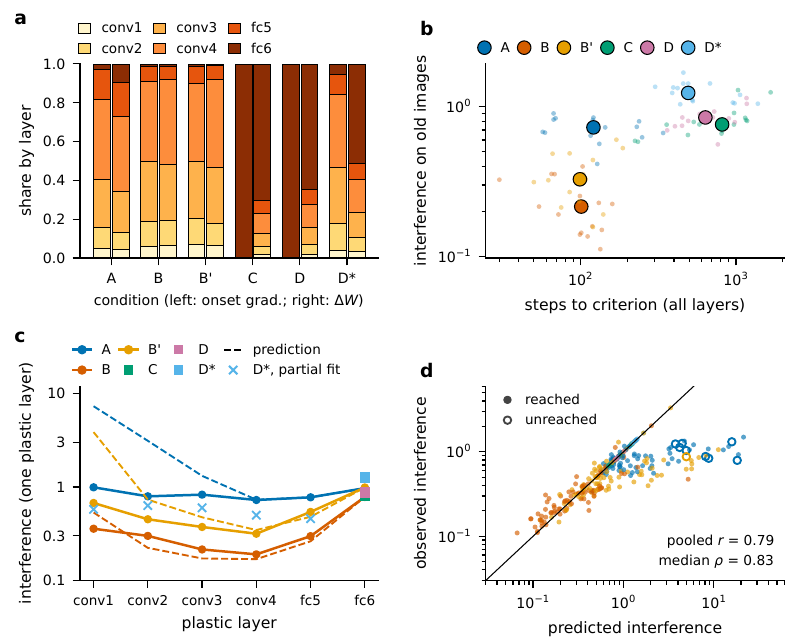}
\caption{\textbf{Six-layer CIFAR-10 convolutional networks} (5 seeds $\times$ 3 draws; squared loss). (a) Onset squared gradient share (left) and total squared weight change after all-layer adaptation (right; means over 15 draws). (b) Steps to the 5\% criterion against retention interference (small dots: runs; large: medians). (c) Single-layer interference medians for reached runs (solid) and prospective layer-kernel predictions (dashed); squares: C, D, D$^*$ at readout; crosses: partial fits of D$^*$ below readout (median 62\% residual loss). (d) Prospective prediction versus observation for 305 reached runs (filled) and 10 unreached runs (open, excluded from correlations).}
\label{fig:cnn}
\end{figure}

\paragraph{Output-side novelty}
For C and D, 100\% of the onset squared gradient norm lay in the readout (\cref{fig:cnn}a), as \cref{prop:relu} requires when error resides on zero-weight output rows. Although growing readout weights opened the plasticity window downward during adaptation, 70\% (C) and 65\% (D) of total squared weight change remained in the readout. Because only one layer contributed at onset, C and D required medians of 815 and 636 steps versus 99--121 for A, B, and B$'$ (\cref{fig:cnn}b), consistent with 27--50-fold lower tangent-kernel onset rates (\cref{cor:speed}). Introducing Gaussian residual weights of relative scale $\varepsilon$ into the reserved readout rows caused the onset gradient share below the readout to scale as $\varepsilon^2$ over three decades ($5\times10^{-6}$, $5\times10^{-4}$, and $0.05$ at $\varepsilon=10^{-3},10^{-2},10^{-1}$), matching SI \cref{prop:soft}, while adaptation sped up only once $\varepsilon\gtrsim0.3$ (SI \cref{fig:cnncontrols}a,b). In class-incremental targets (D$^*$), 32--51\% of initial error energy lay on old units suppressible by every layer, spreading gradients across depth and producing the highest interference (median 1.24).

\paragraph{Natural-image novelty versus controlled pixel-space shifts}
Unseen natural classes (B) and colour-negative images (B$'$) did not constitute input-side novelty in the sense of \cref{def:types}: the fraction $\nu_l$ of their activity outside the 95\%-energy subspace of 2{,}000 old training images differed from held-out old images by at most 0.033 at every layer (SI \cref{tab:nu}). Accordingly, onset gradients peaked at conv3--conv4 rather than conv1, while causing less interference than A (medians 0.22 and 0.33 versus 0.73) and smaller accuracy drops (14 and 23 pp versus 46 pp). Output displacement and classification loss also dissociated: C and D changed old outputs on reserved units as strongly as A (medians 0.76 and 0.85) while lowering old-class accuracy by only 9 and 7 pp.

To test whether stronger input-subspace novelty moves behaviour toward the linear prediction, we evaluated three controlled pixel-space shifts mapped onto familiar labels across the same five pre-trained networks (3 draws per seed): spatially permuted pixels ($\mathrm{B}_{\mathrm{perm}}$, median $\nu_0=0.62$ against the 500-image retention subspace; $0.59$ against 2{,}000 training images), strong Gaussian pixel noise ($\mathrm{B}_{\mathrm{noise}}$, $\sigma=2.0$, $\nu_0=0.78$ / $0.76$), and synthetic patterns projected orthogonal to the 95\%-energy pixel subspace of 2{,}000 old training images ($\mathrm{B}_{\mathrm{orth}}$, $\nu_0=0.992$ against the 500-image retention subspace; $1.000$ against the 2{,}000-image training subspace; SI \cref{sec:cnnmethods}; on $\mathrm{B}_{\mathrm{nat}}$, the 500-image retention median is $0.103$ versus $0.069$ median and $0.071$ mean against the 2{,}000-image reference in SI \cref{tab:nu}). Raising $\nu_0$ from natural unseen classes to $\mathrm{B}_{\mathrm{orth}}$ cut median optimal-layer interference nearly ninefold (conv4: 0.21 to 0.024 bias-free; 0.21 to 0.079 biased; \cref{tab:quant}). Yet conv4 still outperformed conv1 even on $\mathrm{B}_{\mathrm{orth}}$ (0.024 versus 0.34). Direct measurement of $3\times3$ local patches supports an architectural interpretation: 95\% of old-image $3\times3$ patch energy occupies just 4 of 27 patch dimensions, so even when full-image cosine overlap is negligible ($2.9\times10^{-5}$ for $\mathrm{B}_{\mathrm{orth}}$), 11.0\% of local $3\times3$ patch energy still lies inside the old patch subspace (patch novelty $0.890$, mean squared patch cosine overlap $0.0225$ versus $0.4085$ for $\mathrm{B}_{\mathrm{nat}}$), and representation overlap increases immediately after the first convolutional layer (rising from $2.9\times10^{-5}$ at input to $0.071$ after conv1 and $0.126$ after conv2; SI \cref{sec:resnetmethods}).

\paragraph{Layer-resolved kernel prediction and sequential adaptation}
Single-layer interference varied strongly with depth (\cref{fig:cnn}c): for A, B, and B$'$, conv4 was least interfering in 33 of 45 conditions and the readout most interfering. Before any weight update, the prospective layer-resolved kernel captured this profile with median per-condition Spearman $\rho=0.83$ (IQR 0.60--0.90) and pooled log-interference correlation $r=0.79$ across the 305 reached runs (95\% 5-seed cluster bootstrap CI 0.73--0.82; per seed 0.69--0.84; \cref{fig:cnn}d). Excluding the linear readout gave $r=0.83$ ($n=215$) and median $\rho=0.80$. Calibration was tightest from conv4 upward (median absolute $\log_{10}$ error 0.05 at conv4 versus 0.54 at conv1, where long trajectories left the local linear regime) and remained high across all pixel-space shifts (median $\rho=0.94$ for $\mathrm{B}_{\mathrm{perm}}$, $\mathrm{B}_{\mathrm{noise}}$, and $\mathrm{B}_{\mathrm{orth}}$). Biased networks preserved these patterns (100\% readout onset for C and D, 6.4--9-fold slower adaptation, conv4 optimal in 29/45 conditions, prospective $r=0.73$ and $\rho=0.66$; SI \cref{fig:cnncontrols}).

To assess original-task retention beyond isolated single-event probes, we evaluated final displacement $I_T:=\tfrac{1}{P}\sum_o\norm{f_T(x_o)-f_0(x_o)}^2$ and per-event criterion completion after 20-event sequential adaptation streams (SI \cref{tab:seqdetail}). In the pre-trained six-layer CNNs (5 seeds, 100 familiar-output events evaluated on a strictly disjoint 500-image retention set), restricting all 20 events to intermediate stage conv4 achieved both lower final forgetting and a higher event-completion rate (89/100 events reaching the 10\% criterion; median $I_T=0.148$ [IQR 0.145--0.226], accuracy drop 10.0 pp [9.4--20.4]) than all-layer backpropagation (83/100 reached; 0.204 [0.196--0.205] / 14.2 pp [11.0--15.2]) and myopic single-event NTK selection over \texttt{conv2}--\texttt{fc6} (76/100 reached; 0.285 [0.170--0.330] / 21.8 pp [12.2--26.4]), while fixed readout adaptation reached 94/100 events at the highest forgetting (1.019 [0.983--1.054] / 42.4 pp [28.8--42.6]). Thus conv4's lower forgetting did not stem from under-fitting, whereas myopic single-event linearised selection accumulated nonlinear drift across repeated events. In initially balanced four-layer linear networks (10 seeds, 200 mixed B/C/A events), prospective single-layer target-delta selection also showed lower final displacement than iterative all-layer backpropagation (median 2.02 [IQR 1.84--3.32] versus 3.03 [2.58--3.38]), but reached only 129/200 events versus 190/200 for all-layer backpropagation because earlier type-C readout updates rotated $\range W_L$ away from $\spn U$, rendering subsequent type-B targets unreachable by single-layer updates (SI \cref{sec:resnetmethods}).

\subsection{Generalisation to ResNet-18 with cross-entropy loss}
\label{sec:resnet}

We next evaluated five pre-trained seeds of CIFAR-10 ResNet-18 \citep{he2016} (with skip connections, batch normalisation, biases, and five-class cross-entropy loss; mean old-class test accuracy 96.1\%, range 95.7--96.5\%; five reserved readout rows and biases maintained at zero). We probed six hierarchical stages (\texttt{conv1}, \texttt{layer1}--\texttt{layer4}, and \texttt{fc}) across seven conditions (3 draws per seed, 105 conditions, 20 items each): \textbf{A}, \textbf{B}, $\mathbf{B}_{\mathrm{noise}}$, and $\mathbf{B}_{\mathrm{orth}}$ trained with five-class cross-entropy; \textbf{C} and \textbf{D} trained with binary cross-entropy on the single target reserved unit; and \textbf{D$^*$} trained with ten-class cross-entropy. With batch-normalisation running statistics frozen during adaptation, prospective predictions integrated linearised gradient descent on the exact cross-entropy or binary cross-entropy objective in active logit space using the block-wise tangent kernel $\Theta_b$, then mapped the resulting stage update to the 1{,}000 retention images via Jacobian--vector products to predict squared logit change, five-class conditional Kullback--Leibler (KL) divergence, cross-entropy increase, and accuracy drop (SI \cref{sec:resnetmethods}).

\begin{figure}[!htbp]
\centering
\includegraphics[width=\textwidth]{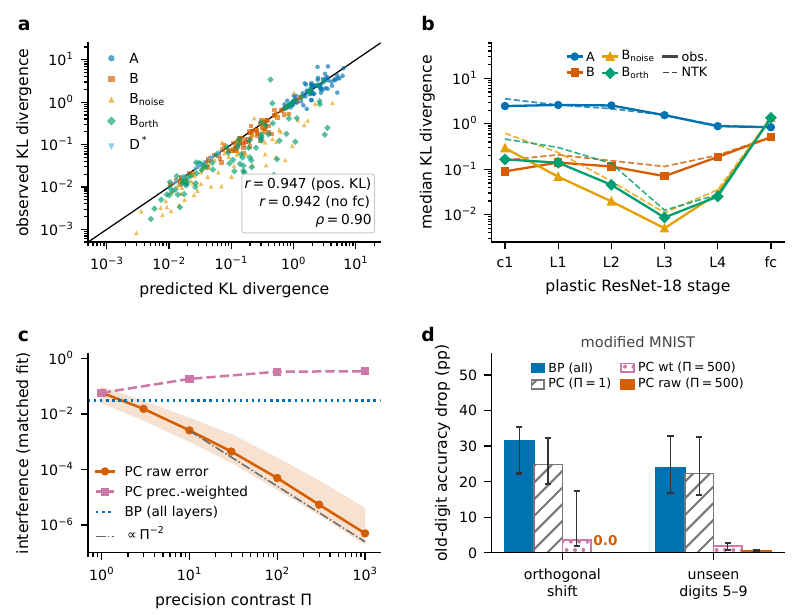}
\caption{\textbf{Generalisation to ResNet-18 (cross-entropy) and nonlinear predictive coding.} (a) Prospective block-wise logit-space NTK prediction versus observed old-class KL divergence across strictly positive-KL reached ResNet-18 runs (5 seeds $\times$ 3 draws; $n=356$ total, $n=298$ nonlinear blocks excluding \texttt{fc}; median per-condition $\rho=0.90$ holds both with and without \texttt{fc}). (b) Median KL divergence by plastic ResNet-18 stage (solid: observed; dashed: prospective NTK). (c) Matched-fit interference versus precision contrast $\Pi$ in four-layer $\tanh$ predictive-coding networks on type-B items ($n=10$ seeds; band: IQR; reference $\propto\Pi^{-2}$). (d) Old-digit accuracy drop (pp) on modified MNIST familiar-output adaptation in three-layer $\tanh$ networks ($n=10$ seeds; bars: medians; whiskers: IQR).}
\label{fig:generalization}
\end{figure}

Of 630 candidate single-stage runs, 150 (C and D below \texttt{fc}) had identically zero gradients and 75 (D$^*$ below \texttt{fc}) had an irreducible loss floor; of the 405 reachable runs, 386 reached the 5\% criterion within 1{,}500 steps (19 did not). Because C and D under \texttt{fc}-only binary cross-entropy leave all five old-class logits strictly untouched (predicted and observed five-class KL identically zero across all 30 such runs), we evaluate logarithmic KL accuracy exclusively on the $n=356$ strictly positive-KL reached runs so zero-KL points cannot inflate log-scale correlation (\cref{fig:generalization}a, \cref{tab:quant}). Across these 356 positive-KL runs, the prospective block-wise kernel predicted log old-class KL divergence with $r=0.947$ (95\% 5-seed cluster bootstrap CI 0.927--0.963; median absolute $\log_{10}$ error 0.115, i.e.\ a $1.30\times$ multiplicative factor) and log logit interference with $r=0.940$ (median absolute $\log_{10}$ error 0.121). Restricting strictly to the $n=298$ purely nonlinear residual-block runs (excluding both zero-KL C/D and linear \texttt{fc}) yielded nearly identical accuracy: $r=0.942$ for log KL (95\% CI 0.923--0.959; median absolute $\log_{10}$ error 0.144, $1.39\times$ factor) and $r=0.934$ for log logit change, with median per-condition rank correlation $\rho=0.900$ (IQR 0.77--0.96) across stages for both metrics. Within individual five-class conditions, positive-KL log correlations were $r=0.871$ (A), $0.973$ (B), $0.902$ ($\mathrm{B}_{\mathrm{noise}}$), and $0.900$ ($\mathrm{B}_{\mathrm{orth}}$). Across these four conditions ($n=341$), the pre-learning kernel also predicted functional old-class accuracy drop ($r=0.885$, $\rho=0.932$) and cross-entropy increase ($r=0.846$, $\rho=0.933$). Crucially for decision-making, selecting the stage with minimum prospective KL across all linearly reachable stages before any run chose a stage that reached the 5\% criterion in 60 of 60 five-class adaptation conditions (A, B, $\mathrm{B}_{\mathrm{noise}}$, and $\mathrm{B}_{\mathrm{orth}}$; 0 unreached selections), matching the exact empirical optimal stage in 52 of 60 conditions (86.7\%) and a top-two stage in 59 of 60 (98.3\%), with median penalty ratio $1.07\times$ (IQR 1.02--1.23) on the 8 non-optimal picks; the same 1{,}000 held-out retention images (disjoint from all adaptation sets) served for prospective estimation and evaluation (SI \cref{tab:resnetdetail}).

Novelty geometry separated both optimal stage and forgetting cost (\cref{fig:generalization}b). Remapping familiar classes (A) caused the largest disruption (median optimal-stage KL 0.82, accuracy drop 30.4 pp), with \texttt{fc} optimal in 10 of 15 conditions. Unseen natural classes (B) reduced disruption roughly tenfold (KL 0.071, accuracy drop 1.1 pp) and shifted the optimal locus to intermediate residual stages (\texttt{layer1}--\texttt{layer3} in 15/15 conditions). High-novelty pixel shifts ($\mathrm{B}_{\mathrm{noise}}$ and $\mathrm{B}_{\mathrm{orth}}$) reduced disruption by another order of magnitude (KL 0.0050 and 0.0087; accuracy drop 0.2 and 0.4 pp), with \texttt{layer3} optimal in all 15 draws of each condition (prospective $\rho_{\mathrm{KL}}=1.00$ and $0.90$). Pure output-side novelty (C and D) concentrated 100\% of onset gradients in \texttt{fc} and slowed all-stage adaptation roughly 24--62-fold: among the 11 of 15 draws reaching criterion within 2{,}000 steps, completed-run medians were 1{,}586 (C) and 1{,}595 (D) steps (all-15 medians including the 4 censored draws at 2{,}000: 1{,}650 and 1{,}684) versus 27--65 steps across A--$\mathrm{B}_{\mathrm{orth}}$. Restricting adaptation to \texttt{fc} left five-class conditional probabilities strictly unchanged (KL and accuracy drop 0.000), whereas all-stage adaptation opened lower stages once \texttt{fc} grew and perturbed five-class probabilities substantially (completed-run medians: KL 1.37 and 3.39, accuracy drop 18.5 and 31.0 pp; all-15 medians: KL 1.09 and 2.59, accuracy drop 14.7 and 29.6 pp), illustrating why readout localisation matters even for pure output novelty.

\begin{table}[!htbp]
\centering\scriptsize
\renewcommand{\arraystretch}{1.08}
\setlength{\tabcolsep}{4pt}
\caption{Compact summary of empirical evaluations across predictive-coding, convolutional, and residual architectures (per-condition, matched-fit, and control breakdowns appear in SI \cref{sec:methods}). Paired Wilcoxon $p$-values for all-seed ($n=10$) and strictly matched-fit subsets are detailed in \cref{sec:pc} and SI \cref{sec:pcnonlinmethods}.}
\label{tab:quant}
\begin{tabular}{@{} >{\raggedright\arraybackslash}p{0.23\textwidth} >{\raggedright\arraybackslash}p{0.23\textwidth} >{\raggedright\arraybackslash}p{0.26\textwidth} >{\raggedright\arraybackslash}p{0.22\textwidth} @{}}
\toprule
Setting & Primary comparison & Main outcome & Control / secondary \\
\midrule
Linear 4-layer PCN, type~B & Raw PC ($\Pi=10^4$) vs.\ $\Pi=1$, BP & $I$: $1.1\times10^{-6}$ vs.\ 5.3, 2.45 & Type~A: 2.2--2.5 (all) \\
4-layer $\tanh$ PCN, type~B ($n=10$) & Raw PC ($\Pi=10^3$) vs.\ $\Pi=1$, wt.\ PC, BP & $5.0\times10^{-7}$ vs.\ $5.9\times10^{-2}$, 0.35, $3.1\times10^{-2}$ & Scaling $\Pi^{-1.87}$; $L_1$ share 1.00 \\
4-layer ReLU PCN, type~B ($n=10$) & Raw PC ($\Pi=10^3$) vs.\ $\Pi=1$, wt.\ PC, BP & 0.25 vs.\ 2.01, 9.08, 1.13 & $L_1$ share 0.96 vs.\ 0.07 (wt.) \\
Modified MNIST 3-layer $\tanh$ ($n=10$) & Raw PC ($\Pi=500$) vs.\ BP, BP-$L_1$, LR-scaled & Orth: 0.0 vs.\ 31.4 pp BP; raw: 0.45 vs.\ 23.8 pp & BP-$L_1$/LR-scaled: 0.0 / 0.65--0.70 pp; PC $5.3\times$ fewer updates \\
6-layer CNN, CIFAR-10 MSE (5 seeds) & Prospective NTK vs.\ observed ($n=305$) & Log $r=0.79$ (CI 0.73--0.82), $\rho=0.83$ & No readout: $r=0.83$; biased: $r=0.73$ \\
6-layer CNN shifts (5 seeds, bias-free) & $\mathrm{B}_{\mathrm{nat}}$ $\to$ $\mathrm{B}_{\mathrm{orth}}$ ($\nu_0$: $0.10\to0.99$) & Conv4 $I$: $0.21 \to 0.024$; $\rho=0.94$ & $3\times3$ patch $\nu=0.89$ on $\mathrm{B}_{\mathrm{orth}}$ \\
ResNet-18, CIFAR-10 CE (5 seeds) & Positive-KL NTK ($n=356$; no \texttt{fc} $n=298$) & KL $r=0.947$ (no \texttt{fc}: 0.942), $\rho=0.90$ & Exact optimal stage 52/60 (86.7\%); top-2 59/60 \\
\bottomrule
\end{tabular}
\end{table}

\section{Discussion}
\label{sec:discussion}

We asked which levels of a hierarchical network change during new learning and what interference cost that adaptation imposes on prior knowledge. In deep linear networks, onset gradients are restricted to the contiguous plasticity window reached by both bottom-up activity and top-down error. Relative to learned input and output subspaces in balanced networks, this geometry separates pure novelty into four types that govern onset locus, acquisition rate, and interference cost, while distinguishing the minimum cost any learner must pay from the extra excess of local rules. Experiments across nonlinear predictive-coding, convolutional, and cross-entropy ResNet-18 models show that precision-selected localisation and layer-resolved tangent-kernel predictions retain quantitative utility in the nonlinear architectures examined here.

\subsection{Relation to prior and recent work}

Our analysis builds on exact deep linear network dynamics \citep{saxe2014,saxe2019}, extensions to prior knowledge and lazy-versus-rich transitions \citep{braun2022,domine2025}, and teacher--student continual learning models \citep{lee2021} that parallel human transfer and forgetting \citep{holton2026}. Whereas those studies treat transfer and forgetting at the whole-network level, novelty types B and C resolve feature and readout similarity across depth. In continual learning algorithms, orthogonal gradient descent and gradient projection memory actively project updates onto subspaces orthogonal to past gradients or layerwise feature spaces \citep{farajtabar2020,saha2021,tang2021,saha2023}. Our work is complementary: rather than imposing explicit projections, we characterise when unconstrained gradient dynamics naturally localise to a protective locus, when joint multi-layer updates leak into shared upper weights (\cref{thm:typeB}c--d), and how local precision contrasts steer plasticity without storing projection bases.

In the neural tangent kernel regime, recent studies show that forgetting concentrates in low-rank old-task eigenmodes \citep{hidekel2026}, link training-time forgetting and generalisation to cross-task kernels \citep{zheng2026}, and use NTK perspectives to guide parameter-efficient continual fine-tuning \citep{liu2026}. \Cref{thm:kernel,thm:cost} complement these analyses by resolving the tangent kernel stage by stage to characterise where adaptation lands and forecast single-event forgetting across nonlinear residual blocks before any weight update. At the same time, our sequential CNN experiment highlights an important distinction between instantaneous interference minimisation and long-term stability: single-event linearised stage selection does not automatically prevent cumulative nonlinear drift across repeated events. Separately, \citet{luo2026} documented two timescales of forgetting in language models, separating rapid output-level suppression from slow latent drift; novelty geometry offers a complementary structural account of why output-side shifts localise at upper stages while deeper rewriting unfolds more slowly. Our vision experiments also refine empirical findings on surgical fine-tuning \citep{surgical2023} and upper-layer forgetting \citep{ramasesh2021}: even under pixel-orthogonal shifts, local $3\times3$ patch overlap offers a plausible mechanism for intermediate convolutional stages remaining optimal rather than the first layer. Loss of plasticity over long non-stationary streams \citep{dohare2024} addresses whether a network retains capacity to adapt, whereas our framework analyses where adaptation lands in a responsive hierarchy.

In predictive coding \citep{rao1999,friston2005,bastos2012,keller2018,whittington2017,millidge2022}, recent studies attribute low interference to prospective configuration \citep{song2024}, explain sample efficiency via target alignment \citep{oliviers2026}, and restore dynamic precision weighting via hierarchical Gaussian filters \citep{baskakovs2026}. Our results isolate a sharp distinction: among learners that reached the adaptation criterion with equal precisions, equilibrium predictive coding interfered \emph{more} than backpropagation on type-B items in linear and smooth nonlinear networks. Protection emerged specifically when a precision contrast localised equilibrium error to layer 1 and plasticity read raw rather than precision-weighted errors; explicit layer-1 and learning-rate-scaled backpropagation controls confirm that localisation itself drives retention, while activity inference improves weight-update efficiency at substantial iterative inference wall-clock cost. Finally, recent macaque recordings report modest inferior temporal cortex plasticity after object-discrimination training, placing most change downstream \citep{sorensen2026}; because new discriminations over familiar visual features may correspond to output-side novelty (type~C), the plasticity window places onset plasticity at upper readout stages in that regime.

\subsection{Mechanistic connections to schemas and perceptual learning}

Two minimal linear models in the SI illustrate qualitative links to classical memory and perceptual phenomena without claiming quantitative data fits. In a schema-consolidation model \citep{tse2007,tse2011} (SI \cref{sec:schema}), a new flavour paired with a known arena location forms a type-B item that support-weighted predictive coding consolidates rapidly with minimal interference, whereas a new-arena pair forms a type-D near-saddle whose escape under controlled residual perturbations requires tens of replays and disrupts prior associations. In a perceptual-learning model (SI \cref{sec:perceptual}), raising discrimination difficulty attenuates task-relevant signals above level 1 and shifts backpropagation's weight change downward, qualitatively echoing reverse hierarchy theory \citep{ahissar2004,wenliang2018}.

\subsection{Implications and testable hypotheses}

For hierarchical learners governed by outer-product gradient-like plasticity (\cref{eq:grad}), the theory suggests five falsifiable hypotheses; biological tests require mapping abstract depth onto cortical areas and controlling for baseline representational drift \citep{deitch2021}.
\begin{enumerate}[leftmargin=2.5em,itemsep=1pt]
\item[H1] \textbf{Locus and interference asymmetry by novelty type.} Adapting a novel cue to a familiar outcome (type~B) should recruit early stages and permit low-interference integration when confined there, whereas adapting a familiar stimulus to a novel outcome (type~C) should recruit upper readout stages and disturb other associations in proportion to high-level code overlap (\cref{cor:types,cor:speed,thm:typeB}). Under balanced unit-mode conditions, types B and C share the same onset rate (\cref{cor:speed}); the asymmetry lies in locus and interference rather than onset speed.
\item[H2] \textbf{Joint multi-level learning disturbs outcome-sharing memories.} For weak, single-mode type-B associations under concurrent all-level gradient flow, interference should remain restricted to memories sharing the target outcome and scale as $\gamma^4$ in association strength, whereas early-stage-confined or one-shot updates avoid this cost (\cref{thm:typeB,thm:modes}).
\item[H3] \textbf{Replay cross-moments govern collateral forgetting.} During partial replay, damage to unrehearsed memories on untouched dimensions should track the uncentred cross-moment $C_R$ between replayed components on touched and untouched modes (\cref{thm:replay}), a quantity measurable in decoded population replay.
\item[H4] \textbf{Precision shifts the locus under raw-error plasticity.} Provided error reachability and nonzero presynaptic support hold, raising prediction-error precision at a given level under raw-error synaptic plasticity should shift plasticity away from that level, whereas precision-weighted plasticity does not guarantee equivalent localisation (\cref{thm:pc}).
\item[H5] \textbf{Saddle dynamics motivate an auxiliary episodic store for doubly novel items.} Motivated by the depth-three saddle trade-off under local gradient dynamics after synaptic downscaling (SI \cref{lem:R1}), associations novel on both input and output sides should face a sharper speed--interference trade-off than familiar-outcome associations when learned directly in a deep hierarchy, creating a stronger demand for buffered interleaved replay (\cref{thm:saddle}).
\end{enumerate}

\subsection{Limitations}
\label{sec:limitations}

\paragraph{Analytical assumptions}
The exact theorems hold for bias-free deep linear networks, most under balance and spectral-structure assumptions (A1--A2). Perturbation bounds quantify leakage when balance is approximate (SI \cref{sec:robust}), while piecewise-linear results apply to a single item within a fixed activation pattern; exact ReLU mode-locality requires segregated assemblies with inhibitory cross-connections (SI \cref{thm:unit}). Closed-form type-B interference assumes continuous joint gradient flow, and the type-D saddle theorem restricts local negative-curvature and aligned-flow escapes without ruling out coordinated non-gradient updates. In predictive coding, doubly exponential share convergence requires the common-eigenvector error condition, support weighting is a heuristic rather than a variational update, and practical iterative nonlinear inference is approximate and not guaranteed to attain the global minimisers assumed in SI \cref{prop:nonlinearpc}.

\paragraph{Empirical and computational scope}
Our vision benchmarks use five pre-trained seeds per architecture, pure output-side novelty (C and D) uses zero-initialised reserved readout rows (relaxed via the residual-weight sweep), and prospective kernel evaluations are conditional on reaching the adaptation criterion and require representative old-data inputs. In convolutional networks, local $3\times3$ patch overlap is consistent with conv1 remaining suboptimal in the tested conditions, kernel calibration degrades at early layers when long trajectories leave the linear regime, and myopic single-event NTK selection can accumulate nonlinear drift across long sequential streams. In nonlinear predictive coding, float32 rounding residuals affect exact-projection ReLU initialisation and produced unequal criterion-reaching rates across baselines in the ReLU experiment, iterative L-BFGS inference adds substantial wall-clock overhead relative to learning-rate-scaled backpropagation, and localising plasticity to layer 1 trades held-out new-digit generalisation for substantial retention gains. Finally, all biological hypotheses await direct multi-area neural recording tests across controlled input and output novelty conditions \citep{poort2015}.

\subsection{Conclusion}

In deep linear hierarchies trained by gradient-based rules, onset plasticity is confined to the band of levels reached by both bottom-up activity and top-down error. Under balanced spectral structure with old inputs supported in the learned input subspace, novel inputs mapped onto familiar outcomes admit interference-free integration at the bottom level, whereas doubly novel items at depth three or more confront negative-curvature and aligned-flow gradient dynamics with a speed--interference trade-off. Empirical tests support selected predictions of the theory in nonlinear networks: raw-error precision contrasts localise plasticity in nonlinear predictive coding (reducing median old-digit accuracy loss from 23.8 to 0.45 percentage points on unseen MNIST digits and from 31.4 to 0.0 percentage points on orthogonalised shifts), and pre-learning layer-resolved tangent kernels forecast stage-wise KL divergence across purely nonlinear ResNet-18 blocks ($r=0.942$, $\rho=0.90$) and optimal stage selection across all candidate stages (52/60 familiar-output conditions). More broadly, characterising where gradient-based plasticity lands provides a concrete mathematical lens for analysing continual adaptation, local credit assignment, and interference in multi-layer neural systems.

\section*{Code and data availability}
The code for analysis and reproducibility will be made publicly available upon publication. The vision experiments use the public CIFAR-10 \citep{krizhevsky2009} and MNIST \citep{lecun1998} benchmarks.

\section*{Declaration of competing interest}
The authors declare no competing financial interests or personal relationships that could appear to influence the work reported in this paper.

\section*{Funding}
This research did not receive any specific grant from funding agencies in the public, commercial, or not-for-profit sectors.

\bibliographystyle{model5-names}
\bibliography{refs}

\clearpage
\appendix
\setcounter{section}{0}
\setcounter{figure}{0}
\setcounter{table}{0}
\setcounter{equation}{0}
\setcounter{theorem}{0}
\setcounter{lemma}{0}
\setcounter{proposition}{0}
\setcounter{corollary}{0}
\renewcommand{\thesection}{S\arabic{section}}
\renewcommand{\thefigure}{S\arabic{figure}}
\renewcommand{\thetable}{S\arabic{table}}
\renewcommand{\theequation}{S\arabic{equation}}
\renewcommand{\thetheorem}{S\arabic{theorem}}
\renewcommand{\thelemma}{S\arabic{lemma}}
\renewcommand{\theproposition}{S\arabic{proposition}}
\renewcommand{\thecorollary}{S\arabic{corollary}}

\begin{center}
{\LARGE\textbf{Supplementary Information}}
\end{center}
\medskip
\noindent Sections, figures, tables, equations, and theorems prefixed with S refer to these appendices; unprefixed references refer to the main text above.

\section{Proofs of the main results}
\label{sec:proofs}

\subsection{Lemmas}

\begin{lemma}[Layer-wise gradient]\label{lem:grad}
$\nabla_{W_l}\ell=-\beta_l(e)h_{l-1}(x)\T$ and $\norm{\nabla_{W_l}\ell}_F=\norm{\beta_l}\norm{h_{l-1}}$.
\end{lemma}
\begin{proof}
Write $f=AW_lb$ with $A=W_{L:l+1}$ and $b=h_{l-1}$. Then $d\ell=-e\T A\,dW_l\,b=-\tr(b\,e\T A\,dW_l)$, so $\nabla_{W_l}\ell=-A\T e\,b\T$. The Frobenius norm of a rank-one matrix is the product of the norms of its factors.
\end{proof}

\begin{lemma}[Nesting]\label{lem:nesting}
$\ker W_{j:1}\subseteq\ker W_{j+1:1}$ and $\ker W_{L:l+1}\T\subseteq\ker W_{L:l}\T$. Hence, for $x\neq0$ and $e\neq0$, $\{l:h_{l-1}(x)\neq0\}=\{1,\dots,\din(x)\}$ and $\{l:\beta_l(e)\neq0\}=\{\dout(e),\dots,L\}$.
\end{lemma}
\begin{proof}
If $W_{j:1}x=0$ then $W_{j+1:1}x=W_{j+1}(W_{j:1}x)=0$; if $W_{L:l+1}\T e=0$ then $W_{L:l}\T e=W_l\T(W_{L:l+1}\T e)=0$. So the levels with nonzero $h_{l-1}$ form a prefix and those with nonzero $\beta_l$ a suffix, both nonempty because $h_0=x\neq0$ and $\beta_L=e\neq0$.
\end{proof}

\begin{lemma}[Balanced networks with learned spectral structure]\label{lem:balanced}
Under A1--A2 there are matrices $R_l\in\R^{n_{l-1}\times k}$ with orthonormal columns, $R_1=V$ and $R_{L+1}=U$, such that $W_l=R_{l+1}S^{1/L}R_l\T$ for $l=1,\dots,L$. Consequently $h_{l-1}(x)=R_lS^{(l-1)/L}V\T x$ for $l\ge2$, $\beta_l(e)=R_{l+1}S^{(L-l)/L}U\T e$ for $l\le L-1$, and $K_l=US^{2(L-l)/L}U\T$ for $l<L$, $K_L=I$.
\end{lemma}
\begin{proof}
Let $W_1=A_1D_1B_1\T$ be a compact SVD. A1 gives $W_2\T W_2=W_1W_1\T=A_1D_1^2A_1\T$. Hence $W_2$ vanishes on $\range(A_1)^\perp$, and $Q:=W_2A_1D_1^{-1}$ has orthonormal columns ($Q\T Q=D_1^{-1}A_1\T W_2\T W_2A_1D_1^{-1}=I$), so $W_2=QD_1A_1\T$. By induction $W_l=A_lD_1A_{l-1}\T$ for all $l$, with $A_0:=B_1$ and orthonormal $A_l$. Then $W_{L:1}=A_LD_1^LB_1\T$ is a compact SVD of $USV\T$. Because the singular values are distinct, $D_1^L=S$, $A_L=U\Sigma$ and $B_1=V\Sigma$ for a diagonal sign matrix $\Sigma$. Absorbing the signs, set $R_1=V$, $R_{l+1}=A_l\Sigma$ for $1\le l<L$ and $R_{L+1}=U$. The consequences follow by substitution.
\end{proof}

\subsection{Window, novelty types and piecewise-linear networks}

\Mref{thm:window} is proved in the main text using \cref{lem:grad,lem:nesting}. For the saddle statement, see \cref{sec:proof-saddle}.

\begin{proof}[Proof of \mref{cor:types}]
By \cref{lem:balanced}, $h_1(x)=R_2S^{1/L}V\T x$, which vanishes if and only if $V\T x=0$; otherwise $h_j(x)=R_{j+1}S^{j/L}V\T x\neq0$ for all $j<L$, so $\din(x)=L$. Likewise $\beta_{L-1}(e)=R_LS^{1/L}U\T e$ vanishes if and only if $U\T e=0$, and otherwise $\beta_l\neq0$ for all $l$. A pure type satisfies exactly one of the four vanishing patterns, and a mixed item satisfies the pattern of its nonzero blocks. The norms follow from \cref{lem:grad,lem:balanced}, because the $R_l$ are isometries on $\R^k$.
\end{proof}

\begin{proof}[Proof of \mref{cor:blocks}]
For $l<L$ the range of $W_{L:l+1}$ lies in $\spn U$, so $P_{U^\perp}$ annihilates the term $W_{L:l+1}\Delta W_lW_{l-1:1}$; for $l>1$ the row space of $W_{l-1:1}$ lies in $\spn V$, so $P_{V^\perp}$ annihilates it on the right. For $l=1$ the first property applies (as $L\ge2$), and for $l=L$ the second. For a gradient step, $\Delta W_l=\eta\beta_lh_{l-1}\T$. The level-1 term is $\eta K_1e\,x\T$ with $K_1e\in\spn U$ depending only on $e_\parallel$, so it writes $(x_\parallel,e_\parallel)$ and $(x_\perp,e_\parallel)$. The level-$L$ term is $\eta e\,h_{L-1}(x)\T W_{L-1:1}$, whose row space lies in $\spn V$ and which depends on $x$ only through $x_\parallel$, so it writes $(x_\parallel,e_\parallel)$ and $(x_\parallel,e_\perp)$. Intermediate levels write only $(x_\parallel,e_\parallel)$.
\end{proof}

\begin{proof}[Proof of \mref{prop:relu}]
For positively homogeneous $\phi$, $\phi(z)=\phi'(z)z$ for every $z\neq0$, so $\phi(z_l)=D_lz_l$ when all pre-activations at $x$ are nonzero. Because the pre-activations depend continuously on the weights, the activation pattern at $x$ is constant in a neighbourhood of the current weights, where the network coincides with the local linear map $\{\tilde W_l\}$ with fixed masks $D_l$. Writing $f(x)=\tilde W_{L:l+1}D_lW_l\tilde h_{l-1}$ and applying \cref{lem:grad} with $A=\tilde W_{L:l+1}D_l$ gives $\nabla_{W_l}\ell=-D_l\tilde W_{L:l+1}\T e\,\tilde h_{l-1}\T=-\tilde\beta_l\tilde h_{l-1}\T$. Both activation-gated signals nest because $\tilde h_l=D_lW_l\tilde h_{l-1}$ and $\tilde\beta_{l-1}=D_{l-1}\tilde W_{L:l}\T e=D_{l-1}W_l\T\tilde\beta_l$ vanish whenever their predecessor vanishes, with $\tilde h_0=x\neq0$ and $\tilde\beta_L=e\neq0$. Hence the plasticity-window characterisation of \mref{thm:window} applies within this fixed activation pattern. For a non-homogeneous activation such as $\phi(z)=\min(\max(z,0),1)$, $\phi(z)\neq\phi'(z)z$ for $z>1$, and the construction fails (\cref{sec:verification}, Example~2).
\end{proof}

\subsection{Kernel, speed and cost}

\begin{proof}[Proof of \mref{thm:kernel}]
Expand $\prod_l(W_l+\Delta W_l)x'$. The zeroth-order term is $f(x')$. The first-order terms are $\sum_lW_{L:l+1}\Delta W_lW_{l-1:1}x'=\eta\sum_lW_{L:l+1}\beta_l\,h_{l-1}\T h_{l-1}(x')$, and $W_{L:l+1}\beta_l=K_le$. The terms containing two or more updates are bounded by $[(M+\eta g)^L-M^L-L\eta gM^{L-1}]\norm{x'}=O(\eta^2)$, where $M=\max\norm{W_l}$ and $g=\max\norm{\nabla_{W_l}\ell}$. With a single nonzero update there are no such terms.
\end{proof}

\begin{proof}[Proof of \mref{cor:speed}]
Under gradient flow, $-\dot\ell=\sum_l\norm{\nabla_{W_l}\ell}^2=\sum_l\norm{h_{l-1}}^2\norm{\beta_l}^2$. For unit $x$ and $e$ on mode $i$, each in-window level contributes $s_i^{2(l-1)/L}s_i^{2(L-l)/L}=s_i^{2(L-1)/L}$ by \cref{lem:balanced}; with $\ell=\tfrac12$, $-\dot\ell/\ell$ is twice the number of in-window levels times $s_i^{2(L-1)/L}$. The windows are given by \mref{cor:types}. For type B the only plastic level is 1, so \mref{thm:kernel} is exact with $\ip{h_0(x)}{h_0(x_o)}=\ip{x}{x_o}$; for type C it is $L$, with $K_L=I$.
\end{proof}

\begin{proof}[Proof of \mref{thm:cost}]
(a) If $h=0$, then $f'(x)=W'_{L:m}h=0=f(x)$ for all new weights of levels $m,\dots,L$. Otherwise, the frozen part maps $x\mapsto h$ and $x_o\mapsto H_{:,o}$, and the updated part $z\mapsto W'_{L:m}z$ is linear. With $a:=H^+h$, which satisfies $Ha=h$ because $h\in\range H$,
\[
\Delta=(W'_{L:m}-W_{L:m})Ha=\sum_oa_o\,[f'(x_o)-f(x_o)]=Da,
\]
where $D$ stacks the old output changes. By Cauchy--Schwarz, $\norm{\Delta}\le\norm{D}_F\norm{a}$, so $I=\norm{D}_F^2\geq\norm{\Delta}^2/\norm{a}^2$, and $\norm{a}^2=h\T(H^+)\T H^+h=h\T G^+h>0$.

(b) Let $W:=W_{L:m+1}$, so that $WW^+e=e$. If $h\in\range H$, set $\Delta W_m:=W^+e\,a\T H^+/\norm{a}^2$. Because $a\in\range(H\T)$, $H^+Ha=a$, so the output change at $x$ is $e\,a\T a/\norm{a}^2=e$, the old output changes are $D=e\,a\T/\norm{a}^2$, and $\norm{D}_F^2=\norm{e}^2/\norm{a}^2=B_m$. If $h\notin\range H$, let $h_\perp:=(I-P_{\range H})h\neq0$ and $\Delta W_m:=W^+e\,h_\perp\T/(h_\perp\T h)$; then $\Delta W_mH=0$ and $W\Delta W_mh=e$.

(c) The update is a single-level update, so the output changes are exact: $W\Delta W_mh=e$ at $x$ and $e\ip{h}{h_o}/\norm{h}^2$ at $x_o$. Summing the squared norms gives $I^\delta_m=\norm{e}^2h\T Gh/\norm{h}^4$. If $h\in\range G=\range H$, the Cauchy--Schwarz inequality applied to $G^{1/2}h$ and $(G^+)^{1/2}h$ gives $\norm{h}^2=\ip{G^{1/2}h}{(G^+)^{1/2}h}\le\norm{G^{1/2}h}\norm{(G^+)^{1/2}h}$, i.e.\ $\rho_m=(h\T Gh)(h\T G^+h)/\norm{h}^4\ge1$, with equality if and only if the two vectors are parallel, i.e.\ $h$ is an eigenvector of $G$. The upper bound is the Kantorovich inequality \citep{kantorovich1948} applied to $G$ restricted to its range.
\end{proof}

\begin{corollary}[Monotonicity of the bound]\label{cor:monotone}
If $h_{m-1}\in\range H_{m-1}$, then $h_{m'-1}\in\range H_{m'-1}$ and $B_{m'}\ge B_m$ for every $m'\ge m$. If $x\in\spn X_o$, every update of any set of levels that fits the item ($\Delta=e\neq0$) causes $I\ge B_1>0$.
\end{corollary}
\begin{proof}
If $H_{m-1}a=h_{m-1}$ then $H_ma=W_mH_{m-1}a=h_m$, so every coefficient vector feasible at level $m-1$ is feasible at level $m$, and the minimum-norm coefficient, whose squared norm is $h\T G^+h$, can only decrease. For the second statement, let $m$ be the lowest updated level; $h_{m-1}\neq0$ because the item is fitted, $h_0=x\in\range H_0$, and (a) with monotonicity gives $I\ge B_m\ge B_1$.
\end{proof}

\begin{proof}[Proof of \mref{cor:aniso}]
$G_{m-1}=W_{m-1:1}X_oX_o\T W_{m-1:1}\T=c\,R_mS^{(m-1)/L}V\T VV\T VS^{(m-1)/L}R_m\T=cR_mS^{2(m-1)/L}R_m\T$ by \cref{lem:balanced} and A4; for $m=1$, $G_0=cVV\T$. For $x\in\spn V$, $x$ is an eigenvector of $G_0$, so $\rho_1=1$. The nonzero eigenvalues are $cs_j^{2(m-1)/L}$, so $\kappa_m=(s_1/s_k)^{2(m-1)/L}$, and $(\kappa+1)^2/4\kappa$ is strictly increasing in $\kappa\ge1$. The worst case is attained by $h$ in the span of the extreme eigenvectors with equal weights, which is a familiar input with reachable error.
\end{proof}

\subsection{Type B and mode-locality}

\begin{proof}[Proof of \mref{thm:typeB}]
(a) $W_{L:2}W_{L:2}^+e=e$ because $e\in\spn U=\range W_{L:2}$, and $\Delta W_1x_o=0$ because $x\perp x_o$. For the single step, the window is $\{1\}$, so \mref{thm:kernel} is exact and gives $f'(x)=f(x)+\eta\norm{x}^2K_1e=f(x)+\eta\norm{x}^2s_i^{2(L-1)/L}e$.

(b) Every update of $W_1$ has the form $\eta_t\beta_1(e_t)x\T$, which annihilates every $x_o\perp x$; the other levels are frozen, so old outputs are unchanged for any step-size sequence $\{\eta_t\}$. The fitted quantity $r:=W_1x$ evolves as $r_{t+1}=r_t+\eta_t\norm{x}^2W_{L:2}\T(y-W_{L:2}r_t)$, so the output error $e_t=y-W_{L:2}r_t\in\spn U$ satisfies $e_{t+1}=(I-\eta_t\norm{x}^2K_1)e_t$. In the eigenbasis $U$ of $K_1=US^{2(L-1)/L}U\T$ (\cref{lem:balanced}), mode $i$ contracts by $(1-\eta_t\norm{x}^2s_i^{2(L-1)/L})$ per step. Thus $e_t\to0$ whenever $\lim_{T\to\infty}\prod_{t=0}^{T-1}(1-\eta_t\norm{x}^2s_i^{2(L-1)/L})=0$ for all $i=1,\dots,k$, and in particular for any fixed step size $0<\eta<2/(\norm{x}^2s_1^{2(L-1)/L})$.

(c) The family is \mref{thm:modes}'s invariant family with $\M=\{i\}$. The coordinates $(p,a_2,\dots,a_L)$ are orthonormal coordinates on it: each derivative $\partial W/\partial(\cdot)$ is a unit-norm rank-one matrix, and they are mutually orthogonal. Hence the matrix gradient flow, which stays in the family, equals the gradient flow of $\ell=\tfrac12(\gamma-Pp)^2$ with $P:=\prod_{l\ge2}a_l$:
\[
\dot p=(\gamma-Pp)P,\qquad\dot a_l=(\gamma-Pp)\,pP/a_l.
\]
The gain of level 1 on $\spn V$ does not change, because $\nabla_{W_1}\ell\propto R_2e_ix\T$ and $x\perp\spn V$. Then $\tfrac{d}{dt}p^2=2(\gamma-Pp)Pp=\tfrac{d}{dt}a_l^2$, with $p(0)=0$ and $a_l(0)=s^{1/L}$.

(d) $\tfrac{d}{dt}(Pp)=(\gamma-Pp)\big[P^2+p^2P^2\sum_la_l^{-2}\big]>0$ while $Pp<\gamma$, so $Pp\uparrow\gamma$. By (c), $P=(s^{2/L}+p^2)^{(L-1)/2}$; writing $p=s^{1/L}q$ gives $Pp=s\,q(1+q^2)^{(L-1)/2}=\gamma$. The left side is strictly increasing in $q$, so the root is unique. An old item's mode-$i$ output is $s^{1/L}Pc$, initially $sc$, and its change is $cs((1+q^2)^{(L-1)/2}-1)=cs(g/q-1)$; the other modes are unchanged (\mref{thm:modes}). For small $g$, $q=g+O(g^3)$ and $(1+q^2)^{(L-1)/2}-1=\tfrac{L-1}2g^2+O(g^4)$.
\end{proof}

\begin{proof}[Proof of \mref{thm:modes}]
Let $E_\M$ be the coordinate projector onto $\M$ in $\R^k$, $x_\perp:=x-VV\T x$ and $\hat x_\perp$ its direction ($p=0$ if $x_\perp=0$). Consider the family $\F$ of weights
\[
W_1=R_2\big[(S^{1/L}+\Delta_1)V\T+p\,\hat x_\perp\T\big],\qquad W_l=R_{l+1}(S^{1/L}+\Delta_l)R_l\T\ (2\le l\le L),
\]
with each $\Delta_l=E_\M\Delta_lE_\M$ and $p\in\range E_\M$. The trained network ($\Delta_l=0$, $p=0$) is in $\F$.
(i) Every factor $S^{1/L}+\Delta_l$ is block diagonal with respect to $\M\oplus\M^c$ and equals $S^{1/L}$ on $\M^c$, and products preserve this.
(ii) Within $\F$, $V\T x$ and $p$ are supported on $\M$, so $h_{l-1}(x)\in R_l\range E_\M$ for $l\ge2$. The target satisfies $U\T y=U\T e_0+SV\T x$, which is supported on $\M$, and $f(x)\in U\range E_\M$, so $e=U(U\T e)$ with $U\T e$ supported on $\M$ at all times. Hence $\beta_l\in R_{l+1}\range E_\M$ for $l<L$, and $\beta_L=e$. For $l\ge2$, $-\nabla_{W_l}\ell=\beta_lh_{l-1}\T=R_{l+1}(\text{an }\M\times\M\text{ block})R_l\T$. For $l=1$, $\beta_1x\T=R_2v(V\T x)\T V\T+R_2v\norm{x_\perp}\hat x_\perp\T$ with $v$ supported on $\M$, which updates $\Delta_1$ within its block and $p$ within $\range E_\M$. So every gradient step, of any size, maps $\F$ into itself, and so does gradient flow.
(iii) Let $x_o\in\spn V$ with $V\T x_o$ supported on $\M^c$. Then $\hat x_\perp\T x_o=0$ and $\Delta_1V\T x_o=0$, so $W_1x_o=R_2S^{1/L}V\T x_o$. By (i) every later factor acts on it as $S^{1/L}$, so $f_t(x_o)=USV\T x_o=f_0(x_o)$.
\end{proof}

\subsection{The type-D saddle}
\label{sec:proof-saddle}

\begin{proof}[Proof of \mref{thm:saddle}]
(a) $V\T x=0$ gives $\din(x)=1$, and $U\T e=U\T y=0$ gives $\dout(e)=L$, so the window is empty (\mref{thm:window}).

(b) Expand $\prod_l(W_l+\varepsilon q_lq_{l-1}\T)x$. Since $W_1x=0$, the first factor contributes $\varepsilon q_1\norm{x}$. At each later level, $W_{l+1}q_l=0$ because $q_l\perp\spn R_{l+1}$, the row space of $W_{l+1}$, so only the $\varepsilon$ term survives, giving $\varepsilon^L\norm{x}q_L$. For an old input $x_o\in\spn V$, the first factor gives $W_1x_o\in\spn R_2$ plus $\varepsilon q_1\ip{\hat x}{x_o}=0$. At every level the old representation lies in $\spn R_l$, which $\varepsilon q_lq_{l-1}\T$ annihilates because $q_{l-1}\perp\spn R_l$. So no $\varepsilon$ term ever acts on an old input. The same argument applies with a different coefficient $a_l$ on each level, which proves the claim about the coordinated update $W_l+a\,q_lq_{l-1}\T$ in the main text: the item's output is $a^L\norm{x}\hat y=y$ for $a=(\norm{y}/\norm{x})^{1/L}$, and old outputs are unchanged.

(c) Expanding $f(x)$ to second order in $\delta W$ gives $\delta_1f(x)=W_{L:2}\delta W_1x$ and $\delta_2f(x)=\sum_{l<l'}W_{L:l'+1}\delta W_{l'}W_{l'-1:l+1}\delta W_lW_{l-1:1}x$; only $l=1$ survives because $W_{l-1:1}x=0$ for $l\ge2$. Contracting with $e=y\perp\range W_L$ removes both $e\T\delta_1f(x)$ and every $l'<L$ in $\delta_2f(x)$, leaving $e\T\delta W_LW_{L-1:2}\delta W_1x$. Hence the second-order term in $\ell(W+\delta W)-\ell(W)=\tfrac12\norm{y-\delta_1f-\delta_2f-\cdots}^2-\tfrac12\norm{y}^2$ is $\tfrac12\delta^2\ell=\tfrac12\norm{W_{L:2}\delta W_1x}^2-e\T\delta W_LW_{L-1:2}\delta W_1x$. Negative curvature requires $e\T\delta W_Lz>0$ for $z:=W_{L-1:2}\delta W_1x$. For $L\ge3$, $z\in\range W_{L-1:2}\subseteq\spn R_L$, which is spanned by the old hidden codes $h_{L-1}(X_o)$ because the old inputs span $\spn V$. So $\hat y\T\delta W_Lz\neq0$ forces $P_{U^\perp}\delta W_Lh_{L-1}(x_o)\neq0$ for some old item: a first-order change of that item's output in $\spn U^\perp$, which the other levels' first-order terms (lying in $\spn U$) cannot cancel. Conversely, if old outputs are unchanged to first order, then $P_{U^\perp}\delta W_L$ vanishes on $\spn R_L$, so $e\T\delta W_Lz=0$ and the second-order term is nonnegative. For $L=2$, $z=\delta W_1x$ is unconstrained: $\delta W_1=q_1\hat x\T$ and $\delta W_2=t\,\hat yq_1\T$ with $q_1\perp\spn R_2$ give negative curvature for suitable $t>0$ and leave every old output unchanged.

(d) On the family, $h_{l-1}(x)=(\prod_{j<l}a_j)\norm{x}q_{l-1}$ and $\beta_l=(\prod_{j>l}a_j)(\hat y\T e)q_l$, using $W_l^{\mathrm{old}\,\top}q_l=0$ and $W_{l+1}^{\mathrm{old}}q_l=0$, both consequences of $q_l\perp\spn R_{l+1}$. So $\nabla_{W_l}\ell\propto q_lq_{l-1}\T$, the family is invariant, balance $a_l\equiv a$ is conserved, and $\dot a=c\,a^{L-1}(\sigma-c\,a^L)$. With $0<\varepsilon<(\sigma/(2c))^{1/L}$, the initial output $c\varepsilon^L$ lies strictly below $\sigma/2$. For $L=2$, $b=a^2$ obeys the logistic equation $\dot b=2cb(\sigma-cb)$ with carrying capacity $\sigma/c$, which gives the stated time to reach $b=\sigma/(2c)$. For $L\ge3$, the early phase $\dot a\approx c\sigma a^{L-1}$ integrates to $a(t)^{-(L-2)}=\varepsilon^{-(L-2)}-(L-2)c\sigma t$, and the remaining time to reach $a^L=\sigma/(2c)$ is $O(1)$.
\end{proof}

\subsection{Predictive coding and learned precision}

\begin{proof}[Proof of \mref{thm:pc}]
(a) Stationarity in $u_l$ ($1\le l<L$) gives $\pi_l\varepsilon_l=W_{l+1}\T\pi_{l+1}\varepsilon_{l+1}$, i.e.\ $\xi_l=W_{L:l+1}\T\xi_L$. Unrolling $u_l=W_lu_{l-1}+\varepsilon_l$ gives $y=u_L=f(x)+\sum_lW_{L:l+1}\varepsilon_l=f(x)+\sum_l\pi_l^{-1}K_l\xi_L$, i.e.\ $e=K_\pi\xi_L$, and $K_\pi\succeq\pi_L^{-1}I\succ0$ because $K_L=I$. Also $u_l=h_l(x)+\sum_{j\le l}W_{l:j+1}\varepsilon_j$.

(b) Let $\delta=1/\Pi$, $A:=K_m$ and $B:=\sum_{l\neq m}K_l$, so that $\xi_L=z_\delta:=(A+\delta B)^{-1}e$. For $m=L$, $A=I$ and $z_\delta=e+O(\delta)$. For $m<L$, $B\succeq K_L=I\succ0$, and since $e\in\range A$, $z_\delta=z_*+O(\delta)$, where $z_*$ is the unique solution of $Az=e$ with $P_{\ker A}Bz=0$ (write $z=z_1+z_0$ with $z_1\in\range A$, $z_0\in\ker A$ and expand the two block equations). With $W:=W_{L:m+1}$, $\varepsilon_m=\xi_m=W\T z_\delta$, and $W\T z_*$ solves $Wv=e$ and lies in $\range W\T$; the only such solution is $W^+e$. For $l\neq m$, $\varepsilon_l=\delta W_{L:l+1}\T z_\delta=O(\delta)$, and $u_{m-1}$ differs from $h$ only through $\varepsilon_{j<m}=O(\delta)$. Hence $\eta\,\varepsilon_mu_{m-1}\T=\eta\,W^+e\,h\T+O(\delta)$, which equals the target delta rule $W^+e\,h\T/\norm{h}^2$ up to $O(\delta)$ for $\eta=1/\norm{h}^2$, and also for $\eta=1/\norm{u_{m-1}}^2=1/\norm{h}^2+O(\delta)$ because $h\neq0$. For type B at $m=1$ with $\eta=1/\norm{x}^2$, $u_0=x$ exactly, the level-1 update $\varepsilon_1x\T/\norm{x}^2$ annihilates every $x_o\perp x$, and the remaining updates are $O(\delta)$ in norm; old outputs therefore change by $O(\delta)$ and the interference is $O(\delta^2)$.

(c) is immediate from (a).

(d) Apply the expansion in the proof of \mref{thm:kernel} to the simultaneous update $\Delta W_l=\eta\varepsilon_lu_{l-1}\T$: the first-order term at $x_o$ is $\eta\sum_lW_{L:l+1}\varepsilon_l\ip{u_{l-1}}{h_{l-1}(x_o)}$ and all products of two or more updates are $O(\eta^2)$. Use $W_{L:l+1}\varepsilon_l=\pi_l^{-1}K_l\xi_L$. With $\pi\equiv1$, the level-1 term vanishes for $x_o\perp x$, and for type B, $u_{l-1}=\sum_{j<l}W_{l-1:j+1}\varepsilon_j$ because $h_{l-1}(x)=0$ for $l\ge2$. The second-order term is nonzero in general (\cref{sec:verification}, Example~6).
\end{proof}

\begin{proof}[Proof of \mref{thm:learnedprec}]
(a) If $K_le=k_le$ for all $l$, then $K_\pi e=Ze$ with $Z=\sum_lk_l/\pi_l$, so $\xi_L=e/Z$ and $\norm{\varepsilon_l}^2=\norm{\xi_l}^2/\pi_l^2=k_l\norm{e}^2/(Z^2\pi_l^2)$. Levels with $k_l=0$ have $\varepsilon_l=0$ and $\sigma_l=0$ identically and are held inactive in the precision update. For active levels ($k_l>0$), the energy share is $\sigma_l=w_l/Z$ with $w_l:=k_l/\pi_l$. The update $\pi_l'=n_l/\norm{\varepsilon_l}^2$ gives $w_l'=k_l/\pi_l'=(\norm{e}^2/n_l)\,w_l^2/Z^2$, and normalising gives $\sigma_l'=(\sigma_l^2/n_l)/\sum_j(\sigma_j^2/n_j)$. Induction gives $\sigma_l^{(t)}\propto(\sigma_l^{(0)})^{2^t}n_l^{-(2^t-1)}=n_l(\sigma_l^{(0)}/n_l)^{2^t}$. Let $r_l:=\sigma^{(0)}_l/n_l$ and $r^*:=\max_lr_l$. For $r_l<r^*$, $\sigma_l^{(t)}/\sigma_j^{(t)}=(n_l/n_j)(r_l/r^*)^{2^t}\to0$ doubly exponentially for every maximiser $j$, while maximisers keep the ratio $n_l/n_j$. The precision ratio $\pi_l/\pi_j=(k_l/k_j)(\sigma_j/\sigma_l)$ therefore diverges for every non-maximiser $l$ with $k_l>0$. Note that this describes convergence of the normalised shares $\sigma_l$ and divergence of ratios $\pi_l/\pi_j$, rather than bounded convergence of absolute precisions $\pi_l$ (which depend on $\norm{e}^2$ and scale freely without altering the equilibrium). With uniform initial precisions, $\sigma_l^{(0)}\propto k_l=s^{2(L-l)/L}$, which is strictly decreasing in $l$ if $s>1$, strictly increasing if $s<1$ and constant if $s=1$; with equal widths the claims follow.

(b) $\varepsilon_l=\xi_l/\pi_l=\pi_l^{-1}W_{L:l+1}\T K_\pi^{-1}e$ by \mref{thm:pc}a. Under A0--A2 with $e\perp\spn U$, $K_le=0$ for $l<L$ (\cref{lem:balanced}), so $K_\pi e=\pi_L^{-1}e$, $K_\pi^{-1}e=\pi_Le$ and $W_{L:l+1}\T e=R_{l+1}S^{(L-l)/L}U\T e=0$. In general $K_\pi^{-1}e$ need not lie in $\ker W_{L:l+1}\T$ even when $K_le=0$ (\cref{sec:verification}, Example~8).

(c) Under the uniform-initial-precision, equal-width, single-mode conditions of (a) with $s<1$, the top level is the unique maximiser, so $\pi_l/\pi_L\to\infty$ for $l<L$ and $\varepsilon_l=O(\pi_L/\pi_l)\to0$. For type B, $h_{L-1}(x)=0$, so $u_{L-1}=\sum_{j<L}W_{L-1:j+1}\varepsilon_j\to0$. Hence the top-level raw-error update $\eta\varepsilon_Lu_{L-1}\T\to0$ and all lower-level updates $\eta\varepsilon_ju_{j-1}\T\to0$, so the limiting raw-error update vanishes and prevents learning in this limit.
\end{proof}

\subsection{Partial replay}

\begin{proof}[Proof of \mref{thm:replay}]
Use the family $\F$ of the proof of \mref{thm:modes} and write $A_l$ for the $\M$-block of $S^{1/L}+\Delta_l$, $A_{b:a}:=A_b\cdots A_a$ ($A_{L:L+1}:=I$) and $\Delta:=A_{L:1}-S_\M$. Within $\F$, a rehearsed item $x_r=Vc_r$ has output change $f(x_r)-f_0(x_r)=U_\M\Delta c_{r,\M}$, so its rehearsal error is $\tilde e_r:=f_0(x_r)-f(x_r)=-U_\M\Delta c_{r,\M}$ and $\beta_l(\tilde e_r)=R_{l+1}[N_lc_{r,\M};0]$ with $N_l:=-A_{L:l+1}\T\Delta$. Its bottom-up activity is $h_{l-1}(x_r)=R_l[A_{l-1:1}c_{r,\M};S_{\M^c}^{(l-1)/L}c_{r,\M^c}]$ for $l\ge2$, and $x_r=Vc_r$ for $l=1$. The rehearsal term contributes $\lambda\sum_r\beta_l(\tilde e_r)h_{l-1}(x_r)\T$ to the descent direction of $W_l$. Its $\M\times\M$ block keeps the weights in $\F$; its $\M^c$ rows vanish; and its block with $\M$ rows and $\M^c$ columns is
\begin{equation}\label{eq:offblock}
\lambda\,N_l\,C_R\,S_{\M^c}^{(l-1)/L}.
\end{equation}
At level 1 the rehearsal term has no component along $\hat x_\perp$, because $x_r\perp\hat x_\perp$. The new item's gradient keeps the weights in $\F$ (proof of \mref{thm:modes}).

(a) If $C_R=0$, \cref{eq:offblock} vanishes, every gradient step and the gradient flow map $\F$ into itself, and part (iii) of the proof of \mref{thm:modes} applies to every $\M^c$-supported old item, rehearsed or not.

(b) At a point of $\F$, the output of $x_o=Vc$ with $c$ supported on $\M^c$ changes at the rate $\sum_lW_{L:l+1}\dot W_lW_{l-1:1}x_o$. Here $W_{l-1:1}x_o=R_l[0;S_{\M^c}^{(l-1)/L}c]$ and $W_{L:l+1}$ maps the $\M$ rows to $U_\M A_{L:l+1}$. The in-block parts of $\dot W_l$ and the new item's gradient do not act on $[0;\cdot]$, and the $\M^c$ rows vanish, so only \cref{eq:offblock} contributes:
\[
\dot f(x_o)=\lambda\sum_lU_\M A_{L:l+1}N_lC_RS_{\M^c}^{2(l-1)/L}c=-\lambda\,U_\M\sum_lA_{L:l+1}A_{L:l+1}\T\Delta\,C_RS_{\M^c}^{2(l-1)/L}c.
\]
For $|\M|=1$ all $A$'s and $\Delta$ are scalars, and the sum equals $\Delta\,C_R\Lambda c$ with $\Lambda=\sum_lA_{L:l+1}^2S_{\M^c}^{2(l-1)/L}$, a positive diagonal matrix because the $l=L$ term is $S_{\M^c}^{2(L-1)/L}$.

(c) With $\lambda>0$ and a type-B item, the point of $\F$ with $\Delta_l=0$ and $p$ chosen so that $U_\M S_\M^{(L-1)/L}p\norm{x_\perp}=e$ has zero loss for the new item and leaves every rehearsed item unchanged, so the minimum of the objective within $\F$ is zero. A global minimiser within $\F$ therefore has $\tilde e_r=0$, i.e.\ $\Delta c_{r,\M}=0$, for all $r\in R$. An old item changes by $U_\M\Delta c_{o,\M}$, which vanishes when $c_{o,\M}\in\spn\{c_{r,\M}\}$. If the new input is familiar, zero loss need not be attainable, and the rehearsal frontier (\cref{prop:frontier}) describes the trade-off (\cref{sec:verification}, Example~9).
\end{proof}

\section{Additional analytical results}
\label{sec:further}

\subsection{Robustness: residual weight, soft windows and perturbations}
\label{sec:robust}

The exact zeros of the main text become small leaks when A1--A2 hold only approximately. The controlling quantity is the residual weight on directions that the training data never use.

\begin{lemma}[Residual weight]\label{lem:R1}
(R1) Let the first layer be linear, possibly followed by a nonlinear network, and let the training inputs lie in a subspace $\mathcal V$. Under gradient descent with step $\eta$ and decoupled weight decay $\lambda$ on any loss of the outputs at the training inputs, $W_1^{(t)}P_{\mathcal V^\perp}=(1-\eta\lambda)^tW_1^{(0)}P_{\mathcal V^\perp}$ exactly.
(R2) With squared loss, a linear last layer and targets in a subspace $\mathcal U$, $P_{\mathcal U^\perp}W_L^{(t+1)}=P_{\mathcal U^\perp}W_L^{(t)}\big((1-\eta\lambda)I-\eta G^{(t)}\big)$, where $G^{(t)}$ is the second moment of the penultimate activity over the training set. Its norm is non-increasing whenever $\eta(\lambda+\norm{G^{(t)}})\le2$ at every step.
\end{lemma}
\begin{proof}
(R1) $\nabla_{W_1}\mathcal L=-\tfrac1P\sum_\mu\delta_\mu x_\mu\T$ with $x_\mu\T P_{\mathcal V^\perp}=0$, so the update does not act on $\mathcal V^\perp$ and only the decay term does. (R2) $\nabla_{W_L}\mathcal L=-\tfrac1P\sum_\mu e_\mu h_\mu\T$ and $P_{\mathcal U^\perp}e_\mu=-P_{\mathcal U^\perp}W_Lh_\mu$; the eigenvalues of $(1-\eta\lambda)I-\eta G$ lie in $[-1,1]$ under the stated condition.
\end{proof}

\begin{proposition}[Soft window]\label{prop:soft}
For any deep linear network and $l\ge2$, $\norm{\nabla_{W_l}\ell}\le\norm{\beta_l}\norm{W_{l-1:2}}\norm{W_1x}$. For a unit probe $x\in\mathcal V^\perp$ with $\norm{\beta_1}>0$, the leak odds $\mathcal O:=\sum_{l\ge2}\norm{\nabla_{W_l}\ell}^2/\norm{\nabla_{W_1}\ell}^2$ satisfy
\[
\frac{\norm{\beta_2}^2}{\norm{\beta_1}^2}\norm{W_1x}^2\le\mathcal O\le P(t)\,r_{\mathrm{in}}^2,\qquad P(t):=\frac{\sum_{l\ge2}\norm{\beta_l}^2\norm{W_{l-1:2}}^2}{\norm{\beta_1}^2},\qquad r_{\mathrm{in}}:=\norm{W_1P_{\mathcal V^\perp}}.
\]
\end{proposition}
\begin{proof}
$\norm{\nabla_{W_l}\ell}=\norm{\beta_l}\norm{h_{l-1}}$ with $h_{l-1}=W_{l-1:2}W_1x$, and $\norm{W_1x}=\norm{W_1P_{\mathcal V^\perp}x}\le r_{\mathrm{in}}$; the lower bound is the $l=2$ term.
\end{proof}

By \cref{lem:R1}, $r_{\mathrm{in}}$ decays exactly as $(1-\eta\lambda)^t$ under training with decoupled weight decay. The prefactor $P(t)$ depends on the evolving weights and errors, so the leak odds are bounded above by the geometrically decaying envelope $P(t)r_{\mathrm{in}}(0)^2(1-\eta\lambda)^{2t}$ wherever $P(t)$ stays bounded; we do not claim an exact geometric law for the leak odds themselves. The leakage bound depends explicitly on residual weights in data-free directions rather than on the imbalance $W_{l+1}\T W_{l+1}-W_lW_l\T$: the balanced orthogonal initialisation $W_l(0)=\omega Q_l$ has zero imbalance for all time under continuous gradient flow (and remains approximately balanced under small-step discrete gradient descent up to $O(\eta^2)$ terms) but carries a residual $r_{\mathrm{in}}=\omega$ that does not decay without weight decay. In simulations with this initialisation, 3--12\% of the type-B gradient fell above level 1, and the ratio of the leak odds to $r_{\mathrm{in}}^2$ was nearly constant within each initialisation family (\cref{tab:robust}). R1 held to relative error $10^{-13}$. The output side is analogous with $r_{\mathrm{out}}:=\norm{P_{\mathcal U^\perp}W_L}$, which obeys the recurrence (R2) rather than an exact geometric law.

\begin{proposition}[Robustness of mode-locality]\label{prop:robust}
Let the balanced network be perturbed by $\norm{\Delta}\le\delta$, and run gradient flow on an item that satisfies the conditions of \mref{thm:modes}.
(a) Over a finite horizon $T$, protected outputs change by at most $C(1+e^{KT})\delta$.
(b) If the unperturbed flow converges to $W_\infty$ with $h_{L-1}(x;W_\infty)\neq0$, there are $\delta_0,C$ such that for $\delta\le\delta_0$, $\sup_{t\ge0}\norm{\tilde f_t(x_o)-\tilde f_0(x_o)}\le C\delta$ for every protected $x_o$, and $\norm{\tilde W_\infty-W_\infty}\le C\delta$.
\end{proposition}
\begin{proof}
(a) Split $\tilde f_T-\tilde f_0$ at $x_o$ into $[\tilde f_T-f_T]+[f_T-f_0]+[f_0-\tilde f_0]$. The middle term vanishes by \mref{thm:modes}; the outer terms are Lipschitz in the weights, and Grönwall's inequality gives $\norm{\tilde W(t)-W(t)}\le\delta e^{Kt}$.
(b) Let $J=\partial f_W(x)/\partial W$. Then $JJ\T=\Theta(x,x)\succeq\norm{h_{L-1}}^2I\succ0$ near $W_\infty$, so $\mathcal S=\{W:f_W(x)=y\}$ is locally a smooth submanifold. The Hessian of $\ell$ on $\mathcal S$ is $J\T J$, whose kernel is $T\mathcal S$, so $\ell$ is Morse--Bott and $\mathcal S$ is a normally attracting manifold of equilibria. By invariant-manifold theory \citep{fenichel1971,hirsch1977}, a neighbourhood of $W_\infty$ in which the flow converges to $\mathcal S$ has a $C^1$ stable foliation, and the flow is uniformly Lipschitz in its initial condition there for all $t\ge0$. The unperturbed flow enters this neighbourhood at a finite time; Grönwall's inequality covers the prefix, and composing the two bounds gives (b). Items that start at an unstable saddle (type D) are excluded, because a uniform-in-time bound cannot generally be guaranteed near unstable saddles (although finite-time continuous dependence still holds).
\end{proof}

Exact invariance (\mref{thm:modes}) holds for any step sizes; \cref{prop:robust} is a gradient-flow statement, and (b) additionally requires a regular endpoint. Numerically, for $\delta$ from $10^{-5}$ to $10^{-2}$ the supremum over time of protected interference scaled with slope 2.00 in $\delta$ (squared norm), with $\sup_tI/\delta^2$ between 1.991 and 1.999, and continuing for 60{,}000 steps after convergence changed protected outputs by $9\times10^{-30}$.

\subsection{Capacity}
\label{sec:capacity}

\begin{theorem}[Capacity bound]\label{thm:capacity}
Let the network be linear from level $j$ upward, $f(x)=W_{L:j+1}g(x)$ with $g$ arbitrary, and let $r:=\min_{j\le l<L}n_l$ be the smallest width in this linear suffix. Let $F_o:=f(X_o)$. For any new weights (at any levels) with $y':=f'(x)\neq0$,
\[
I=\norm{f'(X_o)-F_o}_F^2\ \ge\ \sum_{i\ge r}\sigma_i^2\big(P_{\hat y'^\perp}F_o\big),
\]
where $\sigma_1\ge\sigma_2\ge\dots$ are singular values. In particular, if $y'\perp\range F_o$ and $r=\operatorname{rank}F_o$, interference is bounded below by the energy of the weakest old mode: $I\ge\sigma_{\min}^2(F_o)$.
\end{theorem}
\begin{proof}
All outputs of the new network lie in $\range W'_{L:j+1}$, a subspace $\mathcal S$ of dimension at most $r$ that contains $y'$. For fixed $\mathcal S$, the smallest possible value of $\norm{f'(X_o)-F_o}_F^2$ is $\norm{P_{\mathcal S^\perp}F_o}_F^2$. Writing $\mathcal S=\spn(\hat y')\oplus\mathcal S'$ with $\mathcal S'\perp\hat y'$ and $\dim\mathcal S'\le r-1$, $\norm{P_{\mathcal S^\perp}F_o}^2=\norm{Z}^2-\norm{P_{\mathcal S'}Z}^2$ with $Z=P_{\hat y'^\perp}F_o$, and $\max_{\mathcal S'}\norm{P_{\mathcal S'}Z}^2=\sum_{i<r}\sigma_i^2(Z)$ \citep{eckart1936}.
\end{proof}

The theorem is a lower bound that holds for every update; whether it is attained depends on the architecture and on the inputs. A sufficient condition for attainment is that the network can realise the map that sends $x$ to $y'$ and each $x_o$ to its projection onto the optimal $\mathcal S$, which is the case, for example, when $g(x)$ is linearly independent of $g(X_o)$ and every level of the suffix has width at least $r$. If the new input coincides with an old input, the new output also constrains that old output and the bound can be strict: with $r=2$, old outputs $3u_1$ and $2u_2$ and a novel output $u_3$ for the first old input, the bound is 4 but every fitting update causes at least 10 (\cref{sec:verification}, Example~11). With bottleneck $r=k=4$ and independent inputs, the bound was 1.7995 and Adam on all layers converged to 1.7995 in six of six runs. The bound concerns a linear suffix: with a ReLU layer above the bottleneck, interference fell to $0.0000$ against a linear bound of 2.886, because the new item was placed in a different linear region (\cref{prop:region}).

\subsection{Transfer and interference in one kernel}
\label{sec:transfer}

\begin{theorem}[Transfer--interference kernel]\label{thm:transfer}
Assume A0, $e\in\range W_{L:m+1}$ and $h_{m-1}(x)\neq0$. The target delta rule at level $m$ changes every probe by $f'(x')-f(x')=\kappa_m(x,x')\,e$ with $\kappa_m(x,x'):=\ip{h_{m-1}(x)}{h_{m-1}(x')}/\norm{h_{m-1}(x)}^2$. Under A0--A2, for $x,x'\in\spn V$ with $a=V\T x$, $b=V\T x'$ and $m\ge2$, $\kappa_m=a\T S^{2p}b/a\T S^{2p}a$ with $p=(m-1)/L$; at $m=1$, $\kappa_1=\ip{x}{x'}/\norm{x}^2$.
\begin{enumerate}[leftmargin=2em,itemsep=1pt]
\item[(i)] (Categories defined by a novel feature.) Let $x=z+Va$ and $x_t=z+Va_t$ with $z\perp\spn V$, $z\neq0$, and $a\neq0$. Then $\kappa_1(x,x_t)=(\norm{z}^2+\ip{a}{a_t})/(\norm{z}^2+\norm{a}^2)\to1$ as $\norm{a},\norm{a_t}\to0$, while for $m\ge2$, $\kappa_m(x,x_t)=a\T S^{2p}a_t/a\T S^{2p}a$ does not depend on $z$; if $a_t$ is drawn independently of $a$ with zero mean, its expectation is zero. For an old item $x_o=Vb_o$, $\kappa_1(x,x_o)\to0$ as $\norm{a}\to0$, whereas for $m\ge2$, fixed direction $\hat a=a/\norm{a}$, and nonzero weighted overlap $\hat a\T S^{2p}b_o\neq0$, $\kappa_m(x,x_o)$ grows as $\norm{a}^{-1}$, so the target delta rule's interference grows as $\norm{a}^{-2}$.
\item[(ii)] (Categories sharing a strong mode.) Let $x=\alpha v_1+\beta Vu$ and $x_t=\alpha v_1+\beta Vw$ with $\alpha\neq0$, $\beta\neq0$, and unit $u,w$ supported on weak modes ($s_j\le s_w<s_1$). Then
\[
\kappa_m(x,x_t)=\frac{\alpha^2s_1^{2p}+\beta^2\,u\T S^{2p}w}{\alpha^2s_1^{2p}+\beta^2\,u\T S^{2p}u}.
\]
If $u$ and $w$ have disjoint mode supports, the cross term vanishes for every $p$ and $\kappa_m$ increases strictly with $m$. Orthogonality $u\T w=0$ alone does not suffice. For an old item with mode vector $b_o$ supported on weak modes, $|\kappa_m(x,x_o)|\le(|\beta|/\alpha^2)(s_w/s_1)^{2p}\norm{b_o}$, which decays geometrically in $m$.
\end{enumerate}
\end{theorem}
\begin{proof}
The kernel statement is the single-level computation in the proof of \mref{thm:cost}c, and the balanced form follows from \cref{lem:balanced}. (i) At $m=1$, $\ip{x}{x_t}=\norm{z}^2+\ip{a}{a_t}$ and $\norm{x}^2=\norm{z}^2+\norm{a}^2$. For $m\ge2$, $h_{m-1}(x)=W_{m-1:1}Va$ because $W_1z=0$; the expectation statement is linearity in $a_t$. For the old item, $\kappa_1(x,x_o)=\ip{a}{b_o}/(\norm{z}^2+\norm{a}^2)$, and for $m\ge2$ the numerator is linear and the denominator quadratic in $a$. (ii) Substitute $a=\alpha e_1+\beta u$ and $b=\alpha e_1+\beta w$; the $e_1$ cross terms vanish because $u,w$ are supported on weak modes. If the supports are disjoint, $\kappa_m=1/(1+(\beta^2/\alpha^2)\sum_ju_j^2(s_j/s_1)^{2p})$, which increases strictly in $p$. For the old item, the magnitude of the numerator is $|\beta\,u\T S^{2p}b_o|\le|\beta|s_w^{2p}\norm{b_o}$ and the denominator is at least $\alpha^2s_1^{2p}$.
\end{proof}

With $L=4$, $S=\mathrm{diag}(3,2,0.1)$, $\alpha=0.1$, $\beta=1$, $u=(e_2+e_3)/\sqrt2$ and $w=(-e_2+e_3)/\sqrt2$, which are orthogonal but share support, $\kappa_m$ is $0.0099,-0.6024,-0.8519,-0.9085$ for $m=1,\dots,4$ (\cref{sec:verification}, Example~10). In trained ReLU networks (8 seeds), the bottom locus was the least interfering for categories defined by a novel feature in 8 of 8 networks (median interference 0.3 versus at least 346 at higher loci), and abstract-category transfer rose with the locus in 7 of 8. The linear expectation that the bottom locus also transfers best for novel-feature categories did not hold robustly (median transfer 0.27, interquartile range 0.02--0.79); in ReLU networks the transfer kernel is gated, $J(x_t)J(x)^+$, so related items transfer only as far as they share activation patterns.

\subsection{The rehearsal frontier}

\begin{proposition}[Rehearsal frontier]\label{prop:frontier}
Let $J_\lambda=\tfrac12\norm{f(x)-y}^2+\tfrac\lambda2\sum_o\norm{f(x_o)-f_0(x_o)}^2$ with all old items rehearsed, $G=X_oX_o\T$, $x\in\range G$ and $u=x\T G^+x$, and assume the network can represent the needed change of its input--output map (true for type A in a network of sufficient width). At every global minimiser, $I=\norm{e}^2u/(\lambda+u)^2$ and $M:=\norm{f(x)-y}^2=\norm{e}^2\lambda^2/(\lambda+u)^2$, so $\sqrt I=(\norm{e}-\sqrt M)/\sqrt u$. No update of any linear network does better: $I\ge(\norm{e}-\sqrt M)^2/u$. If $x\notin\range G$ (for example, type B), every global minimiser has $I=M=0$ for every $\lambda>0$.
\end{proposition}
\begin{proof}
In function space, $J=\tfrac12\norm{e-\Delta Fx}^2+\tfrac\lambda2\tr(\Delta FG\Delta F\T)$; stationarity gives $\Delta F(\lambda G+xx\T)=ex\T$, and the Sherman--Morrison formula gives $I$ and $M$. The frontier follows from \mref{thm:cost}a at $m=1$ with $\norm{\Delta}\ge\norm{e}-\sqrt M$. For $x\notin\range G$, $f_0+ex_\perp\T/(x_\perp\T x)$ achieves $J=0$. Deep linear networks of sufficient width have no spurious local minima \citep{kawaguchi2016,laurent2018}.
\end{proof}

The closed forms matched gradient descent to six digits for $\lambda\in\{0.01,0.1,1,10\}$. For a familiar input whose new target conflicts with the old output, zero interference is not attainable at any $\lambda$ (\cref{sec:verification}, Example~9).

\subsection{Errors spread over several modes}
\label{sec:multimode}

Assume A0--A2 and a unit-norm type-B input ($\norm{x}=1$, $x\perp\spn V$) with error $e=U_\M\gamma$ supported on a set $\M$ of $m$ touched modes ($\gamma\in\R^m$). The invariant family of \mref{thm:modes} reduces joint gradient flow exactly to the block ordinary differential equation
\[
\dot p=A_{L:2}\T\epsilon,\qquad\dot A_l=A_{L:l+1}\T\,\epsilon\,(A_{l-1:2}\,p)\T,\qquad\epsilon=\gamma-A_{L:2}\,p,
\]
with $p\in\R^m$ and $A_l\in\R^{m\times m}$, $A_l(0)=S_\M^{1/L}$. Its dimension, $m+(L-1)m^2$, does not depend on the network width or the number of old items. The change of an old item with mode vector $c$ is exactly $U_\M(A_{L:2}-S_\M^{(L-1)/L})S_\M^{1/L}c_\M$. For small $\gamma$, expanding $A_l(t)=S_\M^{1/L}+O(|\gamma|^2)$ gives the first-order dynamics $\dot p^{(1)}=\Sigma(\gamma-\Sigma p^{(1)})$ with $\Sigma:=\mathrm{diag}(\sigma_1,\dots,\sigma_m)$ and $\sigma_j:=s_j^{(L-1)/L}$, so $p_k^{(1)}(t)=(1-e^{-\sigma_k^2 t})\gamma_k/\sigma_k$ and $\epsilon_j^{(1)}(t)=e^{-\sigma_j^2 t}\gamma_j$. Substituting into $\dot A_l$ and integrating $\int_0^\infty e^{-\sigma_j^2 t}(1-e^{-\sigma_k^2 t})\,dt=\sigma_k^2/[\sigma_j^2(\sigma_j^2+\sigma_k^2)]$, then summing the first-order product perturbation $\sum_{l=2}^L S_\M^{(L-l)/L}(\Delta A_l)S_\M^{(l-1)/L}$, yields $\Delta f=\mathrm{diag}(\gamma)K\mathrm{diag}(\gamma)c_\M+O(|\gamma|^4)$ in touched-mode coordinates, with
\[
K_{jk}=\frac{\sigma_k\sum_{l=2}^Ls_j^{2(L-l)/L}s_k^{(2l-3)/L}}{\sigma_j^2(\sigma_j^2+\sigma_k^2)},\qquad\sigma_j:=s_j^{(L-1)/L}.
\]
For $m=1$ this reduces to \mref{thm:typeB}d. The modes couple: $K_{jk}$ is large when output mode $j$ is weak and input mode $k$ is strong, so joint gradual learning moves the content of strong-mode memories into weak-mode outputs. The reduction matched full-network gradient descent to $8\times10^{-15}$, and the relative error of the leading-order formula fell as $|\gamma|^2$ (fitted slopes 1.87--1.98). The expansion is valid only for small association strengths: at $\gamma=(1,1)$ it overestimates the exact interference sixfold, and the exact reduced equation must be used there. A non-perturbative closed form for general spectra remains open.

\subsection{Piecewise-linear (ReLU) networks}
\label{sec:relu}

\begin{lemma}[Neuron-wise conservation; \citealp{du2018}]\label{lem:neuron}
For bias-free networks with positively homogeneous activation, gradient flow on any loss conserves $\norm{w^{\mathrm{in}}_i}^2-\norm{w^{\mathrm{out}}_i}^2$ for every hidden unit $i$.
\end{lemma}

This is the ReLU form of the conservation law of \mref{thm:typeB}c: an increase in the squared norm of a unit's incoming weights is accompanied by an equal increase in the squared norm of its outgoing weights. Recruiting units shared with old items can therefore introduce interference.

\begin{theorem}[Unit-locality with inhibitory segregation]\label{thm:unit}
Use ReLU with the derivative convention $\phi'(0)=0$. Let $x$ be orthogonal to all old inputs, and split each hidden layer into old-item units $\mathcal A_l$ and new-item units $\mathcal B_l$ such that initially: (1) $x$ drives no $\mathcal A$ unit, with strictly negative pre-activation at layer 1; (2) old items drive no $\mathcal B$ unit, and the layer-1 rows of $\mathcal B$ units are orthogonal to the span of the old inputs; (3) cross weights are inhibitory or absent, $W_l[\mathcal A_l,\mathcal B_{l-1}]\le0$ and $W_l[\mathcal B_l,\mathcal A_{l-1}]\le0$. Then under gradient descent on the new item, with any step sizes, every old output is unchanged for all time.
\end{theorem}
\begin{proof}
By induction over steps. While segregation holds, the gradient $D_l\tilde W_{L:l+1}\T e\,\tilde h_{l-1}\T$ has nonzero rows only for units active for $x$ (the $\mathcal B$ units) and nonzero columns only for units active for $x$ at the level below, because $\phi'(0)=0$ and $\phi(0)=0$. So only $W_l[\mathcal B,\mathcal B]$, the $x$-component of $W_1[\mathcal B,:]$ and $W_L[:,\mathcal B]$ change, and the cross weights do not change. An $\mathcal A$ unit's pre-activation for $x$ is at most $0$ (strictly negative at layer 1), and a $\mathcal B$ unit's pre-activation for an old item is at most $0$ by (3) and exactly $0$ at layer 1, because the layer-1 updates are along $x$. So segregation persists, and every changed weight multiplies an old-item activity that is zero.
\end{proof}

The theorem is exact under the stated derivative convention and in exact arithmetic. Units whose pre-activation is exactly zero lie on the kink, and in floating point their behaviour depends on rounding; strict margins make segregation robust. In the reported construction, strictly inhibitory cross weights gave interference $2.9\times10^{-31}$; with zero cross weights, old units sat on the kink for the new item and rounding recruited them, giving $1.6\times10^{-5}$ (in a second construction the same case stayed at $1.3\times10^{-30}$; \cref{sec:verification}, Example~12); excitatory cross weights gave 0.054--2.05. Lateral inhibition between assemblies thus plays the role that mode orthogonality plays in the linear theory.

\begin{proposition}[Capacity is per linear region]\label{prop:region}
\Cref{thm:capacity} holds for a piecewise-linear network whenever the new item and all old items share one activation pattern above the bottleneck in the updated network, and can fail otherwise.
\end{proposition}

\begin{proposition}[Nonlinear predictive coding: penalty limit]\label{prop:nonlinearpc}
Consider the pre-activation energy $E_\Pi(u)=\tfrac12\sum_l\pi_l\norm{u_l-W_l\hat\phi(u_{l-1})}^2$ with $u_0=x$, $u_L=y$, $\hat\phi=\mathrm{id}$ at $l=1$ and $\hat\phi=\phi$ otherwise, continuous $\phi$, $\pi_m=1$ and $\pi_l=\Pi$ for $l\neq m$. Assume (i) reachability: the set $\mathcal C$ of states with $u_l=W_l\hat\phi(u_{l-1})$ for all $l\neq m$ is nonempty; and (ii) for each $\Pi$, $E_\Pi$ has a global minimiser $u^*(\Pi)$, and these minimisers remain in a bounded set. Then $\sum_{l\neq m}\norm{\varepsilon_l(u^*(\Pi))}^2\le 2c^*/\Pi$, where $c^*$ is the minimum of $\tfrac12\norm{u_m-W_m\hat\phi(u_{m-1})}^2$ over $\mathcal C$, and every limit point of $u^*(\Pi)$ as $\Pi\to\infty$ minimises $\tfrac12\norm{u_m-W_m\hat\phi(u_{m-1})}^2$ over $\mathcal C$. The levels below $m$ are then at their feedforward values, and the raw-error update at level $m$ is a nonlinear target-propagation step. At $m=1$ with $\eta=1/\norm{x}^2$, the level-1 update $\varepsilon_1x\T/\norm{x}^2$ leaves $f(x_o)$ unchanged for every $x_o\perp x$ in any network, and the other levels' updates vanish as $\Pi\to\infty$.
\end{proposition}
\begin{proof}
This is the quadratic-penalty argument \citep{fiacco1968}. For any $\bar u\in\mathcal C$, $E_\Pi(u^*)\le E_\Pi(\bar u)=\tfrac12\norm{\bar u_m-W_m\hat\phi(\bar u_{m-1})}^2$; minimising over $\bar u$ gives $E_\Pi(u^*)\le c^*$, hence the bound on the penalty terms. A limit point $\hat u$ exists by boundedness, lies in $\mathcal C$ by continuity, and satisfies $\tfrac12\norm{\hat u_m-W_m\hat\phi(\hat u_{m-1})}^2\le c^*$, so it is a constrained minimiser. With the levels below $m$ constrained, $u_{m-1}$ is the feedforward activity. At $m=1$, $u_0=x$ is clamped, and $\varepsilon_1x\T x_o=0$ for $x_o\perp x$; the other levels' errors vanish by the penalty bound.
\end{proof}

The proposition concerns global minimisers of the energy; it does not guarantee that a particular inference algorithm finds them, and it gives the rate $O(\Pi^{-1/2})$ for the off-locus errors rather than a sharper one. Numerically, with inference started from the target's own activity pattern, interference fell from 202 to 4.3, 0.073 and 0.012 as $\Pi$ increased from 10 to $10^4$. Both assumptions and the initialisation mattered: with an unreachable target no run converged, and with inference started from the feedforward state, dead ReLU units trapped it at every $\Pi$ (\cref{tab:robust}).

\section{Precision dynamics and sensitivity analyses}
\label{sec:precision}

\paragraph{Ideal recursion and implementation}
\Mref{thm:learnedprec}a describes the ideal alternation of equilibrium inference and the free-energy update $\pi_l\leftarrow n_l/\norm{\varepsilon_l}^2$. Only precision ratios affect the equilibrium, so the implementation divides all precisions by their minimum after each update; this leaves the energy shares unchanged. Because losing levels' precision ratios grow doubly exponentially, the implementation caps them at $10^{12}$ and runs a finite number of updates (10 in \mref{fig:precision}, 6 per replay in the schema model, 8 per trial in the perceptual model). With the cap, shares below about $10^{-12}$ are not resolved; the selected level is unaffected in all reported runs. Convergence in the theorem refers to the energy shares, not to the absolute precisions.

\paragraph{Ties}
If several levels maximise $\sigma_l^{(0)}/n_l$, the recursion keeps their shares in the ratio of their widths and the other shares vanish. With equal widths, $s=1$ and uniform initial precisions, all shares stay equal (\cref{sec:verification}, Example~7: shares remain $0.25$ across updates). The implemented rule then does not select a single level.

\begin{figure}[!htbp]
\centering
\includegraphics[width=0.85\textwidth]{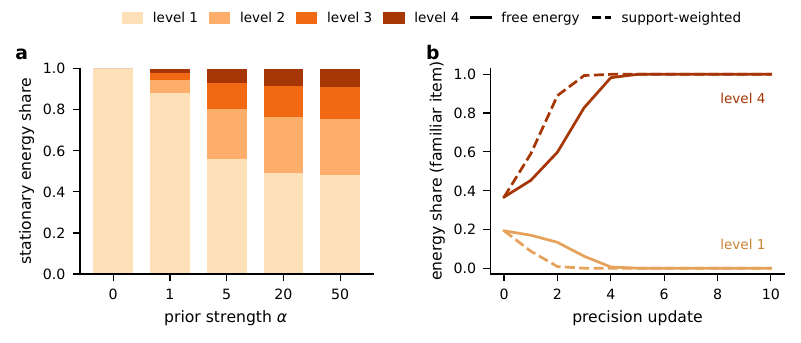}
\caption{\textbf{Graded locus and familiar items under learned precision.} Same network as \cref{fig:precision}. (a) Stationary energy share per level after 40 free-energy updates with a Gamma prior of increasing strength $\alpha$ (rate $\beta=\alpha/100$; shape $\alpha+1$), single-mode type-B item with $s=3$. (b) Energy share of each level over precision updates for a familiar (type-A) item, under the free-energy and the support-weighted updates; both select the top level.}
\label{fig:hyperprior}
\end{figure}

\paragraph{Gamma prior}
With a Gamma prior of shape $\kappa$ and rate $\theta$ on each precision, the negative log posterior adds $-(\kappa-1)\log\pi_l+\theta\pi_l$ to $F$, and stationarity gives
\[
\pi_l=\frac{n_l+2(\kappa-1)}{\norm{\varepsilon_l}^2+2\theta}.
\]
The implementation writes this as $\pi_l=(n_l+2\alpha)/(\norm{\varepsilon_l}^2+2\beta)$, which corresponds to shape $\kappa=\alpha+1$ and rate $\theta=\beta$. The prior bounds the precision ratios and makes the stationary locus graded rather than winner-take-all (\cref{fig:hyperprior}a): with $\beta=\alpha/100$, the stationary energy share of level 1 for a single-mode type-B item fell from 1 ($\alpha=0$) to 0.88, 0.56, 0.49 and 0.48 for $\alpha=1,5,20,50$. The prior, the minimum normalisation, the cap and the finite number of updates act separately: normalisation does not change shares, the cap limits resolution only, the number of updates sets how far the recursion has progressed, and the prior changes the fixed point.

\paragraph{Support-weighted precision}
The support-weighted update $\pi_l\leftarrow n_l/(\norm{\varepsilon_l}^2\norm{u_{l-1}}^2)$ (with the same prior form when used) weights each level by the energy of its raw-error update $\varepsilon_lu_{l-1}\T$. It is proposed here as a heuristic and is not derived from a generative model. In the type-B example with $s<1$, where the free-energy update selects the top and stalls, it selected level 1, after which one update fitted the item (misfit $6\times10^{-11}$). For the familiar item of \mref{fig:precision}, both the support-weighted and the free-energy updates selected the top level (\cref{fig:hyperprior}b), whereas under isotropic old inputs the local-rule analysis favours the bottom (\mref{cor:aniso}). Support weighting therefore resolves the presynaptic-support limitation of \mref{thm:learnedprec}c without being an optimal controller of the locus.

\paragraph{General (non-common-eigenvector) error spectra}
Outside the common-eigenvector error case of \mref{thm:learnedprec}a, the free-energy update (12 iterations) selected the level with the largest initial energy share in 60 of 60 random trials: balanced networks with $L=4$, width 16, five learned modes with singular values drawn log-uniformly from $[0.2,4]$, and random errors spread over all modes for alternating type-A and type-B inputs. This is an empirical observation; the closed-form recursion of \mref{thm:learnedprec}a does not apply there, and reachability must be checked via the general equilibrium formula $\varepsilon_l=\pi_l^{-1}W_{L:l+1}\T K_\pi^{-1}e$.

\section{Numerical illustrations and boundary examples}
\label{sec:verification}

Representative numerical evaluations in float64 arithmetic accompany the analytical results of \cref{sec:proofs,sec:further}. \Cref{tab:robust} summarises controlled comparisons that delineate the scope of the modelling assumptions, and \cref{tab:examples} collects illustrative boundary examples and counterexamples referenced throughout the main text and proofs.

\begin{table}[!htbp]
\centering\footnotesize
\caption{Controlled numerical comparisons delineating assumption scope and quantitative agreement.}
\label{tab:robust}
\begin{tabular}{@{}p{0.19\textwidth}p{0.24\textwidth}p{0.26\textwidth}p{0.23\textwidth}@{}}
\toprule
Analytical condition & Controlled comparison & Representative outcome & Interpretation \\
\midrule
Residual weight on data-free directions & Balanced orthogonal vs.\ small-weight initialisation & 3--12\% of type-B gradient above level 1; leak odds$/r_{\mathrm{in}}^2$ constant within each family & Leakage is set by residual weight (\cref{lem:R1,prop:soft}), not layer imbalance \\
Novelty defined on the learner & Teacher vs.\ learner SVD in trained ReLU networks & Teacher SVD mixes A and C; learner SVD places 97--99.7\% (B) at level 1 and 100\% (C) at top & Novelty must be measured against the trained network's own subspaces \\
Bias-free assumption & Bias-free vs.\ biased CIFAR-10 CNNs (main \cref{sec:cnn}) & Qualitative locus and kernel ordering unchanged (\cref{sec:cnnmethods}) & Biases soften exact zero-overlap bounds without removing locus structure \\
Bound vs.\ local-rule excess & $B_m$ vs.\ target-delta excess in balanced networks (old inputs in $\spn V$) & $B_m$ independent of $m$ for $x\in\spn X_o$; median excess rises 1.00 $\to$ 1.60 & Locus preference arises from the local rule (\mref{cor:aniso}) \\
Type-B learning schedule & One-shot vs.\ level-1-only vs.\ joint gradient descent & $<10^{-29}$ and $<10^{-28}$ vs.\ exact closed-form joint-flow interference & Interference arises from joint all-level dynamics (\mref{thm:typeB}) \\
Depth at type-D saddle & $L=2$ vs.\ $L\ge3$ with aligned and random residuals & Negative-curvature interference-free escape exists only at $L=2$; aligned flow matches \mref{thm:saddle}d to 3 digits & Depth-three transition of \mref{thm:saddle}c \\
Multi-mode type-B updates & Single-mode formula vs.\ exact reduced ODE & Reduced ODE matches full network to $8\times10^{-15}$; modes couple & Multi-mode interference is not additive across modes (\cref{sec:multimode}) \\
Precision structure & Equal precisions vs.\ contrast, raw vs.\ weighted errors & $\Pi=1$: 5.3 vs.\ BP 2.45; raw $\Pi=10^4$: $1.1\times10^{-6}$ (scales $\Pi^{-2}$) & Localisation requires precision contrast and raw-error updates (\mref{thm:pc}) \\
Probe-set structure & Probe sets with distinct support and overlap across 8 ReLU seeds & Bottom locus optimal in 8/8 seeds for novel-feature interference; transfer statistical & ReLU transfer depends on shared activation patterns (\cref{sec:transfer}) \\
Scale of novel-feature items & Varying $\norm{a}$ at loci $m\ge2$ & Interference grows as $\norm{a}^{-2}$ when $\hat a\T S^{2p}b_o\neq0$ & Matches \cref{thm:transfer}(i) \\
ReLU activation kinks & Inhibitory, zero, and excitatory cross weights & $3\times10^{-31}$; $1.6\times10^{-5}$ or $10^{-30}$; 0.054--2.05 & Strict inhibitory margins stabilise unit-locality (\cref{thm:unit}) \\
Nonlinear PC inference & Reachable vs.\ unreachable target; top-down vs.\ feedforward start & Converges when reachable (202 $\to$ 0.012); feedforward start trapped by dead ReLU units & Reachability and non-degenerate initialisation matter (\cref{prop:nonlinearpc}) \\
Presynaptic support & Free-energy vs.\ support-weighted precision, type B with $s<1$ & Free energy selects top level and stalls; support weighting selects level 1 & Illustrates \mref{thm:learnedprec}c and \cref{sec:precision} heuristic \\
Linear regions & Bottleneck followed by a ReLU layer & Interference $0.0000$ vs.\ linear capacity bound 2.886 & Capacity bound applies per linear region (\cref{prop:region}) \\
Mode-locality perturbation & Balanced weights perturbed by $\delta\in[10^{-5},10^{-2}]$ & $\sup_t I/\delta^2 \in [1.991, 1.999]$ & Confirms $O(\delta)$ output robustness (\cref{prop:robust}) \\
\bottomrule
\end{tabular}
\end{table}

\begin{table}[!htbp]
\centering\footnotesize
\caption{Illustrative boundary examples and counterexamples referenced in the main text and proofs.}
\label{tab:examples}
\begin{tabular}{@{}p{0.10\textwidth}p{0.34\textwidth}p{0.50\textwidth}@{}}
\toprule
Example & Configuration & Outcome \\
\midrule
Example 1 & $U=V=\spn(e_1)$, $x=e_2$, $e=e_1+e_2$, $L=3$ & Window $\{1\}$ at onset; level 1 alone leaves squared error 1; joint descent fits the item through the old channel and changes the old output by 1.46 \\
Example 2 & Clipped ReLU, $x=1$, $W_1=2$, $W_2=1$, $y=2$ & $\phi(z)=1$ but $D(z)z=0$; true gradient $-1$, effective linearised gradient 0 (whereas leaky ReLU matches finite differences to $4\times10^{-10}$) \\
Example 3 & Random linear network, $L=3$, loci 1--3 & Reachable $e$ and $h\in\range H$: exact fit with $1\le\rho_m\le$ Kantorovich bound; $h\notin\range H$: interference-free update $<10^{-27}$ vs.\ target delta 0.39--7.0; $h=0$: no output change \\
Example 4 & Type B, $L=2,3,4$, $\gamma=0.8$, $s_i=2$ & Finite-step gradient descent approaches the joint-flow endpoint linearly in step size; level-1-only descent stays $<10^{-28}$ \\
Example 5 & Type D, $L=2,3,4$ & Coordinated finite update achieves misfit $<10^{-15}$ and interference $<3\times10^{-30}$; aligned-flow escape time scales by 2.21 ($L=3$) and 4.07 ($L=4$) upon halving $\varepsilon$ \\
Example 6 & $W_l=\mathrm{diag}(1,0)$, $L=3$, $x=e_2$, $y=e_1$, $x_o=e_1$ & Equal-precision predictive coding changes old output by $\eta/3+2\eta^2/81$ exactly ($\eta=0.1,0.5,1$) \\
Example 7 & Single-mode precision, $s\in\{1,3,0.3\}$; tied shares & $s=1$: shares remain 0.25; unique winner otherwise; tied maximisers preserve equal shares \\
Example 8 & $W_3=\begin{psmallmatrix}1&1\\0&1\end{psmallmatrix}$, $W_2=\mathrm{diag}(1,0)$, $e=(0,1)\T$ & $K_1e=0$ yet $\varepsilon_1=(-1/7,0)\T\neq0$ in a non-balanced network \\
Example 9 & Familiar input: old output 1, new target 2, $\lambda=1$ & Rehearsal compromise optimum is 1.5 with unavoidable interference 0.25 \\
Example 10 & $L=4$, $S=\mathrm{diag}(3,2,0.1)$, orthogonal $u,w$ sharing support & $\kappa_m=0.0099,-0.6024,-0.8519,-0.9085$ (non-monotone); with disjoint support, $\kappa_m$ increases in 200/200 draws \\
Example 11 & Bottleneck $r=2$, old outputs $3u_1$, $2u_2$ & Independent new input attains capacity bound 4; coinciding input $x=x_1$ requires interference at least 10 \\
Example 12 & Segregated ReLU network & Inhibitory cross weights $2\times10^{-31}$; zero cross weights $1\times10^{-30}$ (kink-sensitive); excitatory 4.7 \\
Example 13 & $L=3$, one touched mode, $C_R\neq0$ and $C_R=0$ & Partial-replay leakage-rate formula exact to $3\times10^{-14}$; zero leakage for $C_R=0$ \\
Example 14 & $L=4$, locus 2, type A & Raw-error update discrepancy scales as $\Pi^{-1}$ for both $\eta=1/\norm{h}^2$ and $\eta=1/\norm{u_{m-1}}^2$ \\
\bottomrule
\end{tabular}
\end{table}

\section{Computational methods}
\label{sec:methods}

Linear models use float64 and convolutional networks float32 arithmetic. Squared-gradient-norm shares are $\norm{\nabla_{W_l}\ell}_F^2/\sum_j\norm{\nabla_{W_j}\ell}_F^2$ throughout. Values below $10^{-20}$ are at the level of rounding error in float64 and are drawn at a floor on logarithmic axes; we do not interpret differences between such values.

\subsection{Deep linear networks}
\label{sec:linmethods}

Balanced networks were built as $W_l=R_{l+1}S^{1/L}R_l\T$ with random orthonormal bases.

\emph{\Mref{fig:dln}a.} Networks with $L=4$ and width 20 were trained from weights with standard deviation $0.1/\sqrt n$ by gradient descent (step 0.02, weight decay $10^{-3}$, 30{,}000 steps) on whitened inputs spanning a 5-dimensional subspace, with teacher singular values $3,\dots,1$ (5 networks). Novelty types were defined from each trained network's own SVD, with the rank at the largest singular-value gap, and 20 unit items per type were drawn per network.
\emph{\Mref{fig:dln}b.} Gradient flow (Euler step $2\times10^{-3}$) in a balanced network with $L=4$, width 16 and $s=(3,2.5,2,1.5,1)$; unit items on the mode with $s_i=2$. The network carries an aligned residual weight $\varepsilon=0.1$ on a fresh path so that D can eventually escape; its loss halved at $t=49.4$.
\emph{\Mref{fig:dln}c.} $L=4$, width 20, $s=(4,2.5,1.5,1,0.5)$, 40 isotropic old inputs, 200 familiar items.
\emph{\Mref{fig:dln}d.} Exact endpoint of \mref{thm:typeB}d for $s_i=1$, $c=1$ and $L\in\{2,3,4,6\}$; finite-step gradient descent on all levels (width 10, spectrum $(1,0.7,0.4)$, step $\min(0.02,0.2\gamma^{-3/2})$) to $\norm{e}<10^{-7}$ for the same depths. The one-shot target delta rule and gradient descent on level 1 alone were run for the same items.
\emph{\Mref{fig:saddle}.} $L\in\{2,3,4\}$, width 16, $s=(3,\dots,1)$, 10 old items, Euler step $2\times10^{-3}$. An item counted as half learned when its output along $\hat y$ reached $\sigma/2$. Random residuals were Gaussian with standard deviation $\varepsilon/\sqrt n$ per entry (3 draws).
\emph{\Mref{fig:replay}.} (a) $L=4$, type-B target $1.5u_1$, step 0.01, 20{,}000 steps, 8 mode-sharing and 8 protected old items. (b) $L=3$, width 14, $\lambda=0.5$, step 0.01, 30{,}000 steps; 12 unrehearsed old items on untouched modes and 6 on the touched mode.
\emph{\Mref{fig:precision}.} $L=4$, width 16, $s=(3,2,1,0.6,0.3)$, 20 old items. Raw-error updates used $\eta=1/\norm{u_{m-1}}^2$. The matched-fit criterion was a relative misfit $\norm{f(x)-y}/\norm{y-f_0(x)}\le10^{-3}$; backpropagation used step 0.01. Learned precision used 10 updates, normalised by the minimum after each update and capped at $10^{12}$ (\cref{sec:precision}).

\subsection{Convolutional networks}
\label{sec:cnnmethods}

\paragraph{Architecture and training}
conv($3\to32$, $3{\times}3$) $\to$ conv($32\to64$, stride 2) $\to$ conv($64\to128$, stride 2) $\to$ conv($128\to256$, stride 2) $\to$ global average pooling $\to$ fc($256\to256$) $\to$ fc($256\to10$), with ReLU, Kaiming initialisation and no biases (or, in the control, biases in every layer). The old classes were airplane, automobile, bird, cat and deer (25{,}000 training images, per-channel standardisation). Training used squared loss on one-hot targets, SGD with momentum 0.9, weight decay $5\times10^{-4}$, a one-cycle schedule with peak learning rate 0.05, batch size 128, 40 epochs, random crops (reflection padding 4) and horizontal flips. The five reserved read-out rows (outputs 6--10) were initialised at zero and reset to zero after every step, so their incoming read-out weights are exactly zero. In the biased control, the reserved units' biases were trained like all other parameters (their targets are zero), and all biases were frozen during the learning of new items, which changed weights only. Old-class test accuracy was 81.4\% (bias-free) and 81.9\% (biased), averaged over 5 seeds.

\paragraph{Items and learning}
Learned items (20 per condition) came from two randomly chosen classes (10 each), and 20 further items of the same kind were held out for transfer. Old-class items came from test images 1{,}001--5{,}000 of the old classes; the retention set was the first 1{,}000. New-class items came from the training images of dog, frog, horse, ship and truck. The remappings were A: a cyclic derangement of the old units; B: dog$\to$cat, frog$\to$bird, horse$\to$deer, ship$\to$airplane, truck$\to$automobile. Learning was full-batch gradient descent on $\tfrac12\sum\norm{y-f(x)}^2$ until the loss fell to 5\% of its initial value, with at most 4{,}000 steps. With all layers plastic, the step was $0.5/\lambda_{\max}$ of the full tangent kernel on the learned items, maximised over the conditions of a seed, so it was common across conditions. With one plastic layer $m$ it was $0.5/\lambda_{\max}(\Theta_m)$. The onset rate is $2e\T\Theta e/\norm{e}^2$.

\paragraph{Kernel predictions}
For each single-layer run, the layer-$m$ tangent kernel on the learned items was eigendecomposed, $\Theta_m=Q\Lambda Q\T$. Linearised gradient descent with the same step leaves the residual $(I-\eta\Lambda)^tQ\T e$ and changes the parameters by $J_m\T Q\,\mathrm{diag}\big((1-(1-\eta\lambda_i)^t)/\lambda_i\big)Q\T e$, which was propagated to the retention and transfer images by Jacobian--vector products. The \emph{prospective} prediction chooses $t$ so that the linearised loss reaches the prespecified 5\%, and it is defined only if the linearised model can reach that value (the fraction of $\norm{e}^2$ in the range of $\Theta_m$ exceeds 0.95). The \emph{achieved-fit} prediction chooses $t$ so that the linearised loss equals the loss the real run achieved. For runs that reached the criterion the two agree closely (median rank correlation 0.83 for both); the main text reports the prospective one.

\paragraph{Metrics}
Interference was the mean over the 1{,}000-image retention set (strictly disjoint from all adaptation sets) of the squared change of all ten outputs. In the six-layer CNN benchmark, old-class test accuracy was measured on all 5{,}000 old-class test images (note that for conditions A, B$'$, and C, the 20 adapted old-class images drawn from indices 1{,}001--5{,}000 represent 0.4\% of this accuracy set, bounding any direct overlap effect to at most 0.4 pp; both the 1{,}000-image interference set and the sequential 500-image retention accuracy set are strictly disjoint from all adaptation images). Old-task error was the mean squared error of the ten outputs against one-hot old-class targets on the retention set. Transfer was the fractional reduction of the held-out items' squared error. Related old classes were the source and target classes of the remapping. The novelty profile $\nu_l$ was the fraction of $\norm{h_l(x)}^2$ outside the subspace capturing 95\% of the uncentred energy of 2{,}000 old training images at that layer (\cref{tab:nu}); this is an operational surrogate for the exact condition $x\perp\spn V$, not an exact nullspace test.

\paragraph{Experimental inclusion criteria}
Each seed contributes 18 conditions (6 kinds $\times$ 3 draws) and each condition 6 single-layer runs (\cref{tab:runs}). A run is \emph{reached} if the loss fell to 5\% of its initial value within the step budget, \emph{not reached} if the criterion was reachable but not met within 4{,}000 steps, \emph{unreachable} if more than 5\% of the initial error lies on reserved outputs that the plastic layer cannot change (D$^*$ below the readout; run and reported as partial fits), and \emph{not run} if all initial error lies on zero-weight reserved outputs (C and D below the readout). Correlations and medians of predicted and observed interference use reached runs only; runs that did not reach the criterion and partial fits are reported separately and never pooled with them.

\begin{table}[!htbp]
\centering\footnotesize
\caption{Single-layer run counts (bias-free networks; biased networks in parentheses). Each condition type has 15 draws $\times$ 6 layers = 90 candidate runs.}
\label{tab:runs}
\begin{tabular}{@{}lcccc@{}}
\toprule
Condition & Reached & Not reached & Unreachable (partial fit) & Not run \\
\midrule
A & 81 (88) & 9 (2) & 0 & 0 \\
B & 90 (90) & 0 & 0 & 0 \\
B$'$ & 89 (90) & 1 (0) & 0 & 0 \\
C & 15 (15) & 0 & 0 & 75 (75) \\
D & 15 (15) & 0 & 0 & 75 (75) \\
D$^*$ & 15 (15) & 0 & 75 (75) & 0 \\
\midrule
Total & 305 (313) & 10 (2) & 75 (75) & 150 (150) \\
\bottomrule
\end{tabular}
\end{table}

\begin{table}[!htbp]
\centering\footnotesize
\caption{Bottom-up novelty of the learned items: fraction of activity at the input to each layer lying outside the old images' 95\%-energy subspace (mean over 15 item sets, bias-free networks). New-class and colour-negative images are about as familiar to the network as held-out old images at every layer (maximum difference 0.033).}
\label{tab:nu}
\begin{tabular}{@{}lcccccc@{}}
\toprule
Condition & input & conv1 & conv2 & conv3 & pooled conv4 & fc5 \\
\midrule
A (old images) & 0.066 & 0.143 & 0.138 & 0.142 & 0.052 & 0.050 \\
B (new classes) & 0.071 & 0.152 & 0.150 & 0.152 & 0.054 & 0.059 \\
B$'$ (colour negatives) & 0.060 & 0.135 & 0.130 & 0.153 & 0.063 & 0.083 \\
C (old images) & 0.075 & 0.146 & 0.141 & 0.145 & 0.051 & 0.050 \\
D (new classes) & 0.072 & 0.149 & 0.149 & 0.150 & 0.053 & 0.058 \\
D$^*$ (new classes) & 0.080 & 0.160 & 0.155 & 0.156 & 0.055 & 0.059 \\
\bottomrule
\end{tabular}
\end{table}

\begin{figure}[!htbp]
\centering
\includegraphics[width=\textwidth]{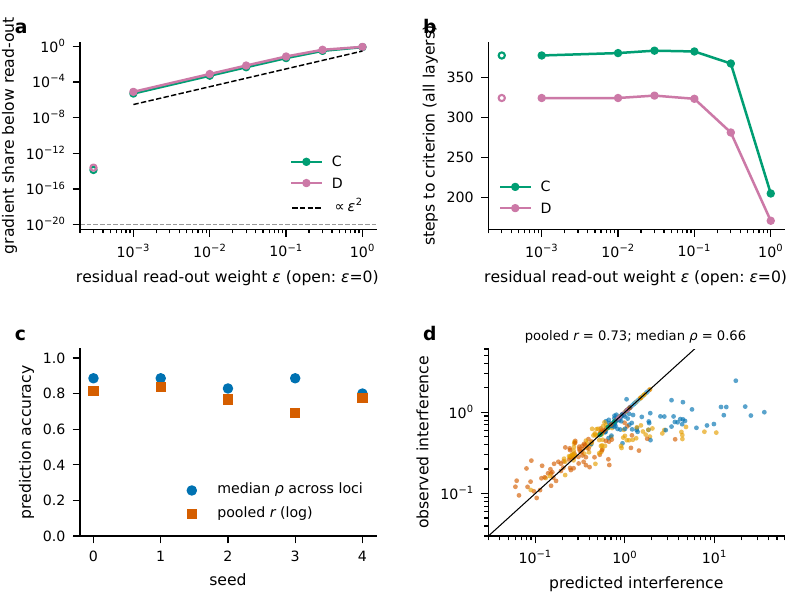}
\caption{\textbf{Controls for the convolutional-network experiment.} (a) Onset share of the squared gradient below the read-out for output-novel items (C, D) when the reserved read-out rows carry residual weights of relative size $\varepsilon$ (median over 15 draws; open symbols: $\varepsilon=0$, at the level of float32 rounding). Dashed: $\propto\varepsilon^2$. (b) Steps to criterion with all layers plastic. (c) Accuracy of the prospective kernel prediction per seed (bias-free networks). (d) Prospective prediction against observation in networks with biases (313 runs that reached the criterion).}
\label{fig:cnncontrols}
\end{figure}

\paragraph{Further results}
Interference was not confined to old classes related to the new items. With all layers plastic, the median interference was 0.14 on the old classes related to a B item set and 0.23 on the others; for A, 0.78 and 0.56. Per seed, the correlation of log predicted and observed interference was 0.69--0.84 and the median rank correlation across loci 0.80--0.89 (\cref{fig:cnncontrols}c). In the biased networks, 313 of 315 reachable single-layer runs met the criterion; C and D were learned in a median of 573 and 487 steps versus 64--76 for A, B and B$'$; the pooled correlation was 0.73 (seed cluster bootstrap 0.67--0.79), 0.79 without the read-out, and the new-class and negative images again differed from held-out old images by at most 0.052 in $\nu_l$.

\paragraph{Residual read-out sweep}
To test how the exact localisation of output-side novelty depends on the zero read-out rows, the reserved rows were set to $\varepsilon\cdot\mathrm{RMS}(W_{\mathrm{old}})\cdot Z$ with standard Gaussian $Z$ and $\varepsilon\in\{0,10^{-3},10^{-2},0.03,0.1,0.3,1\}$ before learning C and D items (5 seeds, 3 draws; step $0.5/\lambda_{\max}$ per condition). The onset gradient share below the read-out grew as $\varepsilon^2$ over three decades ($5\times10^{-6}$, $5\times10^{-4}$ and $0.05$ at $\varepsilon=10^{-3},10^{-2},10^{-1}$ for C), and learning sped up only when $\varepsilon\gtrsim0.3$ (\cref{fig:cnncontrols}a,b). Exact localisation is thus a property of the zero-residual construction, and it degrades smoothly as \cref{prop:soft} predicts.

\subsection{Controlled pixel-space input shifts and ResNet-18 with cross-entropy}
\label{sec:resnetmethods}

\paragraph{Pixel-space input shifts, patch-overlap analysis, and sequential adaptation}
Using the same five pre-trained bias-free and biased checkpoints, we constructed four familiar-output conditions (3 draws per seed, 20 learned items per draw): $\mathrm{B}_{\mathrm{nat}}$ (unseen CIFAR-10 classes mapped onto old units), $\mathrm{B}_{\mathrm{perm}}$ (old-class test images with a fixed random spatial permutation of pixel locations mapped to their original labels), $\mathrm{B}_{\mathrm{noise}}$ (old-class images corrupted by additive zero-mean Gaussian pixel noise with standard deviation $\sigma=2.0$ in standardised pixel coordinates), and $\mathrm{B}_{\mathrm{orth}}$ (synthetic patterns projected strictly orthogonal to the 174-dimensional 95\%-energy uncentred pixel subspace of 2{,}000 old training images, normalised to match old-image root-mean-square amplitude, and mapped to old labels). Step sizes were calibrated per condition from the layer tangent kernel with automatic step-size backoff when large-norm noisy inputs caused initial overshoot.

To reconcile the small difference in input novelty $\nu_0$ between the 500-image retention reference and the 2{,}000-image training reference of \cref{tab:nu}, and to measure local convolutional overlap directly, we extracted all $3\times3\times3$ ($27$-dimensional) spatial patches from old retention images and computed their 95\%-energy subspace (dimension 4 of 27) alongside mean squared cosine Gram overlap across patches and across layer representations. Across the four shifts, $(\nu_0(500), \nu_0(2000), \nu_{\mathrm{patch}}, \text{patch Gram})$ were $(0.103, 0.069, 0.063, 0.409)$ for $\mathrm{B}_{\mathrm{nat}}$, $(0.621, 0.588, 0.486, 0.192)$ for $\mathrm{B}_{\mathrm{perm}}$, $(0.778, 0.755, 0.684, 0.130)$ for $\mathrm{B}_{\mathrm{noise}}$, and $(0.992, 1.000, 0.890, 0.0225)$ for $\mathrm{B}_{\mathrm{orth}}$. Thus, even when full-image cosine overlap is negligible ($2.9\times10^{-5}$ for $\mathrm{B}_{\mathrm{orth}}$), 11.0\% of local $3\times3$ patch energy lies inside the old patch subspace, and post-conv1 representation Gram overlap rises to $0.071$ ($0.126$ after conv2).

In 20-event sequential adaptation streams, we measured final original-task displacement $I_T:=\tfrac{1}{P}\sum_o\norm{f_T(x_o)-f_0(x_o)}^2$ relative to the initial pre-trained network $f_0$ (evaluating retention of the original pre-trained task rather than intermediate adapted events), using identical event sequences across all learning strategies within each seed (\cref{tab:seqdetail}). In initially balanced four-layer linear networks ($n=10$ seeds, width 32, $k=5$ old modes, 30 old inputs), each stream presented 20 alternating single-item events in the fixed order $(\mathrm{B},\mathrm{B},\mathrm{C},\mathrm{A})\times 5$ (distinct orthogonal directions in $V^\perp$ for type B, distinct directions in $U^\perp$ for type C). All-layer backpropagation ran iterative gradient descent to a 5\% relative-error criterion (max 400 steps; reaching 190/200 events, median achieved relative error 0.050), whereas fixed level-1 (with readout fallback for type C) and prospective per-event NTK selection applied exact single-layer target-delta updates on reachable events (recomputing candidate layer interference at current weights before each event). Both single-layer strategies reached 129/200 events (exact fit on reached events; 71/200 subsequent type-B events skipped after earlier type-C readout updates rotated $\range W_L$ away from $\spn U$), so their lower final displacement reflects both single-layer protection on reachable events and lower event completion after output-subspace rotation. In the pre-trained bias-free six-layer CNNs (5 seeds), each stream presented 20 familiar-output 4-item events in the fixed order $(\mathrm{B}_{\mathrm{orth}},\mathrm{B}_{\mathrm{nat}},\mathrm{B}_{\mathrm{noise}},\mathrm{B}_{\mathrm{orth}})\times 5$ (10\% relative-loss stopping criterion, max 300 steps per event; evaluated on a strictly disjoint 500-image old-class retention set). Prospective NTK selection evaluated candidate layers \texttt{conv2}--\texttt{fc6} at the current weights before every event on 200 retention images (excluding \texttt{conv1} because its single-event linearised calibration was poorest and accounted for most unreached single-layer runs; 0/100 events required fallback to \texttt{conv4}). Fixed \texttt{conv4} achieved both lower final forgetting and a higher event-completion rate (89/100) than all-layer BP (83/100) and myopic NTK selection (76/100; picking \texttt{conv4} on 60\%, \texttt{fc5} on 26\%, \texttt{conv2}--\texttt{conv3} on 13\%, and \texttt{fc6} on 1\% of events).

\begin{table}[!htbp]
\centering\scriptsize
\setlength{\tabcolsep}{3pt}
\caption{Sequential 20-event adaptation summary (original-task retention displacement $I_T$ and per-event criterion completion). DLN streams use 10 seeds $\times$ 20 mixed B/C/A events (5\% relative-error target; exact target-delta updates for single-layer rules vs.\ iterative GD for all-layer BP); CNN streams use 5 seeds $\times$ 20 familiar-output events (10\% relative-loss target, disjoint 500-image retention set).}
\label{tab:seqdetail}
\begin{tabular}{@{}lcccc@{}}
\toprule
Model / strategy & Reached & Med.\ rel.\ misfit & Final $I_T$ med.\ (IQR) & Acc.\ drop pp (IQR) \\
\midrule
\multicolumn{5}{l}{\emph{Initially balanced 4-layer linear networks (10 seeds, 200 events)}} \\
All-layer BP (iterative GD) & 190/200 & 0.050 & 3.03 (2.58--3.38) & -- \\
Fixed level 1 (readout fallback) & 129/200 & $<10^{-15}$ (1.0 skipped) & 2.02 (1.84--3.32) & -- \\
Prospective NTK (target-delta) & 129/200 & $<10^{-15}$ (1.0 skipped) & 2.02 (1.84--3.32) & -- \\
\midrule
\multicolumn{5}{l}{\emph{Pre-trained bias-free 6-layer CIFAR-10 CNNs (5 seeds, 100 events)}} \\
Fixed \texttt{conv4} & 89/100 & 0.099 (IQR 0.098--0.100) & 0.148 (0.145--0.226) & 10.0 (9.4--20.4) \\
All-layer BP & 83/100 & 0.099 (IQR 0.098--0.100) & 0.204 (0.196--0.205) & 14.2 (11.0--15.2) \\
Myopic NTK (\texttt{conv2}--\texttt{fc6}) & 76/100 & 0.099 (IQR 0.098--0.100) & 0.285 (0.170--0.330) & 21.8 (12.2--26.4) \\
Fixed readout (\texttt{fc6}) & 94/100 & 0.099 (IQR 0.098--0.100) & 1.019 (0.983--1.054) & 42.4 (28.8--42.6) \\
\bottomrule
\end{tabular}
\end{table}

\paragraph{ResNet-18 architecture, pre-training, and cross-entropy probing}
We implemented a standard CIFAR-10 ResNet-18 \citep{he2016} with a $3\times3$ convolutional stem (64 channels), four residual stages (\texttt{layer1}--\texttt{layer4} with 64, 128, 256, and 512 channels, two basic blocks each, batch normalisation, and ReLU), global average pooling, and a 10-unit linear readout \texttt{fc} whose five reserved rows and biases (outputs 6--10) were initialised and maintained at zero. Five seeds were pre-trained on the five old CIFAR-10 classes using five-class cross-entropy loss, SGD (momentum 0.9, weight decay $5\times10^{-4}$, one-cycle learning rate peaking at 0.1, batch size 128, 45 epochs, random crops and flips), reaching 96.1\% mean old-class test accuracy (range 95.7--96.5\%).

During adaptation, batch-normalisation running statistics were held in evaluation mode so the network defined a deterministic map of its parameters. Six hierarchical stages (\texttt{conv1}, \texttt{layer1}--\texttt{layer4}, and \texttt{fc}) were probed across seven conditions (3 draws per seed, 20 items each): A, B, $\mathrm{B}_{\mathrm{noise}}$, and $\mathrm{B}_{\mathrm{orth}}$ under five-class cross-entropy on the old logits; C and D under binary cross-entropy with logits on the single target reserved unit (pure output-side novelty leaving old-logit gradients identically zero under \texttt{fc}-only adaptation); and D$^*$ under full ten-class cross-entropy. For each stage $b$, the exact active-logit tangent kernel $\Theta_b=J_bJ_b\T$ was computed in memory-bounded row chunks, and linearised gradient descent on the exact cross-entropy or binary cross-entropy loss was integrated in active logit space to the prespecified 5\% relative loss criterion. One vector--Jacobian product recovered the linearised parameter change $\Delta\theta_b=J_b\T v$, which was propagated to the 1{,}000 held-out retention images via chunked Jacobian--vector products to yield prospective predictions of squared logit change, five-class conditional KL divergence, cross-entropy increase, and accuracy drop. Actual single-stage runs used step size $0.8/(\gamma_L\lambda_{\max}(\Theta_b))$ with $\gamma_L=0.5/n$ for cross-entropy ($0.25/n$ for binary cross-entropy) up to 1{,}500 steps (400 steps for unreachable D$^*$ stages below \texttt{fc}). \Cref{tab:resnetdetail} gives the per-condition calibration and optimal-stage selection breakdown across strictly positive-KL reached runs.

\begin{table}[!htbp]
\centering\scriptsize
\setlength{\tabcolsep}{3pt}
\caption{ResNet-18 prospective block-wise NTK calibration and stage-selection accuracy across strictly positive-KL reached runs (5 seeds $\times$ 3 draws). Zero-KL \texttt{fc}-only runs of C and D ($n=30$) are excluded from logarithmic correlations to prevent floor-induced inflation. Prospective selection evaluated all linearly reachable stages before any run across the 60 five-class conditions (A, B, $\mathrm{B}_{\mathrm{noise}}$, $\mathrm{B}_{\mathrm{orth}}$); 0 of 60 selected stages failed to reach the 5\% criterion.}
\label{tab:resnetdetail}
\begin{tabular}{@{}lccccc@{}}
\toprule
Subset / condition & $n$ & Log-KL $r$ & Med.\ $|\Delta\log_{10}\mathrm{KL}|$ & $\rho_{\mathrm{KL}}$ & Stage-selection / empirical opt. \\
\midrule
All positive-KL reached & 356 & 0.947 (0.927--0.963) & 0.115 ($1.30\times$) & 0.900 & 52/60 exact; 59/60 top-2 ($1.07\times$) \\
Nonlinear blocks (no \texttt{fc}) & 298 & 0.942 (0.923--0.959) & 0.144 ($1.39\times$) & 0.900 & -- (subset excl.\ \texttt{fc}) \\
Condition A (5-class CE) & 90 & 0.871 & 0.058 & 0.771 & \texttt{fc} in 10/15 \\
Condition B (5-class CE) & 89 & 0.973 & 0.090 & 0.943 & \texttt{layer1}--\texttt{layer3} (15/15) \\
Condition $\mathrm{B}_{\mathrm{noise}}$ (5-class CE) & 75 & 0.902 & 0.277 & 1.000 & \texttt{layer3} in 15/15 \\
Condition $\mathrm{B}_{\mathrm{orth}}$ (5-class CE) & 87 & 0.900 & 0.194 & 0.900 & \texttt{layer3} in 15/15 \\
\bottomrule
\end{tabular}
\end{table}

\subsection{Nonlinear predictive coding}
\label{sec:pcnonlinmethods}

We tested precision-selected localisation in two bias-free nonlinear predictive-coding settings minimising $E(u)=\tfrac12\sum_{l=1}^L\pi_l\norm{u_l-W_l\phi_l(u_{l-1})}^2$ with $\phi_1(x)=x$ and $\phi_l=\phi$ for $l\ge2$. Activity inference initialised hidden states $u_1,\dots,u_{L-1}$ at the feedforward activations with $u_L=y$ clamped and minimised $E$ using L-BFGS with strong Wolfe line search for smooth $\tanh$ networks and early-stopping Adam with learning rate $0.05/\sqrt{\max_l\pi_l}$ for piecewise-linear ReLU networks. Weight updates used backtracking line search along the raw-error direction ($\Delta W_l\propto\varepsilon_l\phi_l(u_{l-1})\T$) or the precision-weighted direction ($\Delta W_l\propto\pi_l\varepsilon_l\phi_l(u_{l-1})\T$) until a common relative loss criterion was met.

In the structured four-layer benchmark ($L=4$, dimensions $20\to24\to24\to24\to8$, $n=10$ seeds each for $\tanh$ and ReLU), networks were pre-trained with AdamW on 120 old examples drawn from a 6-dimensional subspace $\spn V$ with nonlinear teacher targets (mean pre-training $R^2=0.998$), followed by subspace projection $W_1\leftarrow W_1P_V$ in float32. Type-B items ($x\perp\spn V$, familiar target) and type-A items ($x\in\spn V$) were adapted to relative misfit $\le10^{-2}$ across $\Pi\in\{1,3,10,30,100,300,1000\}$ with $\pi_1=1$ and $\pi_{l>1}=\Pi$, alongside all-layer and layer-1-only backpropagation baselines (max 12{,}000 steps). In $\tanh$ networks, all learners reached the $10^{-2}$ criterion in 10 of 10 seeds (layer-1-only BP median $I=5.55\times10^{-14}$). In ReLU networks using PyTorch's standard $\phi'(0)=0$ convention, float32 projection leaves rounding residuals ($\sim10^{-7}$) along $V^\perp$ so roughly half of hidden ReLU units begin with small positive pre-activations; under layer-1-only BP, 0 of 10 seeds reached the $10^{-2}$ criterion within 12{,}000 steps (median final relative misfit $0.108$, median $I=2.71\times10^{-12}$), all-layer BP reached criterion in 9 of 10 seeds (median $I=1.13$), raw-error PC ($\Pi=10^3$) reached criterion in 10 of 10 seeds in a median of 5 updates (median $I=0.25$), and precision-weighted PC ($\Pi=10^3$) reached criterion in 4 of 10 seeds (median $I=9.08$).

In the modified MNIST familiar-output adaptation benchmark ($L=3$, dimensions $784\to128\to64\to5$, $\tanh$, $n=10$ seeds), networks were pre-trained on MNIST digits 0--4 \citep{lecun1998} to 99.2\% mean old-class test accuracy. Sets of 10 new items (two per unseen digit 5--9, mapped onto old labels 0--4) were adapted to relative misfit $\le0.05$ both as raw unseen digits and after orthogonalisation against the 87-dimensional 95\%-energy pixel subspace of old digits 0--4. Learners were evaluated by mean squared output interference and old-digit accuracy drop (pp) over 1{,}000 held-out digits 0--4, as well as held-out new-item generalisation accuracy over 200 class-balanced unseen examples (40 per class; 5-class chance 20\%), number of weight updates, and full CPU wall-clock time including all inference iterations, line searches, and stopping checks (\cref{tab:mnistcontrols}). Two-sided exact paired Wilcoxon signed-rank tests compared raw-error precision contrast against all-layer BP, equal-precision PC, and precision-weighted PC both on all 10 seeds (endpoint comparison) and on strictly matched-fit paired subsets where both compared learners reached the stopping criterion (exact $p$-values and sample sizes reported in main \cref{sec:pc}).

\begin{table}[!htbp]
\centering\scriptsize
\renewcommand{\arraystretch}{1.12}
\setlength{\tabcolsep}{4pt}
\caption{Modified MNIST familiar-output adaptation controls ($n=10$ seeds, 3-layer $\tanh$ networks, 5\% relative-misfit target). Old-digit accuracy drop (pp) is reported as median (IQR; max); 200 held-out new examples are class-balanced across 5 classes (chance 20\%; pre-adaptation baseline 9.3\% orthogonal shift, 38.3\% raw unseen digits). Achieved adaptation-set relative misfits were 0.047--0.050 across reached runs (10/10 reached for all learners except precision-weighted PC on raw unseen digits: 7/10 reached). Dashes (--) indicate measurements that were unavailable for the corresponding comparison methods.}
\label{tab:mnistcontrols}
\begin{tabular}{@{}lccccc@{}}
\toprule
Shift / learner & Med.\ $I$ & Old drop (pp) & New gen.\ (\%) & Updates & Time (s) \\
\midrule
\multicolumn{6}{l}{\emph{Orthogonalised pixel shift}} \\
All-layer BP & 0.691 & 31.4 (22.3--35.3; max 37.0) & 20.3 & 11.5 & 0.007 \\
Equal-precision PC ($\Pi=1$) & 0.639 & 24.9 (18.4--29.2; max 36.1) & -- & 119.5 & -- \\
Precision-weighted PC ($\Pi=500$) & 0.641 & 3.6 (0.3--14.5; max 30.1) & -- & 15.0 & -- \\
Layer-1-only BP & $1.38\times10^{-4}$ & 0.0 (0.0--0.0; max 0.0) & 12.0 & 41.0 & 0.021 \\
LR-scaled BP ($\eta_l=\eta/500$) & $3.28\times10^{-4}$ & 0.0 (0.0--0.0; max 0.0) & 12.5 & 40.5 & 0.022 \\
Raw-error PC ($\Pi=500$) & $6.50\times10^{-4}$ & 0.0 (0.0--0.0; max 0.0) & 12.0 & 38.5 & 13.04 \\
\midrule
\multicolumn{6}{l}{\emph{Raw unseen digits 5--9 mapped to labels 0--4}} \\
All-layer BP & 0.569 & 23.8 (12.8--39.5; max 61.7) & 59.0 & 28.0 & 0.015 \\
Equal-precision PC ($\Pi=1$) & 0.537 & 22.3 (12.4--35.7; max 55.3) & -- & 242.5 & -- \\
Precision-weighted PC ($\Pi=500$) & 0.273 & 1.9 (0.8--5.0; max 8.6) & -- & 153.0 & -- \\
Layer-1-only BP & $2.95\times10^{-2}$ & 0.70 (0.45--0.88; max 1.0) & 48.0 & 291.5 & 0.148 \\
LR-scaled BP ($\eta_l=\eta/500$) & $3.61\times10^{-2}$ & 0.65 (0.35--0.88; max 1.3) & 48.0 & 265.5 & 0.135 \\
Raw-error PC ($\Pi=500$) & $2.33\times10^{-2}$ & 0.45 (0.13--0.70; max 0.8) & 43.0 & 50.5 & 12.36 \\
\bottomrule
\end{tabular}
\end{table}

\section{Schema-dependent learning and consolidation: a mechanistic illustration}
\label{sec:schema}

Rats that have learned a set of flavour--place associations in an arena (a schema) learn new pairs in a single trial, and the new memories become independent of the hippocampus within 48 hours: hippocampal lesions at 48 hours spare them, lesions at 3 hours do not. Without a consistent schema, one-trial learning fails \citep{tse2007,tse2011}. In the terms of the theory, a new flavour mapped onto a location in a known arena is a type-B item, and a flavour mapped onto a location in a new arena is a type-D item. The model below illustrates these correspondences qualitatively; it is not fitted to behavioural data, and its mapping from replay events to hours is an assumption.

\paragraph{Model}
The neocortex is a linear network with $L=4$, 40 flavour inputs, hidden width 30 and 20 outputs, and 12 orthonormal flavour codes. Places are the two-dimensional coordinates of wells, embedded linearly in the output space by an orthonormal $20\times2$ matrix per arena; the second arena is orthogonal to the first. Wells lie on a $5\times5$ grid on $[-1,1]^2$ without its centre. There are six original wells, two new wells in arena A and two in arena B. The schema was trained for 60 sessions of 25 full-batch gradient steps (step 0.05, weight decay $2\times10^{-3}$) from weights with standard deviation $0.3/\sqrt n$. Choice probability was a softmax over the current wells of $-\norm{f(x)-y_w}^2/0.05$. In the inconsistent arena the flavour--well assignment was reshuffled every two sessions. The hippocampus is a fast episodic store that replays encoded pairs; each replay is one cortical learning event per new pair, and a lesion ends replay. The network was either used directly or first subjected to synaptic homeostasis: 20{,}000 further steps of rehearsal of the six pairs with weight decay 0.05 (step 0.05), which removes residual weight from unused input directions (\cref{lem:R1}).

For backpropagation (BP), a replay was a gradient step of size 0.1. For predictive coding (PC), a replay was six iterations of support-weighted precision with a Gamma prior, $\pi_l=(n_l+2\alpha)/(\norm{\varepsilon_l}^2\norm{u_{l-1}}^2+2\beta)$ with $\alpha=1$ and $\beta=10^{-3}$ (precision ratios clipped to $[1,10^6]$), followed by a raw-error update with $\eta=1/(\norm{u_{m-1}}^2+0.05)$, where $m$ is the least precise level. Interleaved replay took a gradient step on the new pairs together with the six old pairs, whose targets were the schema's outputs before learning. Memory strength was $\max(0,1-\norm{f(x)-y}^2/\norm{y}^2)$. A pair counted as consolidated once its squared error fell below 10\% of $\norm{y}^2$; runs that did not reach this within 2{,}000 replays are \emph{censored}. Interference is the mean squared change of the six old outputs, reported at consolidation for completed runs and at the 2{,}000-replay horizon otherwise.

\begin{figure}[!htbp]
\centering
\includegraphics[width=\textwidth]{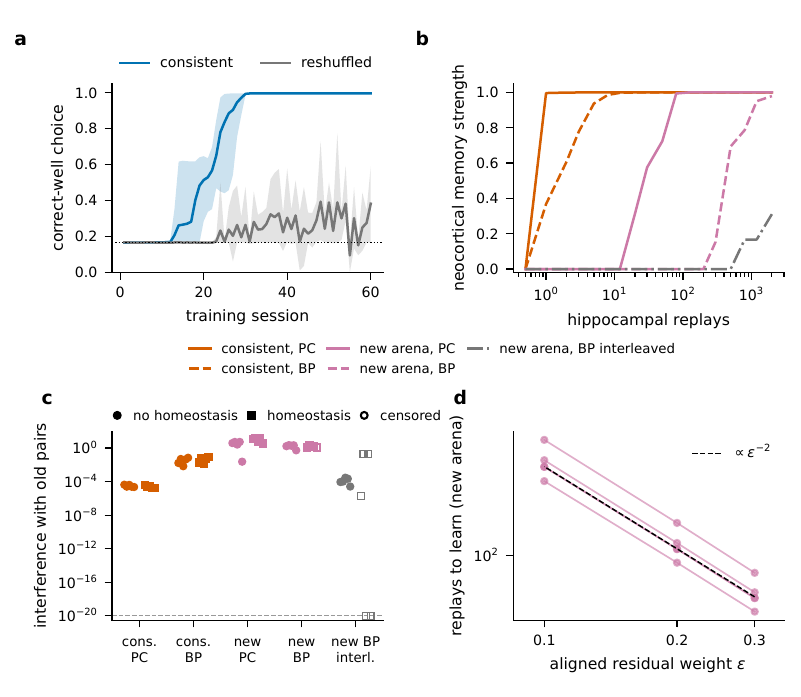}
\caption{\textbf{Schema-dependent learning in a four-level linear model} (5 seeds). (a) Schema acquisition with consistent or reshuffled flavour--place pairs (mean and range). (b) Neocortical memory strength of two new pairs against the number of hippocampal replays, after homeostasis (mean over seeds; 0 replays drawn at 0.5). Interleaved learning is censored at 2{,}000 replays in all seeds. (c) Interference with the six old pairs: at consolidation for completed runs (filled), at the 2{,}000-replay horizon for censored runs (open), without and with homeostasis. (d) Replays needed to learn a new-arena pair along the interference-free path against the aligned residual weight; dashed: $\varepsilon^{-2}$ (\mref{thm:saddle}d).}
\label{fig:schema}
\end{figure}

\paragraph{Results}
With consistent pairs the schema formed after a plateau (24--30 sessions to 90\% correct choices; 5 seeds). With pairs reshuffled every two sessions it did not form (mean correct choice over the last 10 sessions 0.17--0.36; chance 0.17) (\cref{fig:schema}a). After homeostasis, 100\% of the squared gradient for a new schema-consistent pair fell on the bottom level. For a pair in a new arena the gradient norm was $3\times10^{-15}$--$2\times10^{-14}$, a near-saddle: the residual weight on the new flavour's input direction was $7\times10^{-15}$--$8\times10^{-15}$ and that on output directions outside the learned arena $2\times10^{-14}$, at the level of floating-point rounding (in exact arithmetic \cref{lem:R1} predicts $8\times10^{-23}$ for the input side). It is these residuals, amplified along directions of negative curvature, that allow escape.

\Cref{tab:schema} gives the replay counts and interference. PC with support-weighted precision placed plasticity at the bottom level and consolidated each schema-consistent pair after one replay in every seed, with interference at consolidation of $2\times10^{-5}$--$4\times10^{-5}$ (\cref{fig:schema}b,c). BP needed 4--8 replays and caused interference of 0.01--0.08. Thus, ending replay after a single consolidation event spares the new schema-consistent memory under PC while leaving it unintegrated before replay, qualitatively mirroring rapid schema consolidation without imposing a quantitative mapping between replay events and biological hours. A pair in a new arena was learned only after escape from the near-saddle: after homeostasis, PC needed 25--66 replays and BP 378--662 replays in four seeds (censored at 2{,}000 in the fifth), and both severely disrupted the old schema (interference 3.5--17 for PC and 1.1--2.2 for BP) because escape reuses old channels (\mref{thm:saddle}c). Without homeostasis, interleaving the new pairs with replay of the old ones reduced interference at consolidation by about four orders of magnitude ($3\times10^{-5}$--$3\times10^{-4}$ versus 0.5--2.2), at 1.0--1.7 times the replays; after homeostasis, interleaved learning did not finish within 2{,}000 replays in any seed.

Because post-homeostasis escape is otherwise seeded by floating-point rounding residuals, we also ran a controlled perturbation sweep across 10 seeds in which isotropic Gaussian residual weights of explicit scale $\varepsilon\in\{10^{-4},10^{-3},10^{-2},10^{-1}\}$ were injected after homeostasis. Replays to consolidation under PC decreased smoothly and reproducibly with $\varepsilon$, from median 86.5 (IQR 85.2--87.0) at $\varepsilon=10^{-4}$ to 71.0 (69.2--72.0) at $10^{-3}$, 55.0 (54.2--57.0) at $10^{-2}$, and 40.0 (37.2--40.0) at $10^{-1}$, while old-schema interference remained high across all scales (medians 4.72--6.12). Along an interference-free path with an aligned residual weight $\varepsilon$ (\cref{fig:schema}d), replays needed to reach 75\% of target scaled as $\varepsilon^{-2}$ between $\varepsilon=0.3$ and $0.1$ (ratio 9.0--9.4, predicted 9 for $L=4$ by \mref{thm:saddle}d); at $\varepsilon=0.05$, escape along faster interfering directions lowered the count to 366--1{,}060.

The model also makes a prediction beyond the three signatures: the benefit of a schema is specific to the output side. A new cue for a known outcome is fast and safe, whereas a known cue for a new outcome (type C) is learned at the top and interferes through the shared hidden code (\mref{cor:speed}).

\begin{table}[!htbp]
\centering\footnotesize
\caption{Primary schema model after synaptic homeostasis unless stated (5 original seeds; the separate 10-seed controlled Gaussian residual perturbation sweep is reported in the text above). Replays to consolidation and interference at consolidation (range over completed seeds); censored runs did not consolidate within 2{,}000 replays.}
\label{tab:schema}
\begin{tabular}{@{}lllll@{}}
\toprule
Condition & Rule & Completed & Replays & Interference \\
\midrule
schema-consistent (type B) & PC & 5/5 & 1 & $1.7\times10^{-5}$--$3.6\times10^{-5}$ \\
schema-consistent (type B) & BP & 5/5 & 4--8 & 0.012--0.084 \\
new arena (type D) & PC & 5/5 & 25--66 & 3.5--17 \\
new arena (type D) & BP & 4/5 & 378--662 & 1.1--2.2 \\
new arena (type D) & BP, interleaved & 0/5 & censored & (horizon: $2\times10^{-32}$--0.20) \\
\midrule
new arena, no homeostasis & BP & 5/5 & 52--389 & 0.54--2.2 \\
new arena, no homeostasis & BP, interleaved & 5/5 & 87--391 & $2.6\times10^{-5}$--$2.9\times10^{-4}$ \\
\bottomrule
\end{tabular}
\end{table}

\section{Difficulty, locus and specificity in perceptual learning: a mechanistic illustration}
\label{sec:perceptual}

Perceptual learning of fine discriminations is specific to the trained location and orientation, whereas easy discriminations transfer \citep{karni1991,ahissar1997,jeter2009}. Reverse hierarchy theory attributes this to learning at lower levels when high-level representations cannot resolve the task \citep{ahissar2004,hochstein2002}. The model below illustrates how the window produces this pattern.

\paragraph{Model}
The hierarchy is a balanced network with $L=4$, width 16 and $s=(3,2,1,0.6,0.3)$. A discrimination at location A has discriminating direction $d_A=a\,v_1+\sqrt{1-a^2}\,z_A$, where $v_1$ is the first learned input mode (an abstract feature represented up to the top and shared across locations) and $z_A\perp\spn V$ is a location-specific fine feature that the hierarchy discards above level 1. The quantity $1-a^2$, the fraction of discriminating energy lost above level 1, is the task's difficulty. Location B uses $d_B=a\,v_1+\sqrt{1-a^2}\,z_B$ with $z_B\perp z_A$ and a different context. The stimuli are $x_\pm=c\pm0.3\,d$, with targets $f(c)\pm0.6\,u_1$, where $c$ is a random unit context on modes 3--5. Each of 60 trials presented both stimuli. Backpropagation used 10 gradient steps of size 0.05 per trial. Predictive coding used either fixed loci (precision 1 at the locus and $10^6$ elsewhere) or learned support-weighted precision (eight iterations per trial, Gamma prior with $\alpha=1$, $\beta=10^{-4}$, precision ratios clipped to $[1,10^8]$), followed by raw-error updates with $\eta=0.5/\norm{u_{m-1}}^2$. Discrimination is the response difference along $u_1$ divided by its target, 1.2. A task counts as \emph{learned} when discrimination at the trained location reaches 0.9. Transfer is the gain at the untrained location divided by the gain at the trained one; it is reported only for learned tasks, because the ratio is undefined or unstable when the trained gain is near zero. Interference is the mean squared change of 30 old outputs. Results are medians over 10 seeds.

\begin{figure}[!htbp]
\centering
\includegraphics[width=\textwidth]{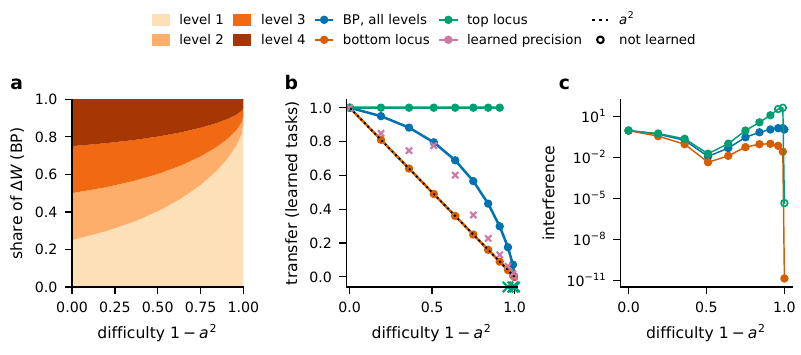}
\caption{\textbf{Difficulty sets the locus and the specificity of perceptual learning} (10 seeds, medians). Difficulty $1-a^2$ is the fraction of discriminating energy lost above level 1. (a) Share of the weight change per level under backpropagation. (b) Transfer to an untrained location for learned tasks (gain there over gain at the trained location). Bottom-locus learning transfers exactly $a^2$ (dotted). Crosses on the axis mark difficulties at which the top-locus learner did not learn the task in at least half of the seeds; no transfer ratio is shown there. (c) Interference with old items; open symbols: task not learned in at least half of the seeds.}
\label{fig:perceptual}
\end{figure}

\paragraph{Results}
Under backpropagation, the share of the weight change at level 1 rose from 0.25 for the easiest task ($a=1$) to 0.87 for the hardest ($a=0$), and transfer to location B fell from 1.0 to 0 (\cref{fig:perceptual}a,b). By \mref{eq:grad}, the fine component contributes only to the level-1 gradient, and it dominates as $a\to0$. Learning at the bottom level transferred exactly the shared discriminating energy, $\kappa_1=a^2$ (to $8\times10^{-7}$; \cref{thm:transfer}), and was interference-free for the purely fine task ($a=0$: $1.5\times10^{-11}$). Learning at the top transferred fully whenever it learned, but it did not learn the tasks with $a\le0.1$ in any seed and learned the task with $a=0.2$ in 4 of 10 seeds; where it partially learned hard tasks, it interfered strongly with old items (median 35--43; \cref{fig:perceptual}c). All other rule--difficulty combinations were learned in 10 of 10 seeds. Support-weighted learned precision placed plasticity at the bottom level in every trial for every task.

Because transfer at the bottom is set by the overlap of the trained and tested discriminating directions, the model is also consistent with the finding that specificity depends on the precision demanded by the transfer task \citep{jeter2009}: a test that requires the fine feature shares little with training at the bottom level. Transfer at higher levels depends on the full weighted kernel of \cref{thm:transfer}, including its cross term, so the model does not imply that higher loci always generalise better.

\end{document}